\documentclass[11pt]{article}
\usepackage{authblk}
\providecommand{\keywords}[1]
{
\small	
\textbf{\textit{Keywords: }} #1
}
\usepackage
[
        a4paper,
        left=2cm,
        right=2cm,
        top=4cm,
        bottom=4cm,
]
{geometry}

\usepackage{bm}
\usepackage{amsmath}
\usepackage{amsfonts, verbatim}
\usepackage{amsthm}            
\usepackage{amssymb}          
\usepackage{bbm}                 
\usepackage{booktabs}          
\usepackage{graphicx}
\usepackage{subcaption}
\usepackage{wrapfig}
\usepackage{float}
\usepackage{algorithm}
\usepackage[noend]{algpseudocode}
\usepackage[normalem]{ulem}
\DeclareMathAlphabet{\mathpzc}{OT1}{pzc}{m}{it}
\usepackage{xcolor}
\usepackage[normalem]{ulem} 
\newcommand{\stkout}[1]{\ifmmode\text{\sout{\ensuremath{#1}}}\else\sout{#1}\fi}
\usepackage{mathtools}
\usepackage{color}
\usepackage{graphicx}
\usepackage{setspace}

\usepackage{tabularx,ragged2e,booktabs,caption}
\newcolumntype{C}[1]{>{\Centering}m{#1}}

\usepackage[colorlinks,citecolor=black,urlcolor=blue,bookmarks=false,hypertexnames=true]{hyperref} 

\usepackage{bm}
\usepackage{amsmath}
\usepackage{amsfonts, verbatim}
\usepackage{amsthm}            
\usepackage{amssymb}          
\usepackage{bbm}                 
\usepackage{booktabs}          
\usepackage{graphicx}
\usepackage{subcaption}
\usepackage{wrapfig}
\usepackage{float}
\usepackage{algorithm}
\usepackage[noend]{algpseudocode}
\usepackage[normalem]{ulem}
\DeclareMathAlphabet{\mathpzc}{OT1}{pzc}{m}{it}
\usepackage{xcolor}
\usepackage[normalem]{ulem} 
\usepackage{mathtools}
\usepackage{color}
\usepackage{graphicx}
\usepackage{setspace}

\usepackage{tabularx,ragged2e,booktabs}
\usepackage[font=small,labelfont=bf]{caption}
\newtheorem{theorem}{Theorem}[section]

\newtheorem{remark}[theorem]{Remark}

\newcommand{\mbf}[1]{\boldsymbol{#1}}

\newcommand{\abs}[1]{\big| #1 \big|}

\newcommand{\rhoT}{\rho_T}
\newcommand{\rhoTL}{\rho_T^L}
\newcommand{\rhoTLM}{\rho_T^{L,M}}
\newcommand{\norm}[1]{\left\| #1 \right\|}
\newcommand{\inprod}[2]{\langle #1, #2 \rangle}
\newcommand{\R}{\mathbb{R}}

\newcommand{\bP}{\mbf{P}}
\newcommand{\br}{\mbf{r}}
\newcommand{\bu}{\mbf{u}}
\newcommand{\bv}{\mbf{v}}
\newcommand{\bV}{\mbf{V}}

\newcommand{\bx}{\mbf{x}}
\newcommand{\bX}{\mbf{X}}
\newcommand{\by}{\mbf{y}}
\newcommand{\bY}{\mbf{Y}}
\newcommand{\bz}{\mbf{z}}
\newcommand{\bZ}{\mbf{Z}}

\newcommand{\bXi}{\mbf{\Xi}}
\newcommand{\muX}{\mu^{\bx}}
\newcommand{\muXi}{\mu^{\xi}}
\newcommand{\muV}{\mu^{\dot\bx}}
\newcommand{\muY}{\mu^{\by}}

\newcommand{\mC}{\mathcal{C}}

\newcommand{\mF}{\mathcal{F}}
\newcommand{\mG}{\mathcal{G}}
\newcommand{\mI}{\mathcal{I}}

\newcommand{\mS}{\mathcal{S}}
\newcommand{\mT}{\mathcal{T}}

\newcommand{\bigO}{\mathcal{O}}
\newcommand{\sdim}{d} 
\newcommand{\numcl}{K}
\newcommand{\idxcl}{k}
\newcommand{\cl}{C}
\newcommand{\clof}{\mathpzc{k}}

\newcommand{\intkernel}{\phi}
\newcommand{\intkernele}{\intkernel^{\bx}}
\newcommand{\intkernela}{\intkernel^{\dot\bx}}
\newcommand{\intkernelxi}{\intkernel^{\xi}}
\newcommand{\bintkernel}{{\bm{\phi}}}
\newcommand{\bintkernele}{\bintkernel^{\bx}}
\newcommand{\bintkernela}{\bintkernel^{\dot\bx}}
\newcommand{\bintkernelxi}{\bintkernel^{\xi}}
\newcommand{\lintkernel}{\hat\intkernel}
\newcommand{\lintkernele}{\lintkernel^{\bx}}
\newcommand{\lintkernela}{\lintkernel^{\dot\bx}}
\newcommand{\lintkernelxi}{\lintkernel^{\xi}}

\newcommand{\blintkernele}{\widehat{\bm{\intkernel}^{\bx}}}
\newcommand{\blintkernela}{\widehat{\bm{\intkernel}^{\dot\bx}}}
\newcommand{\blintkernelxi}{\widehat{\bm{\intkernel}^\xi}}

\newcommand{\intkernelvare}{\varphi^{\bx}}

\newcommand{\bintkernelvar}{{\bm{\varphi}}}
\newcommand{\bintkernelvare}{\bintkernelvar^{\bx}}
\newcommand{\bintkernelvara}{\bintkernelvar^{\dot\bx}}
\newcommand{\bintkernelvarxi}{\bintkernelvar^{\xi}}

\newcommand{\basise}{\psi^{\bx}}

\newcommand{\force}{\mathbf{F}}

\newcommand{\forcex}{\mathbf{F}^{\bx}}
\newcommand{\forcev}{\mathbf{F}^{\dot\bx}}
\newcommand{\forcexi}{\mathbf{F}^{\xi}}
\newcommand{\rhsf}{\mathbf{f}}
\newcommand{\rhsfxnc}{\rhsf^{\text{nc}, \bx}}
\newcommand{\rhsfvnc}{\rhsf^{\text{nc}, \dot\bx}}
\newcommand{\rhsfxinc}{\rhsf^{\text{nc}, \xi}}

\newcommand{\hypspacee}{\mathcal{H}^{\bx}}
\newcommand{\hypspacea}{\mathcal{H}^{\dot\bx}}
\newcommand{\hypspacexi}{\mathcal{H}^{\xi}}

\newcommand{\argmin}[1]{\underset{#1}{\operatorname{arg}\operatorname{min}}\;}

\newcommand{\mand}{\quad \text{and} \quad}

\newcommand{\rev}[1]{\textcolor{black}{{#1}}}

\begin{document}


\title{Data-driven Discovery of Emergent Behaviors \\
in Collective Dynamics}


\author[a, b]{Mauro Maggioni}
\author[a]{Jason Miller}
\author[a]{Ming Zhong}

%


\affil[a]{Department of Applied Mathematics $\&$ Statistics}
\affil[b]{Department of Mathematics}
\affil[ ]{Johns Hopkins University, Baltimore, MD $21218$, USA}

\maketitle

\begin{abstract}
Particle- and agent-based systems are a ubiquitous modeling tool in many disciplines. We consider the fundamental problem of inferring the governing structure, i.e. interaction kernels, \rev{in a nonparametric fashion} from observations of agent-based dynamical systems.  
\rev{In particular, we are interested in collective dynamical systems exhibiting emergent behaviors with complicated interaction kernels, and for kernels which are parameterized by a single unknown parameter.} This work extends the estimators introduced in \cite{PNASLU}, which are based on suitably regularized least squares estimators, to these larger classes of systems.  We provide extensive numerical evidence that the estimators provide faithful approximations to the interaction kernels, and provide accurate predictions for trajectories started at new initial conditions, both throughout the ``training'' time interval in which the observations were made, and often much beyond. We demonstrate these features on prototypical systems displaying collective behaviors, ranging from opinion dynamics, flocking dynamics, self-propelling particle dynamics, synchronized oscillator dynamics, to a gravitational system. Our experiments also suggest that our estimated systems can display the same emergent behaviors of the observed systems, that occur at larger timescales than those in the training data.  Finally, in the case of families of systems governed by a parametric family of interaction kernels, we introduce novel estimators that estimate the parametric family of kernels, splitting it into a common interaction kernel and the action of parameters. We demonstrate this in the case of gravity, by learning both the ``common component'' $1/r^2$ and the dependency on mass, without any a priori knowledge of either one, from observations of planetary motions in our solar system. 
\end{abstract}

\newcommand{\theKeywords}{
\small{\begin{tabular}{c c c c c}
Agent-based Dynamics    & $|$ & Collective Dynamics & $|$ & Emergent Behaviors \\
Nonparametric Inference & $|$ & Machine Learning    & $|$ & Data-driven Method
\end{tabular}}
}


\section{Introduction}
Emergent behavior in collective dynamics, such as clustering of opinions \cite{Krause2000,CKFL2005, BHT2009, MT2014}, flocking of birds \cite{CS2007,CM2008,CD2011}, milling of fish \cite{Chuang2007,Abaid2010,Albi2014,Chuang2016}, and concentric trajectories of planetary motion \cite{logsdon_1998}, is among one of the most interesting phenomena in macroscopic and microscopic scale systems. It occurs in systems used across many disciplines, including biology, social science, particle physics, astronomy, economics, and many more.  Extensive studies have been conducted in order to understand the mechanism behind such intricate and yet geometrically simple behaviors.  As shown in \cite{Vicsek_model,TKIHLC2013,CM2008,GC2004,CHDOB2007,BDT2017,MT2014}, these emergent behaviors are steady-states of various types of collective dynamics, and they can be qualitatively studied when the governing equations are known beforehand.  However, if only the short-time trajectories of the dynamics are observed, it may be challenging to make accurate prediction about the emergent behaviors of the observed dynamics without prior knowledge of the governing equations.  We offer a learning approach to overcome this difficulty by first discovering the governing equations from the observational data, and then use the estimated equations for large-time prediction.  

Research \rev{on} discovering governing equations of dynamical systems has enjoyed a long history in the science and engineering community; it can be traced back to the earlier work of Lagrange, Laplace and Gauss \cite{StiglerHistoryStats}.  Among the many inspiring studies, the lengthy discovery of gravity had immense impact.  In $1605$, Kepler announced his first law of planetary motion, from his work on showing Mars' elliptical orbit based on Tycho Brahe's observational data.  Based on Kepler's first law and the assumption that gravity has a parametric form, namely $\frac{1}{r^p}$, Newton formulated his law of universal gravitation, i.e., that gravity has the form $1/r^2$, in $1687$.  Our learning approach can re-discover the $1/r^2$ form of the law of universal gravitation in a highly efficient and precise manner without the assumption of gravitation having a parametric form and planetary motion being elliptical, for details see Sec. \ref{sec:examples_GSS}.

Our work is focused on discovering governing equations of collective dynamics (also known as self-organized dynamics), a special kind of interacting particle- and agent-based dynamical systems.  
These are a distinguished subset of general autonomous systems of ODEs
\begin{equation}\label{e:dynsysgeneral}
\dot\bX(t) = \rhsf(\bX(t)), \quad \bX(T_0) = \bX_0 \in \R^D \mand t \in [0, T].
\end{equation}
In this general case, given $\{\bX(t)\}_{t \in [T_0, T]}$ ($0 \le T_0 < T$), system identification consists in inferring $\rhsf$ from $\bX(t)$ and $\dot\bX(t)$ observed at various $t$'s. Classical regression techniques (e.g. \cite{Tsybakov:2008, Gyorfi06, devore2006approximation, binev2005universal}) have recently been brought to bear on this problem (e.g. \cite{Schaeffer6634,BPK2016,TranWardExactRecovery,BCGMSVW2012}).  However, lack of independence among the observation data and the curse of dimensionality due to $D$ typically being large, are all obstructions to finding the desired $\rhsf$ effectively and efficiently (see \cite{PNASLU} for an extended discussion) of these points.  
The works in \cite{BFHM17,PNASLU, LTM:MonteCarlo} proposed a learning approach that exploits the special form of collective dynamics to overcome the difficulties mentioned above.  It began with first order systems of the form
\begin{equation}
\dot\bx_i=\frac1N\sum_{i'=1}^N \intkernel(||\bx_{i'}-\bx_i||) (\bx_{i'}-\bx_i)\quad,\quad i=1,\dots,N\,.
\label{e:interactingAgentSystemSimple}
\end{equation}
The $1$-dimensional function $\intkernel$ is referred to as the {\em{interaction kernel}}.  Here and in what follows we assume, with possibly abuse of notation, that the term $i'=i$ in the sum in the r.h.s. is $\bf{0}$, even in cases where $\intkernel$ may not be defined at $0$ (e.g. $\phi(r)=1/r^2$).  The aforementioned works consider the problem of estimating $\intkernel$ given trajectory observations, in terms of positions and velocities of the agents at various times, along one or multiple trajectories (with different initial conditions (ICs), e.g. sampled at random from some probability distribution on the state space).  In \cite{BFHM17,PNASLU} a nonparametric learning approach to construct an estimator $\widehat\intkernel$ for $\intkernel$ is considered, that exploits the governing structure of the dynamics in \eqref{e:interactingAgentSystemSimple}, which is a special (yet ubiquitous) case of the general equations \eqref{e:dynsysgeneral}. The work \cite{BFHM17} considered a first-order model of homogeneous agents (derived from gradient flow), and studied the convergence to its mean field limit and the inference of the mean-field limit interaction kernel from observations of trajectories of the system with a finite and yet increasing number of agents.  The work \cite{PNASLU} extended the approach in \cite{BFHM17} to the situation where the number of agents is fixed, but the number of observations increases, showing that the nonparametric estimators for the interaction kernel converge at the near optimal rate for regression in one dimension, in particular independent of the dimension of the state space. It generalized the estimators to first and second-order of heterogeneous agents with $1$-dimensional interaction kernels based on pairwise distances, providing substantial numerical evidence of the performance of these generalizations.  The work \cite{LTM:MonteCarlo} analyzes in detail the estimators for first order heterogeneous agent-based systems, generalizing the theoretical results of \cite{PNASLU} to that case, while sharpening some of the constructions. 

Here we extend the application of these approaches to rather general classes of agent-based systems, driven by first- and second-order dynamics, interaction kernels depending not only on pairwise distances but also on other pairwise data, which also depends on the states between the agents.  We show that the estimators can be generalized to these settings, and measure their performance at both approximating the interaction kernel $\intkernel$ (in a suitable, dynamics-driven weighted $L^2$-distance) and in predicting trajectories from the same initial conditions (ICs). These estimators may be constructed in a memory-efficient way, i.e. scalable to large data sets with a large number of agents\footnote{Software package available on: \url{https://github.com/MingZhongCodes/LearningDynamics}.}. The estimated interaction kernels, inferred from short-time trajectory data, can provide very good approximations to the original unknown interaction kernels, and yield predictions from new initial conditions.  Furthermore, the estimated interaction kernels can also provide insight into discovering the correct emergent behaviors at large time, as we will demonstrate in several examples in section \ref{sec:examples_emergent} and section \ref{sec:examples_SOD}.  We also extend these estimators to consider not only a single system, but a family of systems, governed by a family of interaction kernels $\{\phi_{\idxcl}\}_{\idxcl}$. We consider the case of gravity in section \ref{sec:examples_GSS}, and show that we can discover both the ``common structure'' of the interaction kernel, namely the $1/r^2$ dependency on pairwise distance, and the dependency on mass. 

The structure of our paper is as follows. In section \ref{sec:model} we discuss in detail the two models, of first and second-order systems respectively, which we are considering. 
In section \ref{sec:algorithm} we outline the learning algorithm for each model. These algorithms are efficient and scalable.  They enjoy highly favorable performance in terms of computational complexity as described in section \ref{sec:computational_comp}.   Section \ref{sec:measures} defines the various performance measures, confusion matrices, and pattern indicator scores.  It also discuses how we set up the numerical experiments.  Section \ref{sec:examples_emergent} to Section \ref{sec:examples_GSS} provide a detailed study of five fundamental dynamical systems that vary across order, interaction kernel form, and agent characteristics, as well as learning of interaction kernels that involve parameters. Finally, we conclude the paper and discuss various future research directions stimulated by these results in section \ref{sec:conclusion}.
\section{Model Description}\label{sec:model}
\rev{The main focus of this work is to numerically investigate the capability of the estimators at predicting emergent behaviors of various collective dynamics using our extended learning approach.  Our extended learning covers dynamical systems much more elaborate than those considered in \cite{PNASLU}.  These systems have a complex balance between non-collective and collective forces, include interaction laws depending on more than one variable, and allow for parametric families of interaction laws.  We motivate these extensions with families of dynamical systems exhibiting these more intricate governing structures, which were motivated by various applications and whose dynamical properties have been studied in the scientific literature.} 

Here we consider particle- and agent-based systems that  model rather general complex systems, beyond those considered in \cite{BFHM17,PNASLU,LTM:MonteCarlo}.
The first-order models are governed by the following system of coupled ODEs
\begin{equation}
\begin{cases}
\dot\bx_i &= \forcex(\bx_i, \xi_i) + \sum_{i'=1}^N \frac{1}{N_{\clof_{i'}}} \intkernele_{\clof_i, \clof_{i'}}(\norm{\bx_{i'} - \bx_i}, s_{i, i'}^{\bx})(\bx_{i'} - \bx_i)\\
\dot\xi_i &= \forcexi(\bx_i, \xi_i) + \sum_{i'=1}^N \frac{1}{N_{\clof_{i'}}} \intkernelxi_{\clof_i, \clof_{i'}}(\norm{\bx_{i'} - \bx_i}, s_{i, i'}^\xi)
\end{cases}\,
\quad,\,i = 1, \cdots, N
\label{eq:1stOrder}
\end{equation}
These systems contain heterogeneous agents: the agents are partitioned into $\numcl$ different types, with $\cl_\idxcl$ containing the indices of the agents of type $\idxcl$, for $\idxcl=1,\dots,\numcl$.  Table \ref{tab:1stOrder_def} shows the definitions of variables in \eqref{eq:1stOrder}.
\begin{table}[H]
\centering
\small{
\small{\begin{tabular}{ c | c }
\hline
Variable                    & Definition \\
\hline\hline
$\bx_i = \bx_i(t) \in \R^d$ & state vector (positions, opinions, etc.) \\
\hline
$\xi_i = \xi_i(t) \in \R$   & auxiliary variable (phase, headings, etc.) \\
\hline
$N$                         & number of agents \\
\hline
$\clof_i$                   & type index of agent $i$ \\
\hline
$N_{\idxcl}$                & number of agents in type $\idxcl$ \\
\hline
$\norm{\cdot}$              & any norm on $\R^d$ (usually an $\ell_2$ norm) \\
\hline
$\forcex, \forcexi$         & non-collective changes on $\dot\bx_i$ and $\dot\xi_i$, respectively \\
\hline
$\intkernele_{\clof_i, \clof_{i'}}, \intkernelxi_{\clof_i, \clof_{i'}}$ & interaction kernels: how the agents in type $\clof_{i'}$ influence agents in type $\clof_i$ \\
\hline
$s_{i, i'}^{\bx}$           & $\mF^{\bx}(\bx_i, \xi_i, \bx_{i'}, \xi_{i'}): \R^{2d + 2} \rightarrow \R$ \\
\hline
$s_{i, i'}^\xi$             & $\mF^{\xi}(\bx_i, \xi_i, \bx_{i'}, \xi_{i'}): \R^{2d + 2} \rightarrow \R$ \\
\hline
\end{tabular}}  
}
\caption{Notation for first-order models}
\label{tab:1stOrder_def} 
\end{table}

\rev{
\begin{remark}
  Compared to Eqn. $(8)$ in \cite{PNASLU}, our new equation \eqref{eq:1stOrder} has the following additions: the new $\xi_i$ variable, non-collective forces $\forcex, \forcexi$, and $2$-dimensional interaction laws (as opposed to only single-variable, pairwise-distance-based interactions).
\end{remark}
}

The second-order models we consider are governed by the following system of coupled ODEs, 
\begin{equation}
\begin{cases}
m_i\ddot\bx_i &= \forcev(\bx_i, \dot\bx_i, \xi_i) + \sum_{i'=1}^N \frac{1}{N_{\clof_{i'}}} \Big[\intkernele_{\clof_i, \clof_{i'}}(\norm{\bx_{i'} - \bx_i}, s_{i, i'}^{\bx})(\bx_{i'} - \bx_i) \\
&\qquad\qquad\qquad\qquad\qquad\qquad\qquad + \intkernela_{\clof_i, \clof_{i'}}(\norm{\bx_{i'} - \bx_i}, s_{i, i'}^{\dot\bx})(\dot\bx_{i'} - \dot\bx_i)\Big]  \\
\dot\xi_i &= \forcexi(\bx_i, \dot\bx_i, \xi_i) + \sum_{i'=1}^N \frac{1}{N_{\clof_{i'}}} \intkernelxi_{\clof_i, \clof_{i'}}(\norm{\bx_{i'} - \bx_i}, s_{i, i'}^\xi)(\xi_{i'} - \xi_i)
\end{cases}\,
\label{eq:2ndOrder}
\end{equation}
for $i = 1, \cdots, N$.  Table \ref{tab:2ndOrder_def} shows the definitions of variables in \eqref{eq:2ndOrder}. 
\rev{Natural regularity and growth assumptions on the functions in the right-hand side are made so that the system has a unique solution for all times. For example assuming that the functions involved are at least Lipschitz and decay sufficiently rapidly at infinity would suffice.}
\begin{table}[H]
\centering
\small{
\small{\begin{tabular}{ c | c }
\hline
Variable                    & Definition \\
\hline\hline
$m_i$                       & mass of agent $i$ \\
\hline
$\forcev, \forcexi$         & non-collective changes on $\ddot\bx_i$ and $\dot\xi_i$ respectively \\
\hline
$\intkernele, \intkernela, \intkernelxi$ & energy, alignment, and environment-based interaction kernels respectively \\
\hline
$s_{i, i'}^{\bx}$           & $\mF^{\bx}(\bx_i, \dot\bx_i, \xi_i, \bx_{i'}, \dot\bx_{i'}, \xi_{i'}): \R^{4d + 2} \rightarrow \R$ \\
\hline
$s_{i, i'}^{\dot\bx}$       & $\mF^{\dot\bx}(\bx_i, \dot\bx_i, \xi_i, \bx_{i'}, \dot\bx_{i'}, \xi_{i'}): \R^{4d + 2} \rightarrow \R$ \\
\hline
$s_{i, i'}^{\xi}$           & $\mF^{\xi}(\bx_i, \dot\bx_i, \xi_i, \bx_{i'}, \dot\bx_{i'}, \xi_{i'}): \R^{4d + 2} \rightarrow \R$ \\
\hline
\end{tabular}}  
}
\caption{Notation for second-order models}
\label{tab:2ndOrder_def} 
\end{table}
\rev{
\begin{remark}
  Compared to Eqn. $(11)$ in \cite{PNASLU}, our new equation \eqref{eq:1stOrder} has the following additions: slightly different non-collective forces, $\forcev, \forcexi$, and $2$-dimensional interaction laws.
\end{remark}
}
We are given observation data, namely $\{\by_i^m(t_l), \dot\by_i^m(t_l)\}_{i, m = 1}^{N, M}$ ($\by_i = \begin{bmatrix} \bx_i \\ \xi_i \end{bmatrix}$ for first order systems or $\by_i = \begin{bmatrix} \bx_i \\ \dot\bx_i \\ \xi_i \end{bmatrix}$ for second order systems) at time instances $T_0 = t_1 < \cdots < t_L = T$.  In the case of missing derivative data, namely $\dot\by_i^m$, we will approximate $\dot\by_i^m$ using appropriate finite difference schemes.  The observation data is generated from $M$ initial conditions (ICs), $\{(\by_i^m(0))_i\}_m$, which are i.i.d samples from a (typically unknown) probability distribution $\mu^{\by}$ ($\mu^{\by} = \mu^{\bx}\oplus\mu^{\xi}$ for first order and $\mu^{\by} = \mu^{\bx}\oplus\mu^{\dot\bx}\oplus\mu^{\xi}$ for second order).  The unknowns in these systems are the interaction laws and the distribution of the initial conditions, while everything else is assumed known. We construct estimators for $\intkernele_{\clof_i, \clof_{i'}}, \intkernelxi_{\clof_i, \clof_{i'}}$ (resp. $\intkernele_{\clof_i, \clof_{i'}}, \intkernela_{\clof_i, \clof_{i'}}, \intkernelxi_{\clof_i, \clof_{i'}}$ for second-order systems) that are close to the true interaction laws with high probability.  Moreover, such estimators yield approximate systems, whose dynamics are approximations to the dynamics of the original system within the training time interval $[T_0, T]$, but can also provide approximations for emergent behaviors of collective dynamics, ranging from first-order opinion dynamics to second-order gravitational dynamics governing the planetary movement in our solar system. A key component of evaluating the emergent dynamics are appropriate measures of the presence of a specific emergent behavior, which will be discussed in \ref{sec:measures}.  
\section{Learning Algorithm}\label{sec:algorithm}
Similar to the the algorithm presented in \cite{PNASLU}, the learning algorithm which we use for the more complex dynamics considered here starts from the introduction of suitable cost functions whose minimizers, over a suitable approximation space, determine the estimators. Equation \eqref{eq:1stOrder} can be rewritten in a more compact form:
\[
\begin{cases}
\dot\bX &= \rhsfxnc(\bX, \bXi) + \rhsf^{\bintkernele}(\bX, \bXi) \\
\dot\bXi &= \rhsfxinc(\bX, \bXi) + \rhsf^{\bintkernelxi}(\bX, \bXi).
\end{cases}
\]
Here $\bX = \begin{bmatrix} \bx_1^\top & \cdots & \bx_N^\top \end{bmatrix}^\top \in \R^{N\sdim}$, $\bXi = \begin{bmatrix} \xi_1 & \cdots \xi_N \end{bmatrix}^\top \in \R^N$; for the interaction kernels, we use the vectorized notations, $\bintkernele = \{\intkernele_{\idxcl, \idxcl'} \in \hypspacee_{\idxcl, \idxcl'}\}_{\idxcl, \idxcl' = 1}^{\numcl}$ and $\bintkernelxi = \{\intkernelxi_{\idxcl, \idxcl'} \in \hypspacexi_{\idxcl, \idxcl'}\}_{\idxcl, \idxcl' = 1}^{\numcl}$, and $\rhsf^{\intkernele}, \rhsf^{\intkernelxi}$ are the collection of the corresponding right hand side terms in \eqref{eq:1stOrder} respectively.  \rev{Lastly,} $\rhsfxnc(\bX, \bXi)$ \rev{is defined as the vectorization of the non-collective forces} $\forcex(\bx_i, \xi_i) \in \mathbb{R}^d$ \rev{and} $\rhsfxinc(\bX, \bXi)$ \rev{is defined as the vectorization of the non-collective forces} $\forcexi(\bx_i, \xi_i) \in \mathbb{R}$.
Our estimators are defined as the minimizers of the loss functions
\[
\begin{cases}
\blintkernele &= \argmin{\bintkernelvare \in \hypspacee} \sum_{m, l = 1}^{M, L}\frac{1}{LM}||\dot\bX^m(t_l) - \rhsfxnc(\bX^m(t_l), \bXi^m(t_l)) - \rhsf^{\bintkernelvare}(\bX^m(t_l), \bXi^m(t_l))||_{\mS(\sdim)}^2 \\
\blintkernelxi &= \argmin{\bintkernelvarxi \in \hypspacexi} \sum_{m, l = 1}^{M, L}\frac{1}{LM}||\dot\bXi^m(t_l) - \rhsfxinc(\bX^m(t_l), \bXi^m(t_l)) - \rhsf^{\bintkernelvarxi}(\bX^m(t_l), \bXi^m(t_l))||_{\mS(1)}^2
\end{cases}\,,
\]
where the $\norm{\cdot}_{\mS(\cdot)}$ norm is defined as
\[
\norm{\bZ}_{\mS(d')}^2 = \sum_{i = 1}^N\frac{1}{N_{\clof_i}}\norm{\bz_i}
\]
for $\bZ = \begin{bmatrix} \bz_1^\top & \cdots \bz_N^\top \end{bmatrix}^\top$ with each $\bz_i \in \R^{d'}$ ($d' = d$ or $1$).  Here $\norm{\cdot}$ is the same norm used in \eqref{eq:1stOrder} and \eqref{eq:2ndOrder}; $\hypspacee = \bigoplus_{\idxcl, \idxcl' = 1}^{\numcl} \hypspacee_{\idxcl, \idxcl'}$ and $\hypspacexi = \bigoplus_{\idxcl, \idxcl' = 1}^{\numcl} \hypspacexi_{\idxcl, \idxcl'}$ are finite-dimensional hypothesis spaces.  We choose each of the hypothesis space $\hypspacee_{\idxcl, \idxcl'}$ be to a finite dimensional function space of piece-wise polynomials of degree $p$, with $p = 0$ or $1$ (polynomials of higher degree can be used and other type of basis functions are also possible, e.g., clamped B-splines,  see \cite{PNASLU}), with polynomial pieces supported on intervals that form a uniform partition of the observed range of variables $[R_{\idxcl, \idxcl', \min}^{\bx}, R_{\idxcl, \idxcl', \max}^{\bx}] \times [S_{\idxcl, \idxcl', \min}^{\bx}, S_{\idxcl, \idxcl', \max}^{\bx}]$.  Hence, each $\intkernelvare_{\idxcl, \idxcl'}$ can be expressed in terms of the linear combination of the basis functions as follows
\[
\intkernelvare_{\idxcl, \idxcl'}(r, s^{\bx}) = \sum_{\eta_{\idxcl, \idxcl'}^{\bx} = 1}^{n_{\idxcl, \idxcl'}^{\bx}} \alpha_{\idxcl, \idxcl', \eta_{\idxcl, \idxcl'}^{\bx}}^{\bx}\basise_{\idxcl, \idxcl', \eta_{\idxcl, \idxcl'}^{\bx}}(r, s^{\bx}).
\]
Similar definitions are used for each $\hypspacexi_{\idxcl, \idxcl'}$.  Substituting this expression into the functionals above, the minimization becomes a set of linear equations,
\[
A^{\bx}\vec{\alpha}^{\bx} = \vec{b}^{\bx} \mand A^{\xi}\vec{\alpha}^{\xi} = \vec{b}^{\xi}.
\]
Here $\vec{\alpha}^{\bx} \in \R^{n^{\bx}}$ is the vector of $\alpha_{\idxcl, \idxcl', \eta_{\idxcl, \idxcl'}^{\bx}}^{\bx}$'s, and $A^{\bx} \in \R^{n^{\bx} \times n^{\bx}}$ with $n^{\bx} = \sum_{\idxcl, \idxcl' = 1}^{\numcl} \eta_{\idxcl, \idxcl'}^{\bx}$; similarly for $\vec{\alpha}^{\xi}$ and $A^{\xi}$.  

\rev{
\begin{remark}
In the case of missing $\dot\bx_i(t)$ (for first order system) or $\ddot\bx_i(t)$ (for second order system), we will approximate it using an appropriate finite difference scheme.  See Sec. \ref{sec:setup_numerics} for details on how we setup the examples with or without derivative information.
\end{remark}
}

In the case of the second-order dynamics described in \eqref{eq:2ndOrder}, we introduce a new variable $\bv_i(t) = \dot\bx_i(t) \in \R^{\sdim}$ and let $\bV = \begin{bmatrix} \bv_1^\top & \cdots & \bv_N^\top\end{bmatrix}^\top$, a compact form of \eqref{eq:2ndOrder} is given as follows,
\[
\begin{cases}
\dot\bX &= \bV \\
\dot\bV &= \rhsfvnc(\bX, \bV, \bXi) + \rhsf^{\bintkernele}(\bX, \bV, \bXi) + \rhsf^{\bintkernela}(\bX, \bV, \bXi) \in \R^{N\sdim} \\
\dot\bXi &= \rhsfxinc(\bX, \bV, \bXi) +  \rhsf^{\bintkernelxi}(\bX, \bV, \bXi) \in \R^N
\end{cases}
\]
Here $\bintkernela = \{\intkernela_{\idxcl, \idxcl'} \in \hypspacea_{\idxcl, \idxcl'}\}_{\idxcl, \idxcl' = 1}^{\numcl}$.  We find the estimators from the following minimizations
\[
\begin{cases}
(\blintkernele, \blintkernela) &= \argmin{\bintkernelvare \in \hypspacee, \bintkernelvara \in \hypspacea}\{\sum_{m, l = 1}^{M, L}\frac{1}{LM}|| \dot\bV^m(t_l) - \rhsfvnc(\bX^m(t_l), \bV^m(t_l), \bXi^m(t_l)) \\
&\hspace{2cm} \qquad - \rhsf^{\bintkernelvare}(\bX^m(t_l), \bV^m(t_l), \bXi^m(t_l)) \\
&\hspace{2cm}\qquad- \rhsf^{\bintkernelvara}(\bX^m(t_l), \bV^m(t_l), \bXi^m(t_l))||_{\mS(d)}^2\} \vspace{0.25cm}\\
\blintkernelxi &= \argmin{\bintkernelvarxi \in \hypspacexi}\{\sum_{m, l = 1}^{M, L}\frac{1}{LM}|| \dot\bXi^m(t_l) - \rhsfxinc(\bX^m(t_l), \bV^m(t_l), \bXi^m(t_l)) \\
&\hspace{2cm}\qquad - \rhsf^{\bintkernelvarxi}(\bX^m(t_l), \bV^m(t_l), \bXi^m(t_l))||_{\mS(1)}^2\}
\end{cases}
\]
Here $\hypspacea = \bigoplus_{\idxcl, \idxcl' = 1}^{\numcl} \hypspacea_{\idxcl, \idxcl'}$.  By choosing appropriate finite dimensional hypothesis spaces for $\hypspacee, \hypspacea$ and $\hypspacexi$, e.g., piece-wise polynomials of degree $p$, we can simplify the least square problems down to the following linear systems which we solve to generate the necessary coefficients:
\[
A\vec{\alpha} = \vec{b} \mand A^{\xi}\vec{\alpha}^{\xi} = \vec{b}^{\xi}.
\]
Here, $\vec{\alpha} = \begin{bmatrix} (\vec{\alpha}^{\bx})^\top & (\vec{\alpha}^{\dot\bx})^\top \end{bmatrix}^\top$ with $\vec{\alpha}^{\bx}$ being the collection of $\alpha_{\idxcl, \idxcl', \eta_{\idxcl, \idxcl'}^{\bx}}^{\bx}$'s and $\vec{\alpha}^{\dot\bx}$ being the collection of $\alpha_{\idxcl, \idxcl', \eta_{\idxcl, \idxcl'}^{\dot\bx}}^{\dot\bx}$'s.
\rev{
\begin{remark}
For further details regarding the construction of the learning matrices, $A, A^{\xi}$, and the right hand side vectors, $\vec{b}, \vec{b}^{\xi}$, we refer the reader to Section $2: Algorithm$ of the Supplementary Information of \cite{PNASLU}.
\end{remark}
}

\subsection{Computational Complexity}\label{sec:computational_comp}
The learning approach, which is described in Sec. \ref{sec:algorithm}, can be easily parallelized in the $m$ (number of initial conditions) variable.  Although it takes $MLD$ double-precision floating-point numbers ($D = Nd + N$ for a first-order system, and $D = 2Nd + N$ for a second-order system) to store the discrete trajectory data, each computing core $j$ only needs to store $M_jLD$ floating-point numbers, with $M_j \approx \frac{M}{\text{Number of Cores}}$.  Furthermore, each computing core does not need to hold all of the trajectory data in memory, since the assembly of the learning matrix and the right-hand-side vector needs only $LD$ floating-point numbers (one system trajectory at a time).  The sizes for the learning matrix and right hand side vector are: $n \times n$ and $n \times 1$ ($n = n^{\bx}$ or $n = n^{\xi}$ for a first-order system and $n = n^{\bx} + n^{\dot\bx}$ or $n = n^{\xi}$ for a second-order system), respectively.  Since we have $n^2 \ll LD$, $n^2 \ll MLD$, which makes solving for our estimators extremely memory efficient.  At each time instance, we have to compute the various pairwise variables, requiring $\bigO(N^2)$ distance calculations, hence the algorithm performs a total of $\bigO(MLN^2)$ computations of pairwise variables.  In solving the linear system, it performs $\bigO(n^3)$ operations (or $\bigO(n^2\log(n))$, we take the worst cases scenario since we use the built-in pseudo-inverse routine in MATLAB to avoid any possible issues with numerical stability).  The total computational complexity is $\bigO(MLN^2 + n^3)$.  Online learning can be built into our learning approach: as trajectory data from different initial conditions comes in, one can simply average the estimators from previous trajectory data with the estimators from the new trajectory data to obtain a better approximation.
\section{Performance Measures}\label{sec:measures}
We consider three different kinds of performance measures: how close the estimated interaction kernel(s) are to the true one(s), how well the trajectories of the system driven by the estimated interaction kernel(s) approximate the trajectories of the original system, and finally how well emergent patterns are reproduced/predicted in the system driven by the estimated interaction kernels. We use appropriate dynamics adapted measures, specified in section \ref{sec:measures2} and the Appendix. 

\subsection{Estimation error of interaction kernels} \label{sec:measures2}
\rev{Following the definitions in \cite{PNASLU}, we introduce a set of probability measures to calculate the learning error between $\blintkernele$ and $\lintkernele$, for any first-order system.}  We define the following probability measures, $\rho_T^{E, \idxcl, \idxcl'}, \rho_T^{L, E, \idxcl, \idxcl'}, \rho_T^{L, M, E, \idxcl, \idxcl'}$, to measure the performance of our estimators.  The first-order measures are as follows,
\begin{equation}
\begin{cases}
\rho_T^{E, \idxcl, \idxcl'}(r, s^{\bx}) &= \frac{1}{N_{\idxcl, \idxcl'}T}\int_{t = 0}^T \mathbb{E}_{\bY_0 \sim \muY}\Big[ \sum_{\substack{i \in \cl_{\idxcl} \\ i' \in \cl_{\idxcl'} \\ i \neq i'}} \delta_{r_{i, i'}(t), s^{\bx}_{i, i'}(t)}(r, s^{\bx}) \Big] \, dt, \\
\rho_T^{L, E, \idxcl, \idxcl'}(r, s^{\bx}) &= \frac{1}{N_{\idxcl, \idxcl'}L}\sum_{l = 1}^L \mathbb{E}_{\bY_0 \sim \muY}\Big[ \sum_{\substack{i \in \cl_{\idxcl} \\ i' \in \cl_{\idxcl'} \\ i \neq i'}} \delta_{r_{i, i'}(t_l), s^{\bx}_{i, i'}(t_l)}(r, s^{\bx})\Big], \\
\rho_T^{L, M, E, \idxcl, \idxcl'}(r, s^{\bx}) &= \frac{1}{N_{\idxcl, \idxcl'}LM}\sum_{l, m = 1}^{L, M} \sum_{\substack{i \in \cl_{\idxcl} \\ i' \in \cl_{\idxcl'} \\ i \neq i'}} \delta_{r_{i, i'}(t_l), s^{\bx}_{i, i'}(t_l)}(r, s^{\bx}). 
\end{cases}
\label{eq:rhoTE} 
\end{equation}
\rev{$\rho_T^{E, \idxcl, \idxcl'}$ (for continuous trajectory) and $\rho_T^{L, E, \idxcl, \idxcl'}$ (for discrete trajectory) are only used in the theoretical setting; in practice, we use $\rho_T^{L, M, E, \idxcl, \idxcl'}$ (with large $M$ and $L$) for actual implementations and applications.}  These measures depend on the dynamical system and the distribution of initial conditions, weighting the areas of pairwise distances (the variable $r$) and of variables $s^{\bx}$ based on how often trajectories of the system explore them. 

Table \ref{tab:rhoT_def} explains the definitions of the variables in \eqref{eq:rhoTE}.
\begin{table}[H]
\centering
\small{
\small{\begin{tabular}{ c | c }
\hline
Variable                    & Definition \\
\hline\hline
$\bY$                       & $\begin{bmatrix} \bX^\top & \bXi^\top \end{bmatrix}^\top$ \\
\hline
$\muY$                      & $\begin{bmatrix} \muX & \muXi \end{bmatrix}^\top$ \\
\hline
$r_{i, i'}(t)$              & $\norm{\bx_{i'}(t) - \bx_i(t)}$ \\
\hline
$s^{\bx}_{i, i'}(t)$        & $\mF^{\bx}(\bx_i(t), \xi_i(t), \bx_{i'}(t), \xi_{i'}(t)): \R^{2d + 2} \rightarrow \R$ \\
\hline
$N_{\idxcl, \idxcl'}$       & $\left\{\begin{array}{ll} N_{\idxcl}(N_{\idxcl} - 1) & \quad \text{if $\idxcl = \idxcl'$}, \\ N_{\idxcl}N_{\idxcl'} & \quad \text{if $\idxcl \neq \idxcl'$.}\end{array}\right.$\\
\hline
\end{tabular}}
}
\caption{$\rho_T$'s, Definition of the Variables}
\label{tab:rhoT_def} 
\end{table}
In the case of $N_{\idxcl, \idxcl'} = 0$, we define the corresponding $\rho_T^{E, \idxcl, \idxcl'}(r, s^{\bx})$ to a zero function.  

We measure the error of the interaction kernel estimators, $\intkernele_{\idxcl, \idxcl'} - \lintkernele_{\idxcl, \idxcl'}$, using the dynamics-induced weighted $L^2$ norm
\begin{equation}\label{eq:L2rhoTE}
\norm{\intkernele_{\idxcl, \idxcl'} - \lintkernele_{\idxcl, \idxcl'}}_{L^2(\rho_T^{E, \idxcl, \idxcl'})}^2 = \int_{r = 0}^\infty \int_{s^{\bx} =-\infty}^\infty (\intkernele_{\idxcl, \idxcl'}(r, s^{\bx}) - \lintkernele_{\idxcl, \idxcl'}(r, s^{\bx})^2  \, r^2d\rho_T^{E, \idxcl, \idxcl'}(r, s^{\bx}).
\end{equation}
However since $\rho_T^{E, \idxcl, \idxcl'}$ is not calculable, we use $\rho_T^{L, M, E, \idxcl, \idxcl'}$ instead.  \rev{The weight, $r^2$, comes from the governing structure of \eqref{eq:1stOrder}.}  Theoretical guarantees such as those in \cite{PNASLU,LTM:MonteCarlo} bound these errors, with high probability, as $M$ grows. Extending those bounds to the general types of systems considered here will be investigated in future work. \rev{The results of our numerical experiments suggest that the learning rate, i.e. the rate of decrease of the error in \eqref{eq:L2rhoTE}, as a function of $M$ is independent of the dimension of the state space of the system, and only depends crucially on the number of variables in the interaction kernel. The curse of dimensionality (of the state space) is therefore avoided.}
\subsection{Trajectory errors}
We consider another performance measure, which might be estimated from data, especially when the true interaction kernel is not known, that quantifies the prediction capability of our estimators, by comparing the observed trajectories to the estimated trajectories evolved from the same initial conditions but using the estimated interaction laws.  We will consider both $\bX(t) = \begin{bmatrix} \bx_1^\top(t) & \cdots \bx_N^\top(t) \end{bmatrix}^\top$ and $\bXi(t) = \begin{bmatrix} \xi_1(t) & \cdots & \xi_N(t) \end{bmatrix}^\top$ for $t \in [T_0, T]$.  Let $\bX_{[T_0, T]} = \{\bX(t)\}_{t \in [T_0, T]}$, then the following norm is used
\begin{equation}\label{eq:traj_norm_x}
\norm{\bX_{[T_0, T]} - \hat\bX_{[T_0, T]}}_{\mT(d)} = \frac{\max_{t \in [0, T]} \norm{\bX(t) - \hat\bX(t)}_{\mS(d)}}{\max_{t \in [0, T]} \norm{\bX(t)}_{\mS(d)}}.
\end{equation}
Here $\hat\bX_{[T_0, T]}$ is the estimated trajectory using our estimators with the same initial condition as in $\bX_{[T_0, T]}$.  The scaling by $\max_{t \in [0, T]} \norm{\bX(t)}_{\mS(d)}$ enables us to compare trajectory errors for different kinds of dynamics, especially those with large $\norm{\bx_i}$.
Similarly, 
\begin{equation}\label{eq:traj_norm_v}
\norm{\bV_{[T_0, T]} - \hat\bV_{[T_0, T]}}_{\mT(1)} = \frac{\max_{t \in [0, T]} \norm{\bV(t) - \hat\bV(t)}_{\mS(d)}}{\max_{t \in [0, T]} \norm{\bV(t)}_{\mS(d)}}, 
\end{equation}
and
\begin{equation}\label{eq:traj_norm_xi}
\norm{\bXi_{[T_0, T]} - \hat\bXi_{[T_0, T]}}_{\mT(1)} = \frac{\max_{t \in [0, T]} \norm{\bXi(t) - \hat\bXi(t)}_{\mS(1)}}{\max_{t \in [0, T]} \norm{\bXi(t)}_{\mS(1)}}.
\end{equation}
For performance measures defined for $\blintkernelxi$ and the second order systems, please see sec. \ref{sec:PM_app} in the appendix.
\subsection{Confusion Matrix and Pattern Indicator Scores}
\rev{When a system is highly sensitive on small perturbations, or even chaotic, it is hopeless to expect that the estimated system will produce trajectories that are accurate approximations of the trajectories of the original system, except perhaps for very small times. However we have observed that certain large-time aspects of the dynamics of the system, such as certain emergent behavior including flocking or milling or clustering, are preserved in our estimated system, even when the trajectory-wise errors are relatively large. On the one hand, this may seem surprising, as at no point do we inject any knowledge about such emergent behaviors, into our estimator; on the other hand if such emergent behaviors are thought of as being ``structurally robust'' to perturbations of the system, and even perturbations of the laws of the system (the interaction kernels), then it becomes reasonable to expect that our estimated systems should preserve, at least to some degree, such emergent behaviors. We therefore introduce a way of measuring quantitatively the presence of such emergent behaviors, and quantify the performance in reproducing them in our estimated systems.}

In order to accurately describe the capability of our estimators to predict the correct \textit{emergent} behaviors at large time $T_f \gg T$, we consider confusion matrices and ``pattern indicator scores''. These are defined differently for each dynamical system to measure its unique emergent behavior. 

Several aspects of the emergent behaviors that we are interested in are observables (i.e. functions defined on the state variables of the system).  We define various emergent behavior scores, such as the flocking score, the milling score, etc., and choose a target range for the score to be in as an indicator of occurrence of the emergent behavior.  For example, if the flocking score is within $(0.99, 1]$, then flocking occurs.  We calculate these scores on the true and estimated systems (systems with the same initial conditions as the true systems but evolved using the learned interaction law(s)).  From this indicator of whether the emergent behavior occurred in the true/estimated system, we construct a confusion matrix, given as follows (in the case of learning flocking systems), 
\begin{table}[H]
\centering
\small{
\rev{
\begin{tabular}{| c || c | c |} 
\hline
                   & Predicted Non-flocking & Predicted flocking\\
\hline
True Non-flocking  & $p_{1, 1}$             & $p_{1, 2}$ \\
\hline
True flocking      & $p_{2, 1}$             & $p_{2, 2}$\\  
\hline
\end{tabular}
}
}
\caption{\rev{General form of a confusion matrix. Each $p_{i, j}$ shows a probability (represented in percentages) of the combination, e.g., $p_{1, 1}$ is the probability of the predicted system (evolved using the estimated interaction laws) showing non-flocking behavior given that the true system shows non-flocking behavior with the same initial conditions.}}
\label{tab:general_CM}
\end{table}
It is used to present the probability of the occurrence of the desired emergent behaviors in the true and estimated systems.  Namely, if the true systems exhibit flocking with high probability, then the estimated systems should ideally show flocking with similar probability. 

\rev{In order to provide deeper insight about the prediction of emergent behavior via confusion matrices, we also consider the following statistics from the confusion matrix.
\begin{table}[H]
\centering
\small{
\rev{
\begin{tabular}{ c | c | c | c } 
\hline
Accuracy                                                   & Precision                              & Recall     & F Score\\
\hline
$\frac{p_{1, 1} + p_{2, 2}}{\sum_{1 = i, j = 2}p_{i, j}}$  & $\frac{p_{2, 2}}{p_{2, 1} + p_{2, 2}}$ & $\frac{p_{2, 2}}{p_{1, 2} + p_{2, 2}}$ & $\frac{2}{\frac{1}{\text{Prec.}} + \frac{1}{\text{Recal.}}}$\\
\hline
\end{tabular}
}
}
\caption{\rev{Definition of the statistical terms related to the confusion matrix.}}
\label{tab:general_CM_stat}
\end{table}
}

Next we use a more refined measurement, a so-called `pattern indicator score', to further demonstrate the capabilities of the estimated system at predicting emergent behaviors. Besides the emergent behavior scores, there are other quantitative descriptions of the emergent behaviors, such as the center-of-mass velocity in flocking, the common rotational axis in milling, the conservation of total energy in concentric trajectories, etc. The pattern indicator scores use these, sometimes together with the previously defined emergent behavior scores, to measure how well the estimated systems are predicting these observables compared to the true systems.  Details of the definition of the confusion matrices and pattern indicator scores for each dynamics are in Sec. \ref{sec:examples_emergent} to Sec. \ref{sec:examples_GSS}.
\subsection{Setup of the Numerical Experiments}\label{sec:setup_numerics}
Here we describe the general setup for the subsequent sections of experiments. The various dynamical systems we consider exhibit a wide variety of emergent behaviors: clustering, flocking, milling, synchronization, and concentric trajectories.  Different forms of interaction kernels are also considered, i.e., $\phi(r)$, $\phi(r, s)$ and $\phi(r; \bP)$, where $\bP$ is an unknown vector of parameters.  These dynamics range from first-order dynamics of homogeneous agents to second-order dynamics of heterogeneous agents.  \rev{We arrange the examples in three major sections based on the different types of the interaction laws.}

The experiments are setup as follows: we first run $M_\rho$ different initial conditions generated i.i.d from the probability measure $\mu^{\by}$ for initial condition, and evolve\footnote{The evolution of the dynamical system is done using the built-in integrator, $\text{ode}15\text{s}$, of MATLAB with the relative tolerance set at $10^{-8}$ and absolute tolerance set at $10^{-11}$.} the dynamics from $0$ to $T$: the dynamics observed in $[T_0, T]$ is used to compute the probability measures $\rhoTL$'s, which are empirical approximations to the probability measures $\rhoT$'s. We do this only to compute and report the $L^2(\rhoT)$ approximation errors; in practice this step is not required nor needed. Next, we generate another set of $M$ random initial conditions and corresponding trajectories of the dynamics for $t \in [0, T]$, with each dynamics observed at $L$ equidistant times $T_0 = t_1 < t_1 < \cdots < t_L = T$, producing the observation data, i.e. $\{\by_i(t_l)^m\}_{i, l, m = 1}^{N, L, M}$, without the corresponding derivative information (i.e., $\dot\by_i(t_l)^m$ is not given, except for Synchronized Oscillator Dynamics and Gravitational Dynamics), as input to our estimation procedure.  We construct the hypothesis spaces, where the estimators are found, on the learning intervals, e.g. $[r_{\min}^{\idxcl, \idxcl'}, r_{\max}^{\idxcl, \idxcl'}]\times[s_{\min}^{\idxcl, \idxcl'}, s_{\max}^{\idxcl, \idxcl'}]$, derived from the observation data; the numbers of basis functions, as well as their degrees, are reported in each section.  We report the $L^2(\rhoT)$ errors between the estimated and true interaction kernels, as well as the trajectory errors based on the statistics over the training set and over a testing set (with new initial conditions), in the form of $\text{(mean value)} \pm \text{(standard deviation)}$. Then we consider the emergent behavior of the true dynamics and the predicted dynamics at $T_f \gg T$, and evaluate ``pattern indicator scores'' and confusion matrices corresponding to the various kinds of emergent behaviors.  The parameters used by all experiments are reported in table \ref{tab:common_param}.
\begin{table}[H]
\centering
\small{
\small{\begin{tabular}{ c | c | c | c | c }
\hline
$N$  & $M_\rho$  & $T_0$ & $L$   & $\#$ Learning Trials\\
\hline\hline
$20$ & $2000$    & $0$   & $500$ & $10$\\
\hline
\end{tabular}}  
}
\caption{Common Parameters}
\label{tab:common_param} 
\end{table}
\rev{Each section/subsection is presented in a similar manner: we introduce the model and discuss why such model interesting for our learning approach; then, we relate the model equation to the learning paradigm presented in \eqref{eq:1stOrder} and \eqref{eq:2ndOrder}.  Next, we present our learning results in figures and tables, in terms of approximation error, trajectory error, confusion matrix and pattern indicators.  We end with a brief discussion of the learning results.}
\section{Emergent Behaviors Induced by $\intkernel(r)$}\label{sec:examples_emergent}
\rev{We consider here three prototypical types of emergent behavior: clustering, flocking and milling.  We examined four different examples of collective dynamics in order to comprehensively explore all three types of emergent behaviors, with very different dynamical behaviors.}
\subsection{Opinion Dynamics}\label{sec:examples_OD}
The opinion dynamics (OD) model, first introduced in \cite{Krause2000}, is a prototypical first-order model of homogeneous agents which describes the interaction of people's opinions through time, see details and extensions in \cite{CKFL2005,BHT2009,MT2014,BT2015,JABIN2014,Dolfin2015}.  These models have gained popularity in modeling human's social behavior, and they can be used to predict interesting social phenomena, namely, clustering/consensus of opinions.

The governing equations ($\bx_i \in \R^d$ being a vector of opinions) are:
\[
\dot\bx_i =\sum_{i'=1}^N \frac{1}{N} \intkernele(\norm{\bx_{i'} - \bx_i})(\bx_{i'} - \bx_i), \quad \text{for $i = 1, \cdots, N$}.
\]
Here $\intkernele(r) \ge 0$ for all $r \ge 0$.  With the interaction kernels giving attractive influences only, these models are bound to have clusters of opinions at large time.  Table \ref{tab:OD_notation} shows how this dynamical system is mapped to the general form \eqref{eq:1stOrder}.  The parameters used for setting up the experiment used are shown in table \ref{tab:OD_params}.
\begin{table}[H]
    \begin{subtable}{.5\linewidth}
	\centering
	\renewcommand{\arraystretch}{1}
	\small{\begin{tabular}{c | c | c | c | c | c} 
	\hline
	Category          & $\xi_i$     & $\numcl$ & $s^{\bx}_{i, i'}$ & $\forcex(\bx_i)$ & $\intkernele$ \\
	\hline
	Value             & $\emptyset$ & $1$      &$\emptyset$        & $\emptyset$      & non-negative \\
	\hline
	\end{tabular}}
	\renewcommand{\arraystretch}{1}
	\caption{(OD) Mapping to \eqref{eq:1stOrder}}
	\label{tab:OD_notation}
    \end{subtable}%
    \begin{subtable}{.5\linewidth}
	\centering
	\renewcommand{\arraystretch}{1}
	\small{
	\small{\begin{tabular}{ c | c | c | c | c }
	\hline
	$M$   & $d$ &$T_f$ & $T$  & $\muX$   \\
	\hline
	$250$ & $2$ & $50$ & $10$ & Unif. on $[0, 5]^2$  \\
	\hline
	\end{tabular}}  
	}
	\renewcommand{\arraystretch}{1}
	\caption{(OD) Parameters for Experiment Setup}
	\label{tab:OD_params} 
    \end{subtable} 
\caption{Opinion Dynamics (OD)}
\end{table}

\begin{figure}[H]
\begin{subfigure}{\textwidth}
  \centering
  \includegraphics[width=0.7\textwidth]{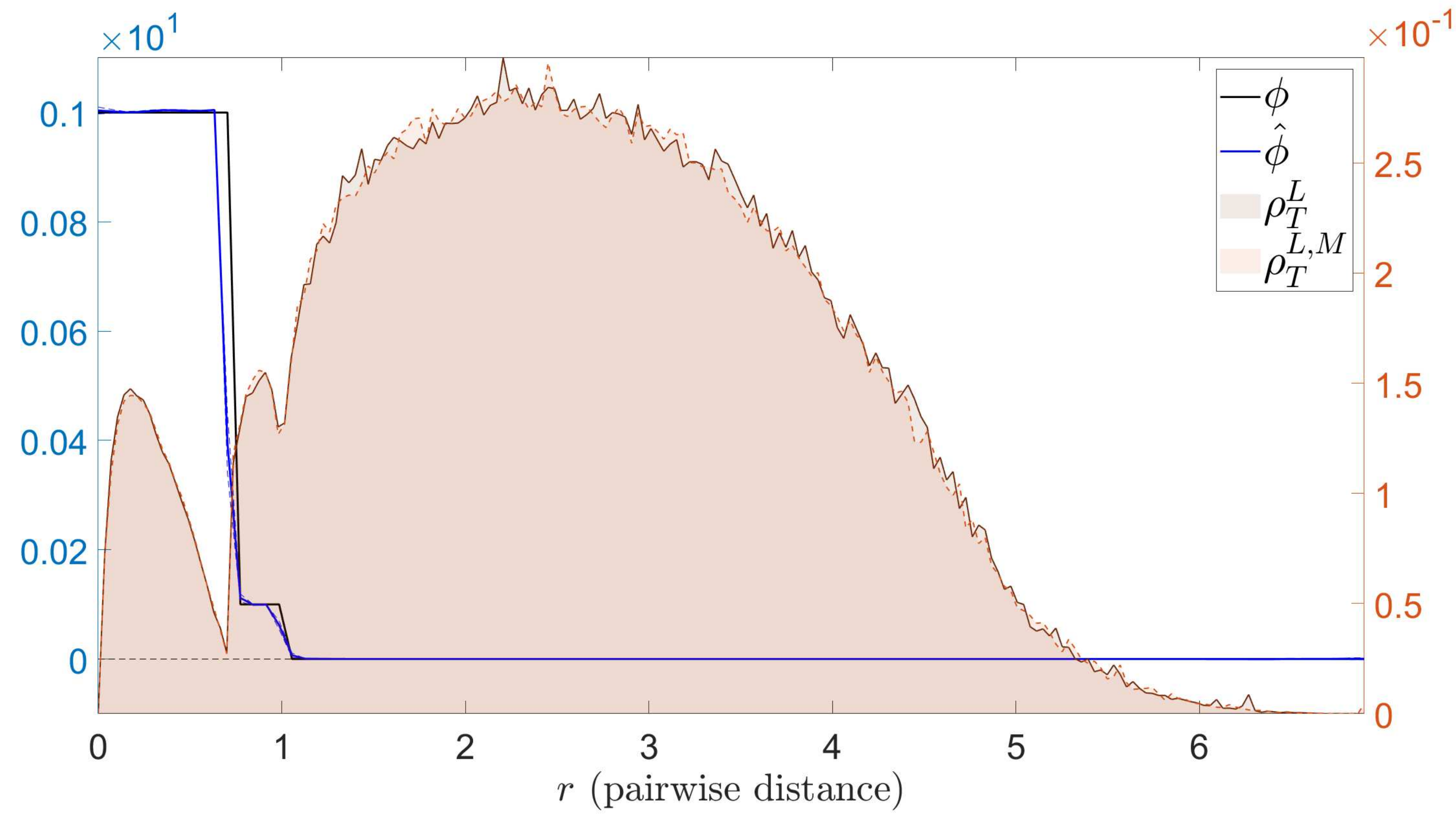}
\end{subfigure}
\caption{(OD) Comparison of $\intkernele$ and $\lintkernele$, with the relative error being $1.4 \cdot 10^{-1} \pm 2 \cdot 10^{-2}$\rev{ (calculated using \eqref{eq:L2rhoTE})}. The true interaction kernel is shown in black solid line, whereas the mean estimated interaction kernel is shown in blue solid line with its confidence interval shown in blue dotted lines.  Shown in the background is the comparison of approximated $\rhoTL$ versus the empirical $\rhoTLM$.}
\label{fig:OD_phiEs}
\end{figure}

\begin{figure}[H]
\begin{subfigure}{\textwidth}
  \centering
  \includegraphics[width=0.7\textwidth]{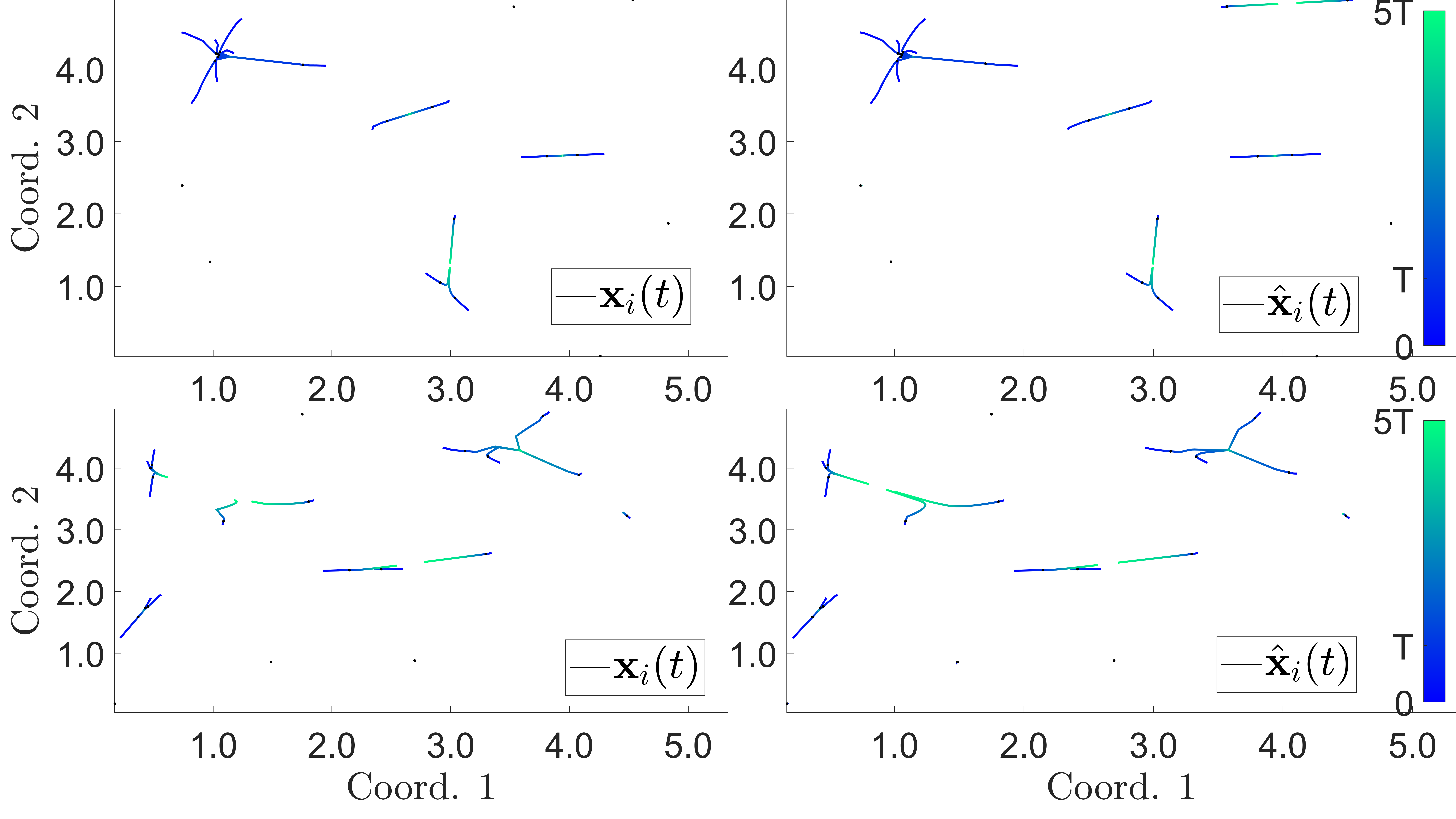} 
\end{subfigure}
\caption{(OD) Comparison of $\bX$ and $\hat\bX$, with the errors reported in table \ref{tab:OD_traj_err}.  The first row of trajectories are generated from an initial condition taken from the observation data.  The second row of trajectories are generated from another randomly chosen initial condition.  The first column of trajectories are generated from the true interaction kernel, whereas the second column of trajectories are generated from our estimated kernel with the same initial conditions.  \rev{The color of the trajectory indicates the flow of time, from deep blue (at $t = T_0$) to light green (at $t = T_f$).}}
\label{fig:OD_trajs}
\end{figure}

We consider the following interaction law,
\[
\intkernele(r) = \left\{ 
\begin{array}{l l }
1    & \quad \text{if $0 \le r < \frac{1}{\sqrt{2}}$} \\
0.1 & \quad \text{if $\frac{1}{\sqrt{2}} \le r < 1$} \\
0    & \quad \text{otherwise}
\end{array}
\right.
\]
Piece-wise constant polynomials with $n^{\bx} = 99$ basis functions are used to approximate $\intkernele$.  The comparison of the true $\intkernele$ and the estimated $\lintkernele$ is shown in Fig.\ref{fig:OD_phiEs}.
As it is shown in Fig. \ref{fig:OD_phiEs}, not only can our estimator detect the discontinuity in the $\intkernel$, but also can it detect the compact support of $\intkernel$.  Meanwhile, there is higher uncertainty in learning the interaction kernel at $r = 0$ (the information of $\intkernele(0)$ is lost since it is weighted by corresponding $\br_{i, i'}$) and at those discontinuity points.  Since $\intkernele$ is non-negative, the agents in the system would eventually converge to clusters, this decreases the effective number of pairwise distance data for inferring $\intkernele$.  However, we are still able to provide an accurate estimator of $\intkernele$ by the continuity of the estimator.  \rev{The comparison of a trajectory driven by the true $\intkernele$ versus the other one driven by the estimated $\lintkernele$} is shown in Fig. \ref{fig:OD_trajs}: there is no major visual difference between the true and predicted trajectories (generated from the training initial condition); the differences are quantified in table \ref{tab:OD_traj_err}.
\begin{table}[H]
\centering
\small{\begin{tabular}{| c || c | c |} 
\hline
                                        & $[0, T]$                                  & $[T, T_f]$\\
\hline
$\text{mean}_{\text{IC}}$: Training ICs & $6.5 \cdot 10^{-3} \pm 7 \cdot 10^{-4}$ & $2.5 \cdot 10^{-2} \pm 2 \cdot 10^{-3}$\\
\hline
$\text{std}_{\text{IC}}$:  Training ICs & $7.4 \cdot 10^{-3} \pm 7 \cdot 10^{-4}$ & $2.5 \cdot 10^{-2} \pm 2 \cdot 10^{-3}$\\
\hline   
\hline         
$\text{mean}_{\text{IC}}$: Random ICs   & $6.6 \cdot 10^{-3} \pm 6 \cdot 10^{-4}$ & $2.73 \cdot 10^{-2} \pm 9 \cdot 10^{-4}$\\
\hline
$\text{std}_{\text{IC}}$:  Random ICs   & $7.4 \cdot 10^{-3} \pm 1 \cdot 10^{-3}$ & $2.7 \cdot 10^{-2} \pm 2 \cdot 10^{-3}$\\
\hline   
\end{tabular}}
\caption{(OD) Trajectory Errors: Initial Conditions (ICs) used in the training set (first two rows), new ICs randomly drawn from $\muX$ (second set of two rows).  \rev{The trajectory errors is calculated using \eqref{eq:traj_norm_x}.}}
\label{tab:OD_traj_err}
\end{table}
The confusion matrix and pattern indicator scores used to examine the capability of our estimators predicting the proper emergent behaviors associated with the Opinion Dynamics model are defined as follows. First, a confusion matrix is used to show the accuracy of our estimator to display the same clustering behavior as the true systems, see the results in table \ref{tab:OD_CM}.
\begin{table}[H]
\centering
\small{\begin{tabular}{| c || c | c |} 
\hline
                                   & Predicted Non-Clustering & Predicted Clustering\\
\hline
True Non-Clustering: Training ICs  & $88 \pm 2 \%$            & $2 \pm 1 \%$ \\
\hline
True Clustering:     Training ICs  & $1.2 \pm 0.6 \%$         & $8.9 \pm 0.2 \%$\\  
\hline
\hline
True Non-Clustering: Random ICs    & $88 \pm 2 \%$            & $1.6 \pm 0.9 \%$ \\
\hline
True Clustering:     Random ICs    & $1.7  \pm 0.8 \%$        & $8 \pm 2 \%$\\
\hline 
\end{tabular}}
\caption{(OD) Confusion Matrix: ICs used in the training set (first two rows), new ICs randomly drawn from $\muX$ (second set of two rows). \rev{It is generated using table \ref{tab:general_CM}.}}
\label{tab:OD_CM}
\end{table}
We provide more statistics about the confusion matrix in order to understand our prediction of clustering better in table \ref{tab:OD_CM_details}.
\begin{table}[H]
\centering
\small{\begin{tabular}{| c || c | c | c | c |} 
\hline
             & Accuracy        & Precision       & Recall           & $F$-Score\\
\hline
Training ICs & $97 \pm 1.5 \%$ & $84 \pm 11.0 \%$ & $88.0 \pm 4.9 \%$ & $85.5 \pm 7.1 \%$ \\
\hline
\hline
Random ICs   & $96.6 \pm 1.2 \%$ & $83.6 \pm 7.5 \%$ & $82.5 \pm 8.4 \%$  & $82.8 \pm 6.5 \%$ \\
\hline
\end{tabular}}
\caption{(OD) Confusion Matrix Statistics: ICs used in the training set, new ICs randomly drawn from $\muX$. \rev{It is generated using table \ref{tab:general_CM_stat}.}}
\label{tab:OD_CM_details}
\end{table}
Next, when the true system has clustering, we want to know if the predicted system can have the same number of clusters as the true system has.  Hence, we assign a score of $1$ when the predicted system shows the same number of clusters as the true systems; and a score of $0$ when it predicts the wrong number of clusters.  $\text{PI}_1$ is the average of those scores over $M$ trials.  Lastly, we want to compare the clusters between the true and predicted systems.  Let $\mC$ contain the centers of the clusters at time $T$\footnote{\rev{The clusters are collection of points such that for each cluster $\mC$, if $\bx_i, \bx_j \in \mC$, then $\norm{\bx_i - \bx_j} < \delta$, and if $\bx_i \in \mC$ and $\bx_j \not\in \mC$, then $\norm{\bx_i - \bx_j} > \delta$.  Here we chose $\delta = 0.01$.}} from the true system, $\hat{\mC}$ contain the centers of clusters from the estimated system; we shall use Hausdorff distance to calculate the distance between $\mC$ and $\hat{\mC}$.  $\text{PI}_2$ is the average of $M$ trials of such distances.  See table \ref{tab:OD_PIs} for details.
\begin{table}[H]
\centering
\small{\begin{tabular}{| c || c | c |} 
\hline
                                        & $\text{PI}_1 $                             & $\text{PI}_2$\\
\hline
$\text{mean}_{\text{IC}}$: Training ICs & $9.2 \cdot 10^{-1} \pm 2 \cdot 10^{-2}$  & $3.2 \cdot 10^{-2} \pm 3 \cdot 10^{-3}$\\
\hline
$\text{std}_{\text{IC}}$:  Training ICs & $2.7 \cdot 10^{-1} \pm 3 \cdot 10^{-2}$  & $4.3 \cdot 10^{-2} \pm 2 \cdot 10^{-3}$\\
\hline            
\hline
$\text{mean}_{\text{IC}}$: Random ICs   & $9.3 \cdot 10^{-1} \pm 2 \cdot 10^{-2}$  & $3.4 \cdot 10^{-2} \pm 3 \cdot 10^{-2}$\\
\hline
$\text{std}_{\text{IC}}$:  Random ICs   & $2.6 \cdot 10^{-1} \pm 3 \cdot 10^{-2}$  & $4.5 \cdot 10^{-2} \pm 2 \cdot 10^{-3}$\\
\hline   
\end{tabular}}
\caption{(OD) Pattern Indicator Scores: ICs used in the training set (first two rows), new ICs randomly drawn from $\muX$ (second set of two rows).}
\label{tab:OD_PIs}
\end{table}
The smaller $\text{PI}_1$ is, the better we are at predicting the number of clusters right.  Meanwhile, not only could we predict the number of clusters with high confidence, but also could we predict the actual location of the clusters.

\subsection{Cucker-Smale Dynamics}\label{sec:examples_CS}
Modeling how animals (or other living agents) move in a cohesive group formation has been a challenging and well-studied problem \cite{CS2007, CS2007a, Choi2017}, \cite{Chuang2007,GAZI2013159,Chuang2016}, \cite{Niwa1994, KTIHC2011,TKIHLC2013}.  There are different degress of cohesion in a collective system: flocking (where each agent shares a common velocity), milling (where each agent rotates around the same axis or about the same oint), and swarming (transition state between flocking and milling).  We first consider the simplest cohesion in a collective system, namely flocking (see detailed work in \cite{Cucker2007,Cucker2010, CD2011, CD2014, Ahn2012, Choi2017, VCBJCS1995}, its mean field limit in \cite{Ha2009,Nourian2011,Shvydkoy2017}, and extension to a stochastic system in \cite{Erban2016} and references therein), and investigate the learnability of these flocking systems.

The Cucker-Smale (CS) dynamics is one of the prototypical examples of flocking agents\footnote{\rev{The Vicsek model in \cite{VCBJCS1995} is a seminal work in modeling flocking of birds, but it uses a different paradigm (different from \eqref{eq:2ndOrder}) to model the flocking behavior.}}.  Its governing equations are
\[
\ddot\bx_i = \sum_{i'=1}^N a_{i, i'}(\bX)(\dot\bx_{i'} - \dot\bx_i), \quad \text{for $i = 1, \cdots N$}.
\]
Here $a_{i, i'}(\bX) = {H}\cdot{(1 + \norm{\bx_{i'} - \bx_i}^2)^{-\beta}}$ where $H, \beta$ are chosen parameters.  Table \ref{tab:CS_notation} shows how this dynamical system is mapped to the general form \eqref{eq:2ndOrder}.
\begin{table}[H]
\centering
	\renewcommand{\arraystretch}{1}
	\small{\begin{tabular}{c | c | c | c | c | c | c | c} 
	\hline
	Category & $m_i$ & $\xi_i$     & $\numcl$ & $s^{\dot\bx}_{i, i'}$ & $\forcex(\bx_i, \dot\bx_i)$ & $\intkernele$ & $\intkernela(r)$\\
	\hline
	Value    & $1$   & $\emptyset$ & $1$      &$\emptyset$            & $\emptyset$      & $\emptyset$   & $\dfrac{H}{(1 + r^2)^\beta}$\\
	\hline
	\end{tabular}}
	\renewcommand{\arraystretch}{1}
	\caption{(CS) Mapping to \eqref{eq:2ndOrder}}
	\label{tab:CS_notation}
	\vspace{0.3cm}
	\renewcommand{\arraystretch}{1}
	\small{\begin{tabular}{ c | c | c | c | c | c }
	\hline
	$M$   & $d$ & $T_f$ & $T$ & $\muX$                & $\muV$\\
	\hline
	$500$ & $2$ & $50$  & $5$ & Unif. on $[-5, 5]^2$  & Unif. on $[-5, 5]^2$\\
	\hline
	\end{tabular}} 
	\renewcommand{\arraystretch}{1}
	\caption{(CS) Parameters for Experiment Setup}
	\label{tab:CS_params} 
\end{table}
With certain choices of $H$ and $\beta$, the Cucker-Smale system is guaranteed to produce flocking (where all agents have the same final velocity) see \cite{Cucker2007}.  For example, when $\beta < \frac{1}{2}$, the system is guaranteed to have flocking regardless of initial conditions; when $\beta = \frac{1}{2}$, the system has conditional flocking depending on the initial configuration of velocities; when $\beta > \frac{1}{2}$, the system has conditional flocking depending on the initial configuration of both positions and velocities.

\begin{figure}[H]
\begin{subfigure}{\textwidth}
  \centering
  \includegraphics[width=0.7\textwidth]{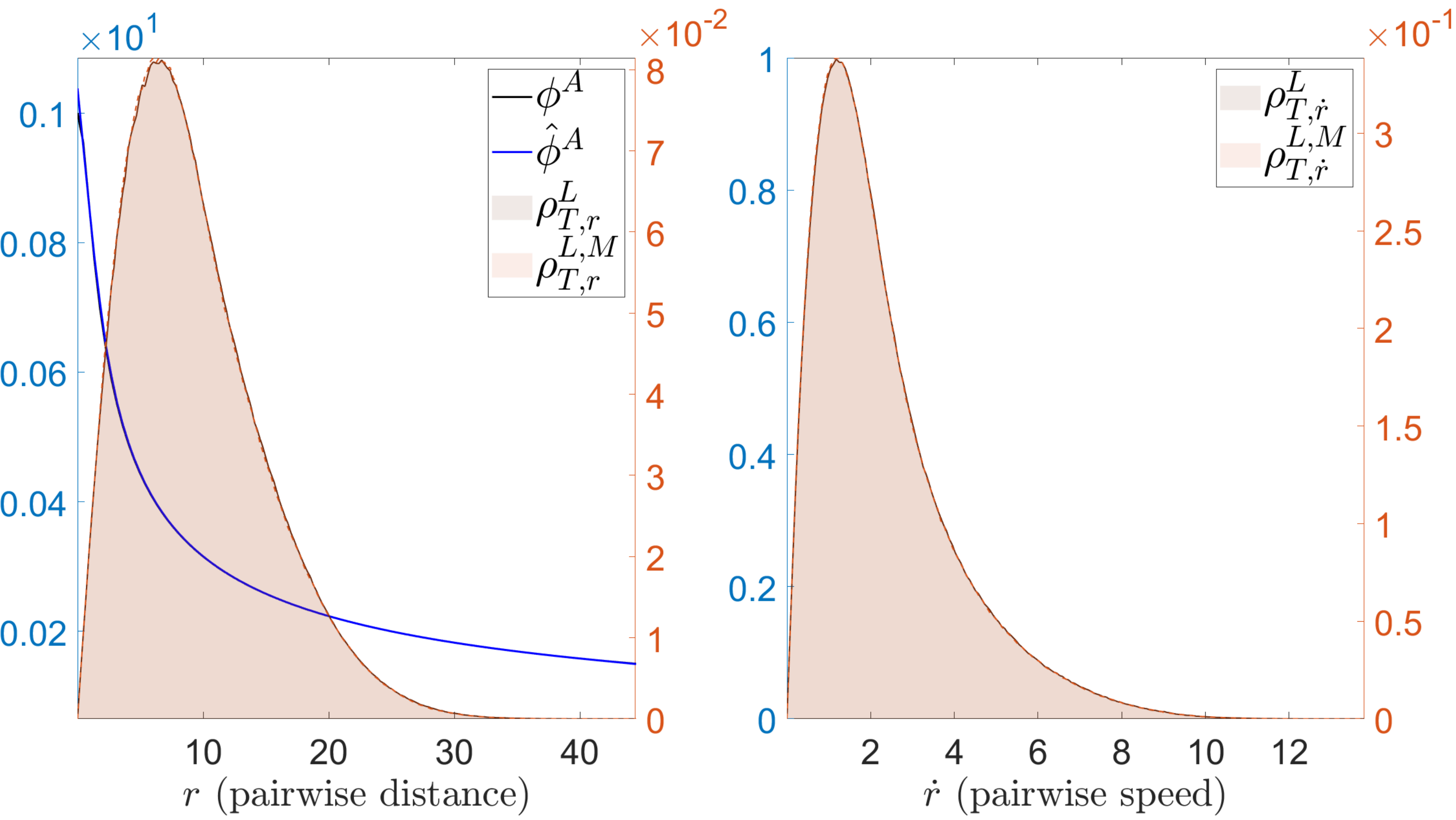}
\end{subfigure}
\caption{(CS) Comparison of $\intkernela$ and $\lintkernela$ together with a plot of $\rho_{T, \dot{r}}^{L}$ versus $\rho_{T, \dot{r}}^{L, M}$, with the relative error being $4.7 \cdot 10^{-3} \pm 2 \cdot 10^{-4}$ \rev{ (calculated using \eqref{eq:L2rhoTA})}. The true interaction kernel is shown by in a black solid line, whereas the mean estimated interaction kernel is shown in a blue solid line with its confidence interval shown in blue dotted lines. Shown in the background is the comparison of approximated $\rhoTL$ versus the empirical $\rhoTLM$.}
\label{fig:CS_phiAs}
\end{figure}

\begin{figure}[H]
\begin{subfigure}{\textwidth}
  \centering
  \includegraphics[width=0.7\textwidth]{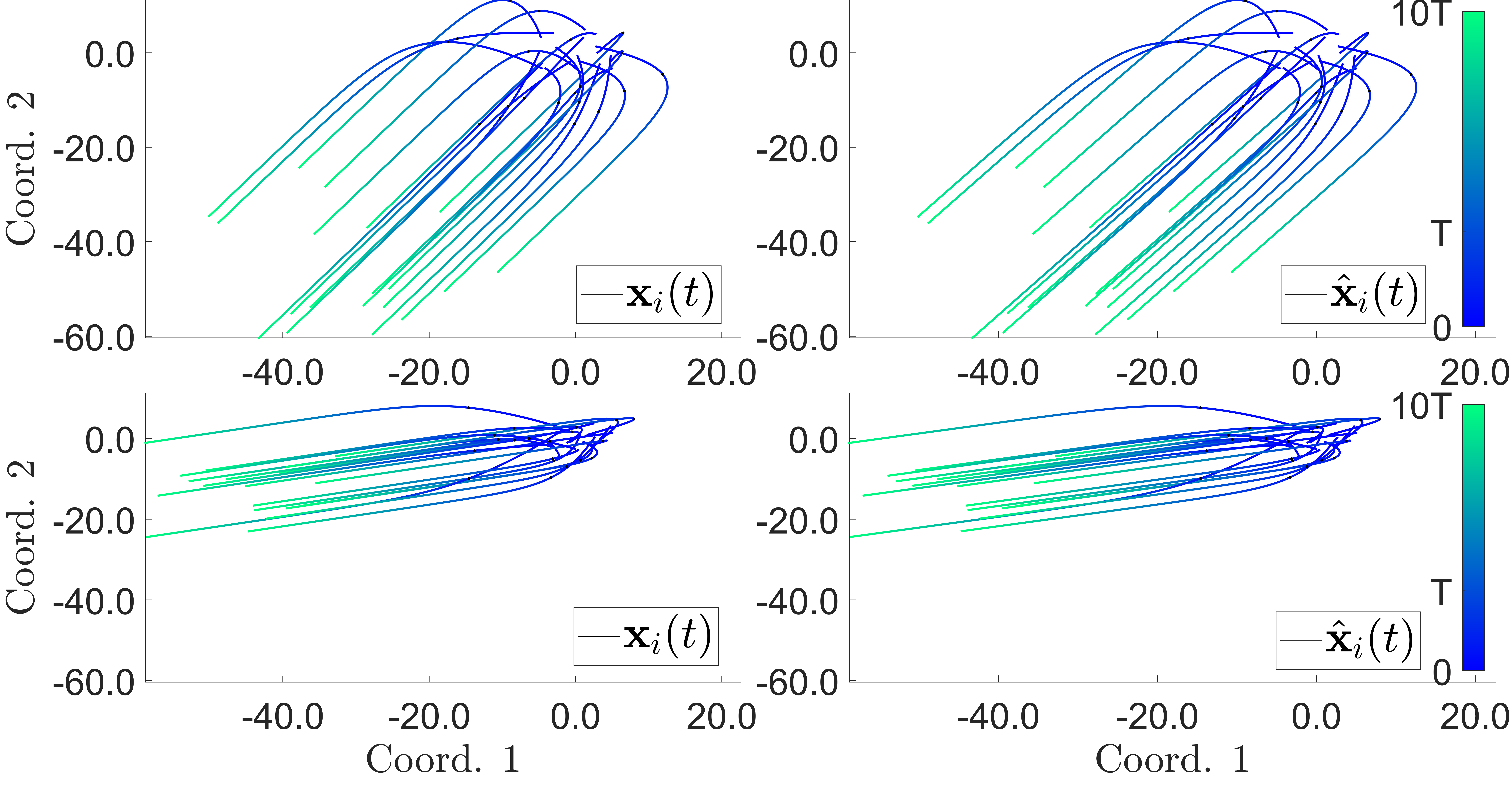} 
\end{subfigure}
\caption{(CS) Comparison of $\bX$ and $\hat\bX$, with the errors reported in table \ref{tab:CS_traj_err}.  The first row of trajectories are generated from an initial condition taken from the observation data. The second row of trajectories are generated from another randomly chosen initial condition. The first column of trajectories are generated from the true interaction kernel, whereas the second column of trajectories are generated from our estimated kernel with the same initial conditions.  \rev{The color of the trajectory indicates the flow of time, from deep blue (at $t = T_0$) to light green (at $t = T_f$).}}
\label{fig:CS_trajs}
\end{figure}

We consider the following interaction law,
\[
\intkernela(r) = \dfrac{1}{(1 + r^2)^{\frac{1}{4}}}.
\]
With this interaction kernel, the agents are guaranteed to flock (see theorem $2, 3$ in \cite{Cucker2007}).  We use the following parameters in table \ref{tab:CS_params} to set up the experiment.  Piece-wise linear polynomials with $n^{\dot\bx} = 100$ basis functions are used to approximate $\intkernela$.  The comparison of the true $\intkernela$ and the estimated $\lintkernela$ is shown in Fig.\ref{fig:CS_phiAs}.
As it is shown in Fig. \ref{fig:CS_phiAs}, our learning approach produce faithful approximation to $\intkernela$, especially capturing the tail behavior of the original interaction law, notwithstanding the scarcity of samples in that region of pairwise distances and speeds; towards $r = 0$, the estimated kernel is also close to the true kernel.  The comparison of true trajectory $\bX(t)$ and learned $\hat\bX(t)$ is shown in Fig. \ref{fig:CS_trajs}.
Fig. \ref{fig:CS_trajs} shows no visual difference between the true trajectories and the learned trajectories (for the training initial condition and a randomly chosen initial condition), we provide a quantitative insight into the difference between trajectories in table \ref{tab:CS_traj_err}.
\begin{table}[H]
\centering
\small{\begin{tabular}{| c || c | c |} 
\hline
                                                 & $[0, T]$                                 & $[T, T_f]$\\
\hline
$\text{mean}_{\text{IC}}$: Training ICs on $\bx$ & $1.55 \cdot 10^{-3} \pm 9 \cdot 10^{-5}$ & $1.9 \cdot 10^{-3} \pm 1 \cdot 10^{-4}$\\
\hline
$\text{mean}_{\text{IC}}$: Training ICs on $\bv$ & $2.7 \cdot 10^{-3} \pm  2 \cdot 10^{-4}$ & $2.8 \cdot 10^{-3} \pm 0 \cdot 10^{-4}$\\
\hline
$\text{std}_{\text{IC}}$:  Training ICs on $\bx$ & $4.1 \cdot 10^{-4} \pm  4 \cdot 10^{-5}$ & $5.5 \cdot 10^{-4} \pm 6 \cdot 10^{-5}$\\
\hline         
$\text{std}_{\text{IC}}$:  Training ICs on $\bv$ & $8.3 \cdot 10^{-4} \pm  8 \cdot 10^{-5}$ & $1.20 \cdot 10^{-3} \pm 9 \cdot 10^{-5}$\\
\hline    
\hline
$\text{mean}_{\text{IC}}$: Random ICs on $\bx$   & $1.5 \cdot 10^{-3} \pm  1 \cdot 10^{-4}$ & $1.8 \cdot 10^{-3} \pm 1 \cdot 10^{-4}$\\
\hline
$\text{mean}_{\text{IC}}$: Random ICs on $\bv$   & $2.6 \cdot 10^{-3} \pm  2 \cdot 10^{-4}$ & $2.7 \cdot 10^{-3} \pm 2 \cdot 10^{-4}$\\
\hline
$\text{std}_{\text{IC}}$:  Random ICs on $\bx$   & $4.1 \cdot 10^{-4} \pm  3 \cdot 10^{-5}$ & $5.5 \cdot 10^{-4} \pm 4 \cdot 10^{-5}$\\
\hline   
$\text{std}_{\text{IC}}$:  Random ICs on $\bv$   & $8.4 \cdot 10^{-4} \pm  6 \cdot 10^{-5}$ & $1.2 \cdot 10^{-3} \pm 1 \cdot 10^{-4}$\\
\hline  
\end{tabular}}
\caption{(CS) Trajectory Errors: Initial Conditions (ICs) used in the training set (first two rows), new ICs randomly drawn from $\muX$ (second set of two rows).  \rev{The trajectory errors in $\bx$/$\bv$ is calculated using \eqref{eq:traj_norm_x}/\eqref{eq:traj_norm_v}.}}
\label{tab:CS_traj_err}
\end{table}
We consider the Flocking score (at $t = T_f$) taken from \cite{CS2007},
\[
I_{\text{flock}} = \sum_{1 = i < i' = N} \norm{\bv_i(T_f) - \bv_{i'}(T_f)}
\]
When $I_{\text{flock}} = 0$, perfect flocking occurs; however we use $I_{\text{flock}} < 0.1$ to indicate flocking.  The prediction capability of the estimated systems in the form of confusion matrix is reported in table \ref{tab:CS_CM}.
\begin{table}[H]
\centering
\small{\begin{tabular}{| c || c | c |} 
\hline
                                 & Predicted Non-Flocking & Predicted Flocking\\
\hline
True Non-Flocking: Training ICs  & $0.0 \pm 0.1 \%$       & $0 \%$ \\
\hline
True Flocking: Training ICs      & $0.0 \pm 0.1 \%$       & $99.9 \pm 0.2 \%$\\  
\hline
\hline
True Non-Flocking: Random ICs    & $0.1 \pm 0.2 \%$       & $0 \%$ \\
\hline
True Flocking: Random ICs        & $0 \%$                 & $99.9 \pm 0.2 \%$\\
\hline 
\end{tabular}}
\caption{(CS) Confusion Matrix: ICs used in the training set (first two rows), new ICs randomly drawn from $\muX$ (second set of two rows).  \rev{It is generated using table \ref{tab:general_CM}.}}
\label{tab:CS_CM}
\end{table}
With $\beta = \frac{1}{4} < \frac{1}{2}$, the true system is guaranteed to show flocking. Since we have no control over when the flocking would occur, we use $I_{\text{flock}}$ to help us capture the essence of flocking behavior, i.e. $I_{\text{flock}} = 0$ (or close to $0$).  And our estimated system can capture the same behavior (flocking or not) with high probability.  We provide more statistics about the confusion matrix in order to understand our prediction of clustering better in table \ref{tab:CS_CM_details}.
\begin{table}[H]
\centering
\small{\begin{tabular}{| c || c | c | c | c |} 
\hline
             & Accuracy           & Precision & Recall             & $F$-Score\\
\hline
Training ICs & $100.0 \pm 0.1 \%$ & $100 \%$  & $100.0 \pm 0.1 \%$ & $99.98 \pm 0.06\%$ \\
\hline
\hline
Random ICs   & $100 \%$           & $100 \%$  & $100 \%$           & $100 \%$ \\
\hline
\end{tabular}}
\caption{(CS) Confusion Matrix Statistics: ICs used in the training set, new ICs randomly drawn from $\muX$. \rev{It is generated using table \ref{tab:general_CM_stat}.}}
\label{tab:CS_CM_details}
\end{table}
In order for us to provide more quantitative insight into the predication capability of our estimator for the case of flocking, we consider two different pattern indicator scores.  First, $\text{PI}_1$ is the relative error of $I_{\text{flock}}$ between true and predicted systems, averaged over $M$ trials.  Second, we consider another important quantity, the center of mass velocity, $\bv_{\text{CM}}$.  It is given by $\bv_{\text{CM}} = \frac{\sum_{i = 1}^N m_i\bv_i(T_f)}{\sum_{i = 1}^N m_i}$. In the case of the CS dynamics ($m_i = 1$ for $i = 1, \cdots, N$), $\bv_{\text{CM}} = \frac{1}{N}\sum_{i = 1}^N \bv_i(T_f)$.  Then we define $\text{PI}_2$ to be the relative error of the predicted center of mass velocity and true center of mass velocity averaged over $M$ trials.  The scores are reported in table \ref{tab:CS_PIs}.  
\begin{table}[H]
\centering
\small{\begin{tabular}{| c || c | c |} 
\hline
                                        & $\text{PI}_1 $                           & $\text{PI}_2$\\
\hline
$\text{mean}_{\text{IC}}$: Training ICs & $1.13 \cdot 10^{-2} \pm 8 \cdot 10^{-4}$ & $6 \cdot 10^{-15} \pm 3 \cdot 10^{-15}$\\
\hline
$\text{std}_{\text{IC}}$:  Training ICs & $5.1  \cdot 10^{-3} \pm 7 \cdot 10^{-4}$ & $3 \cdot 10^{-14} \pm 3 \cdot 10^{-14}$\\
\hline       
\hline     
$\text{mean}_{\text{IC}}$: Random ICs   & $1.12 \cdot 10^{-2} \pm 7 \cdot 10^{-4}$ & $7 \cdot 10^{-15} \pm 4 \cdot 10^{-15}$\\
\hline 
$\text{std}_{\text{IC}}$:  Random ICs   & $5.4  \cdot 10^{-3} \pm 7 \cdot 10^{-4}$ & $4 \cdot 10^{-14} \pm 6 \cdot 10^{-14}$\\
\hline   
\end{tabular}}
\caption{(CS) Pattern Indicator Scores: ICs used in the training set (first two rows), new ICs randomly drawn from $\muX$ (second set of two rows).}
\label{tab:CS_PIs}
\end{table}
As it is shown in table \ref{tab:CS_PIs}, our estimated system can predict $I_{\text{flock}}$ with relatively high accuracy.  Surprisingly, our estimated system can reproduce $\bv_{\text{CM}}$ down to numerical accuracy.

\subsection{Fish Milling in $2$ dimensions}\label{sec:examples_FM2D}
Next we consider a more complicated cohesive collective system: a dynamical system which produces milling patterns, where each agent rotates around the same axis or about the same point.  The models we consider have been proposed in \cite{Chuang2007,Chuang2016} (see references therein for a variety of sources for the biological roots of these models). Useful background references for the two-dimensional models are \cite{Lukeman2008,Abaid2010} as well as the primer \cite{Bernoff2011}. Further theoretical study of models of this type has been done in \cite{Carrillo2009,Degond2013,Albi2014}. 

The governing equations of the Fish Milling Dynamics in $\R^2$ (FM2D) of \cite{Chuang2007} are,
\begin{equation}
m_i\ddot\bx_i = (\alpha - \beta\norm{\dot\bx_i}^2)\dot\bx_i - \nabla_{\bx_i}U_i, \quad \text{for $i = 1, \cdots N$}.
\label{e:FM2Deq}
\end{equation}
Here, $U_i$ is the Morse potential describing the interaction of the $i^{th}$ agent with the other agents in the system, defined as follows
\[
U_i = \sum_{\substack{i' = 1 \\ i' \neq i}}^N \big(-C_{a}\mathrm{e}^{-\frac{\norm{\bx_i - \bx_{i'}}}{\ell_{a}}} + C_{r}\mathrm{e}^{-\frac{\norm{\bx_i - \bx_{i'}}}{\ell_{r}}}\big).
\]
Here $C_a/C_r$ are the attraction/repulsion strengths and $\ell_a/\ell_r$ are the effective attraction/repulsion lengths.  Table \ref{tab:FM2D_notation} shows how the FM2D dynamics fits into the framework of \eqref{eq:2ndOrder}.
\begin{table}[H]
\centering
\renewcommand{\arraystretch}{1}
\small{\begin{tabular}{c | c | c | c | c | c | c} 
\hline
Category & $\xi_i$     & $\numcl$ & $\forcex(\bx_i, \dot\bx_i)$                   & $\intkernela$ & $s^{\bx}_{i, i'}$ & $\intkernele(r)$\\
\hline
Value    & $\emptyset$ & $1$      & $(\alpha - \beta\norm{\dot\bx_i}^2)\dot\bx_i$ & $\emptyset$  & $\emptyset$       & $\frac{N}{r}\big(\frac{C_a}{\ell_a}\mathrm{e}^{-\frac{r}{\ell_a}} - \frac{C_r}{\ell_r}\mathrm{e}^{-\frac{r}{\ell_{r}}}\big)$\\
\hline
\end{tabular}}
\renewcommand{\arraystretch}{1}
\caption{(FM2D) Mapping to \eqref{eq:2ndOrder}}
\label{tab:FM2D_notation}
\vspace{0.3cm}
\renewcommand{\arraystretch}{1}
\small{\begin{tabular}{ c | c | c | c | c | c | c | c | c}
\hline
$M$   & $d$ & $\alpha$ & $\beta$ & $T_f$ & $T$ & $\muX$              & $\muV$\\
\hline
$500$ & $2$ & $1.6$    & $0.5$   & $20$ & $4$ & Unif. on $[0, 1]^2$  & Unif. on $[0, 0]^2$\\
\hline
\end{tabular}}
\renewcommand{\arraystretch}{1}
\caption{(FM2D) Parameters for Experiment Setup}
\label{tab:FM2D_params} 
\end{table}
The delicate balance between the self-propelling force produced by $\forcex(\bx_i, \dot\bx_i)$ and the collective force induced by the energy kernel $U_i$ can create a wide range of patterns for different initial conditions. Milling patterns (single or double milling) are one of the most interesting ones.  Unlike the well-understood Cucker-Smale model, necessary and sufficient conditions on the interaction kernels and ICs that guarantee the existence milling patterns seem to be unknown.  These milling patterns result from the balance of the non-collective force and the collective force induced by the energy kernel $U_i$ (especially when $U_i$ is not $H$-stable, double-milling would occur, see Fig. $1$ in \cite{Chuang2007}), and are therefore rather sensitive to the selection of parameters: relatively small differences in the interaction laws can correspond to dynamical systems with very different dynamical patterns.  The estimator error between the true and estimated interaction kernel may therefore offer little insight information into how well our estimated dynamics can re-produce milling patterns at large time. The confusion matrix and pattern indicator scores are finer indicators of performance in this case.

\begin{figure}[H]
\begin{subfigure}{\textwidth}
  \centering
  \includegraphics[width=0.7\textwidth]{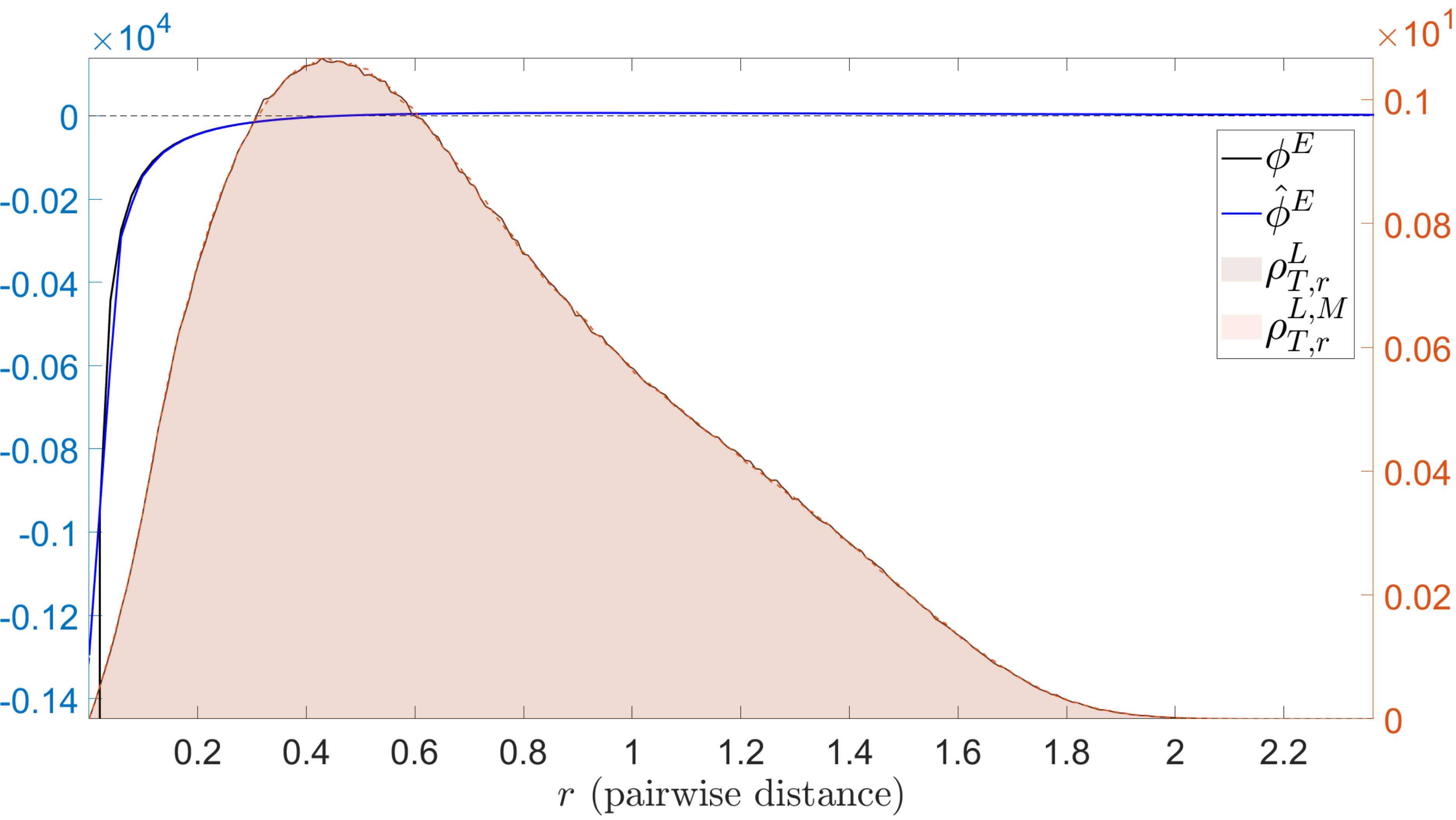}
\end{subfigure}
\caption{(FM2D) Comparison of $\intkernele$ and $\lintkernele$, with the relative error being $6.0 \cdot 10^{-2} \pm 2 \cdot 10^{-3}$ \rev{ (calculated using \eqref{eq:L2rhoTE})}. The true interaction kernel is shown in black solid line, whereas the mean estimated interaction kernel is shown in blue solid line with its confidence interval shown in blue dotted lines.  
Shown in the background is the comparison of approximated $\rhoTL$ versus the empirical $\rhoTLM$.}
\label{fig:FM2D_phiEs}
\end{figure}
\begin{figure}[H]
\begin{subfigure}{\textwidth}
  \centering
  \includegraphics[width=0.7\textwidth]{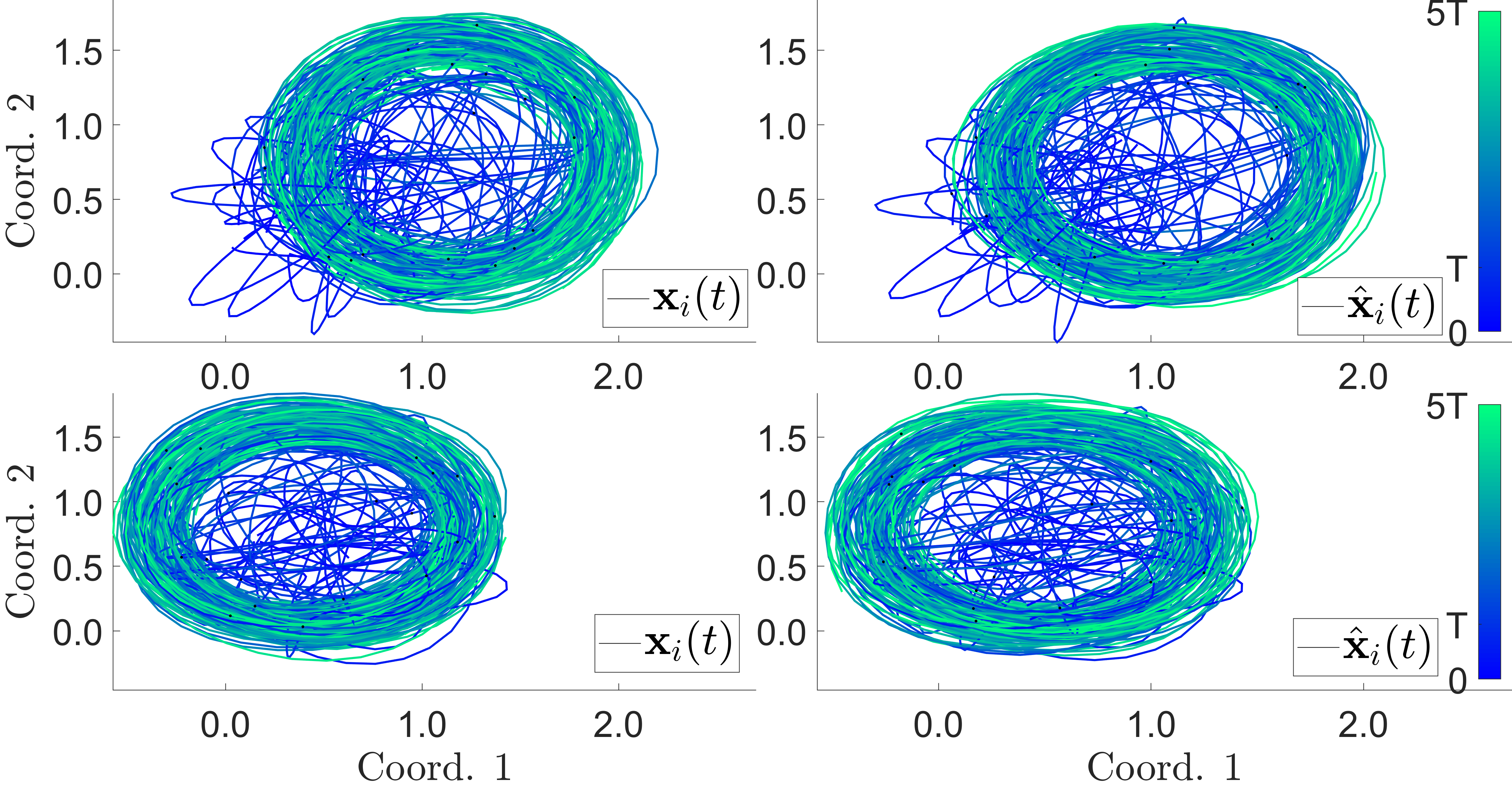} 
\end{subfigure}
\caption{(FM2D) Comparison of $\bX$ and $\hat\bX$, with the errors reported in table \ref{tab:FM2D_traj_err}. The first row of trajectories are generated from an initial condition taken from the observation data. The second row of trajectories are generated from another randomly chosen initial condition. The first column of trajectories are generated from the true interaction kernel, whereas the second column of trajectories are generated from our estimated kernel with the same initial conditions. \rev{The color of the trajectory indicates the flow of time, from deep blue (at $t = T_0$) to light green (at $t = T_f$).}}
\label{fig:FM2D_trajs}
\end{figure}

We consider the following interaction law,
\[
\intkernele(r) = \frac{N}{r}\big(\mathrm{e}^{-\frac{r}{2}} - 2\mathrm{e}^{-\frac{r}{0.5}}\big),
\]
With this setup, a double-milling pattern appears $100\%$ of the time (see \cite{Chuang2016}).  The other parameters are reported in table \ref{tab:FM2D_params}.  We use piecewise constant polynomials with $n^{\bx} = 122$ basis functions to approximate $\intkernele$.  The comparison of the true $\intkernela$ and the estimated $\lintkernela$ is shown in Fig.\ref{fig:FM2D_phiEs}.
As it is shown in Fig. \ref{fig:FM2D_phiEs}, our estimator closely resembles $\intkernele$, however when $r$ is close to $0$, there is a sharp drop of $\intkernele$ to $-\infty$, the availability of $r$ data close to $0$ becomes scarcer, and since we are using a uniform basis to approximate $\intkernele$, the difference between $\intkernele$ and $\lintkernele$ is apparent in this range.  The comparison of the true trajectory $\bX(t)$ and learned $\hat\bX(t)$ is shown in Fig. \ref{fig:FM2D_trajs}.
Our predicted system can still estimate the position/velocity of the agents in large time, i.e., for $t \gg T$, with relatively small error, around $10^{-2}$.  Moreover, when the dynamics enters its milling state, our predicted system is also in the same milling state.  We provide a quantitative insight into the difference between trajectories in table \ref{tab:FM2D_traj_err}.
\begin{table}[H]
\centering
\small{\begin{tabular}{| c || c | c |} 
\hline
                                                 & $[0, T]$                                 & $[T, T_f]$\\
\hline
$\text{mean}_{\text{IC}}$: Training ICs on $\bx$ & $2.35 \cdot 10^{-1} \pm 8 \cdot 10^{-3}$ & $8.38 \cdot 10^{-1} \pm 9 \cdot 10^{-3}$\\
\hline
$\text{mean}_{\text{IC}}$: Training ICs on $\bv$ & $3.6 \cdot 10^{-1} \pm 1 \cdot 10^{-2}$  & $1.13 \pm 1 \cdot 10^{-2}$\\
\hline
$\text{std}_{\text{IC}}$:  Training ICs on $\bx$ & $1.18 \cdot 10^{-1} \pm 7 \cdot 10^{-3}$ & $2.59 \cdot 10^{-1} \pm 9 \cdot 10^{-3}$\\
\hline     
$\text{std}_{\text{IC}}$:  Training ICs on $\bv$ & $1.64 \cdot 10^{-1} \pm 6 \cdot 10^{-3}$ & $3.1 \cdot 10^{-1} \pm 1 \cdot 10^{-2}$\\
\hline       
\hline
$\text{mean}_{\text{IC}}$: Random ICs   on $\bx$ & $2.34 \cdot 10^{-1} \pm 8 \cdot 10^{-3}$ & $8.35 \cdot 10^{-1} \pm 9 \cdot 10^{-3}$\\
\hline
$\text{mean}_{\text{IC}}$: Random ICs   on $\bv$ & $3.6 \cdot 10^{-1} \pm 1 \cdot 10^{-2}$  & $1.12 \pm 1 \cdot 10^{-2}$\\
\hline
$\text{std}_{\text{IC}}$:  Random ICs   on $\bx$ & $1.15 \cdot 10^{-1} \pm 5 \cdot 10^{-3}$ & $2.5 \cdot 10^{-1} \pm 1 \cdot 10^{-2}$\\
\hline   
$\text{std}_{\text{IC}}$:  Random ICs   on $\bv$ & $1.63 \cdot 10^{-1} \pm 8 \cdot 10^{-3}$ & $3.1 \cdot 10^{-1} \pm 1 \cdot 10^{-2}$\\
\hline
\end{tabular}}
\caption{(FM2D) Trajectory Errors: Initial Conditions (ICs) used in the training set (first two rows), new ICs randomly drawn from $\muX$ (second set of two rows).  \rev{The trajectory errors in $\bx$/$\bv$ is calculated using \eqref{eq:traj_norm_x}/\eqref{eq:traj_norm_v}.}}
\label{tab:FM2D_traj_err}
\end{table}
We are getting $10^{-1}$ relative accuracy for estimating positions/velocities within the learning time interval (i.e. $[0, T]$); however, as time goes on, we can see roughly linear growth of the errors ($T_f = 5T$). 

We consider the center of mass position $\bx_{\text{CM}}(t) = \frac{\sum_{i = 1}^N m_i\bx_i(t)}{\sum_{i = 1}^N m_i}$.  In the case of $m_i = 1$ for FM2D, it becomes, $\bx_{\text{CM}}(t) = \frac{1}{N}\bx_i(t)$.  We consider the indicator score $I_s$ (at $t = T_f$), $I_s = I_{\text{flock}} - I_{\text{mill}}$, where $I_{\text{flock}}, I_{\text{mill}}$ are taken from \cite{Chuang2007}.  Here, the flocking score $I_{\text{flock}}$ is defined as,
\[
I_{\text{flock}} = \norm{\frac{\sum_{i = 1}^N \bv_i(T_f)}{\sum_{i = 1}^N\norm{\bv_i(T_f)}}}.
\]
Again $I_{\text{flock}} = 1$ when perfect flocking occurs.  Then we consider the the milling score $I_{\text{mill}}$ as follows,
\[
I_{\text{mill}} = \norm{\frac{\sum_{i = 1}^N\norm{(\bx_i(T_f) - \bx_{\text{CM}}(T_f))\times\bv_i(T_f)}}{\sum_{i = 1}^N\norm{\bx_i(T_f) - \bx_{\text{CM}}(T_f)}\norm{\bv_i(T_f)}}}.
\]
When $I_{\text{mill}} = 1$ when perfect milling\footnote{The reason why we choose to use $\norm{(\bx_i(T_f) - \bx_{\text{CM}}(T_f))\times\bv_i(T_f)}$ is because it covers the case of double milling: half of the agents are rotating around the same axis, and the other half rotate around the same axis but in the exact opposite direction.} (around the same axis) occurs; meanwhile $I_{\text{flock}} = 0$ if $I_{\text{mill}} = 1$.  Therefore $I_s \in [-1, 1]$.  As suggested by the thresholds in \cite{Chuang2016}, we use the case, $I_s \le -0.5$, to indicate milling.  The true systems always show milling pattern (in fact, it shows double milling), and $100\%$ of our estimated systems also show milling.

Next for the pattern indicator scores, let $\text{PI}_1$ be the relative error of $I_s$ over $M$ trials.  And $\text{PI}_2$ is the relative error between the pair $(I_{\text{flock}}, I_{\text{mill}})$ (in $\ell_2$ norm) over $M$ trials.  The scores are reported in table \ref{tab:FM2D_PIs}. 
\begin{table}[H]
\centering
\small{\begin{tabular}{| c || c | c |} 
\hline
                                        & $\text{PI}_1$                            & $\text{PI}_2$\\
\hline
$\text{mean}_{\text{IC}}$: Training ICs & $2.47 \cdot 10^{-2} \pm 6 \cdot 10^{-4}$ & $2.21 \cdot 10^{-2} \pm 5 \cdot 10^{-4}$\\
\hline
$\text{std}_{\text{IC}}$:  Training ICs & $2.2 \cdot 10^{-2} \pm 2 \cdot 10^{-3}$  & $1.8 \cdot 10^{-2} \pm 1 \cdot 10^{-3}$\\
\hline        
\hline    
$\text{mean}_{\text{IC}}$: Random ICs   & $2.49 \cdot 10^{-2} \pm 3 \cdot 10^{-4}$ & $2.22 \cdot 10^{-2} \pm 3 \cdot 10^{-4}$\\
\hline
$\text{std}_{\text{IC}}$:  Random ICs   & $2.26 \cdot 10^{-2} \pm 6 \cdot 10^{-4}$ & $1.83 \cdot 10^{-2} \pm 6 \cdot 10^{-4}$\\
\hline   
\end{tabular}}
\caption{(FM2D) Pattern Indicator Scores: ICs used in the training set (first two rows), new ICs randomly drawn from $\muX$ (second set of two rows).}
\label{tab:FM2D_PIs}
\end{table}
Milling patterns in dynamics are very delicate.  The intricate balance $\alpha/\beta$ and the $H$-stability of $U_i$ decides the appearance of milling in a dynamics, especially when $U_i$ is not $H$-stable for double milling patterns.  In the case of the true dynamics showing milling (to be exact, double milling), our predicted systems can capture the same behavior (with high accuracy) both in terms of $I_s$ and the pair, $(I_{\text{flock}}, I_{\text{mill}})$.
\subsection{Fish Milling in $3$ dimensions}\label{sec:examples_FM3D}
Next, we consider a cohesive collective dynamics in $3D$ of self-propelled particles within a fluid environment, introduced in \cite{Chuang2016}.  It is a more complicated $3D$ extension of the FM2D model, where agents could experience self-propelling force in a fluid.

The governing equations of the Fish Milling Dynamics in $\R^3$ (FM3D) are,
\[
\ddot\bx_i = -\gamma(\dot\bx_i - \bu(\bx_i)) + \force_M(\dot\bx_i, \bu(\bx_i)) - \nabla_{\bx_i}U_i, \quad \text{for $i = 1, \cdots N$}.
\]
Here, $\bu$ is the lab-frame fluid velocity generated at position $\bx_i$, $-\gamma(\dot\bx_i - \bu(\bx_i))$ gives the drag force ($\gamma > 0$), $\force_M(\dot\bx_i, \bu(\bx_i))$ represents the self-propelling motility force, and $-\nabla_{\bx_i}U_i$ is the agent-to-agent interaction force on agent $i$, and the energy potential $U_i$ is the same Morse potential defined in sec. \ref{sec:examples_FM2D}.  $\force_M$ is defined as follows
\[
\force_M(\dot\bx_i, \bu(\bx_i)) = (\alpha - \beta\norm{\dot\bx_i - \lambda\bu(\bx_i)}^2)(\dot\bx_i - \lambda\bu(\bx_i)).
\]
The parameters, $\alpha, \beta > 0$, give the self-acceleration and deceleration, respectively; $0 \le \lambda \le 1$ is a perception coefficient, with $\lambda = 0$ showing a ``clear'' fluid (and it gives the classical Rayleigh-Helmholtz friction), and $\lambda = 1$ for an ``opaque'' fluid; and the lab-frame fluid velocity $\bu$ is given as follows
\[
\bu(\bx_i) = \sum_{i'=1}^N \frac{Gv_{i'}}{r_{i, i'}^2}\big[3\inprod{\hat\br_{i, i'}}{\hat{\dot\bx}_i}^2 - 1\big]\hat\br_{i, i'}.
\]
Here, $v_i = \norm{\dot\bx_i}$, $\hat{\dot\bx}_i = \frac{\dot\bx_i}{v_i}$, $\br_{i, i'} = \bx_{i'} - \bx_i$, $r_{i, i'} = \norm{\br_{i, i'}}$, $\hat\br_{i, i'} = \frac{\br_{i, i'}}{r_{i, i'}}$, and $\inprod{\cdot}{\cdot}$ is the normal inner product on $\R^3$. 

Table \ref{tab:FM3D_notation} shows how the FM3D dynamics fits into the framework of \eqref{eq:2ndOrder}.
\begin{table}[H]
\centering
\renewcommand{\arraystretch}{1}
\small{\begin{tabular}{c | c | c | c | c | c | c} 
\hline
Category & $\xi_i$     & $\numcl$ & $\forcex(\bx_i, \dot\bx_i)$                                          & $\intkernela$ & $s^{\bx}_{i, i'}$ & $\intkernele(r)$\\
\hline
Value    & $\emptyset$ & $1$      & $-\gamma(\dot\bx_i - \bu(\bx_i)) + \force_M(\dot\bx_i, \bu(\bx_i))$  & $\emptyset$  & $\emptyset$       & $N \cdot \big(\frac{C_a}{r\ell_a}\mathrm{e}^{-\frac{r}{\ell_a}} - \frac{C_r}{r\ell_r}\mathrm{e}^{-\frac{r}{\ell_{r}}}\big)$\\
\hline
\end{tabular}}
\renewcommand{\arraystretch}{1}
\caption{(FM3D) Mapping to \eqref{eq:2ndOrder}}
\label{tab:FM3D_notation}
\vspace{0.3cm}
\renewcommand{\arraystretch}{1}
\small{\begin{tabular}{ c | c | c | c | c | c | c | c | c | c | c | c}
\hline
$M$    & $d$ & $\alpha$  & $\beta$            & $G$       & $\lambda$ & $\gamma$  & $T_f$ & $T$ & $\muX$                            & $\muV$\\
\hline
$500$  & $3$ & $10^{-4}$ & $\frac{\alpha}{3}$ & $10^{-4}$ & $1.0$     & $10^{-4}$ & $20$  & $4$ & Unif. on $[0, 2.8\sqrt[3]{3}]^3$  & Unif. on $[0, 0]^3$\\
\hline
\end{tabular}}  
\renewcommand{\arraystretch}{1}
\caption{(FM3D) Parameters for Experiment Setup}
\label{tab:FM3D_params} 
\end{table}

The delicate balance between the self-propelling force (in the presence of a fluid environment) and the collective force induced by the energy kernel $U_i$ can create a wide range of patterns for such dynamics.  And the $H$-stability of $U_i$ is the key at producing milling patterns.  Again, we want to understand the milling pattern and predict such a pattern with our estimators when the true system shows milling.

\begin{figure}[H]
\begin{subfigure}{\textwidth}
  \centering
  \includegraphics[width=0.7\textwidth]{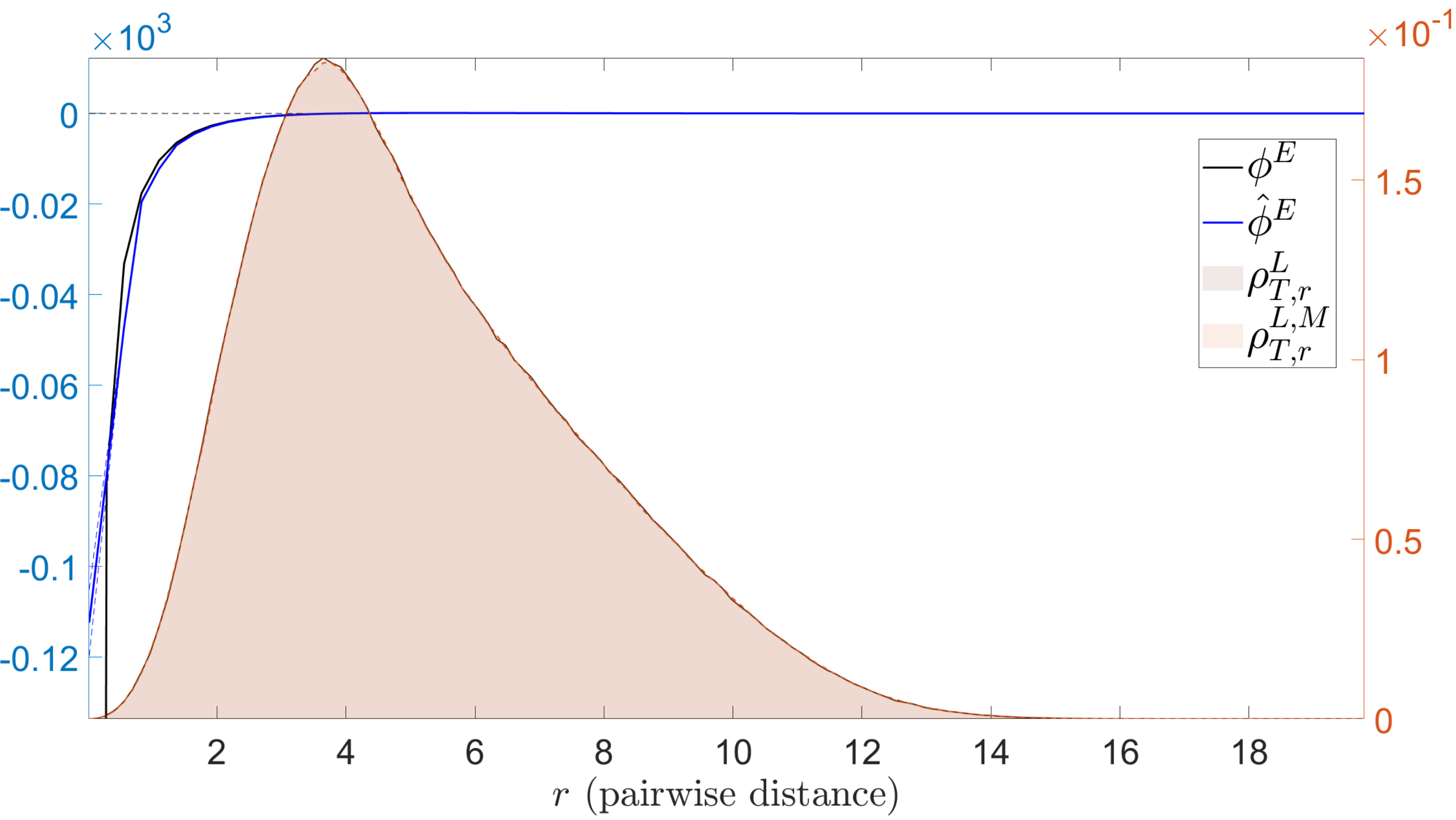}
\end{subfigure}
\caption{(FM3D) Comparison of $\intkernele$ and $\lintkernele$, with the relative error being $1.49 \cdot 10^{-1} \pm 3.4 \cdot 10^{-3}$ \rev{ (calculated using \eqref{eq:L2rhoTE})}. The true interaction kernel is shown in black solid line, whereas the mean estimated interaction kernel is shown in blue solid line with its confidence interval shown in blue dotted lines.  Shown in the background is the comparison of approximated $\rhoTL$ versus the empirical $\rhoTLM$.}
\label{fig:FM3D_phiEs}
\end{figure}

\begin{figure}[H]
\begin{subfigure}{\textwidth}
  \centering
  \includegraphics[width=0.7\textwidth]{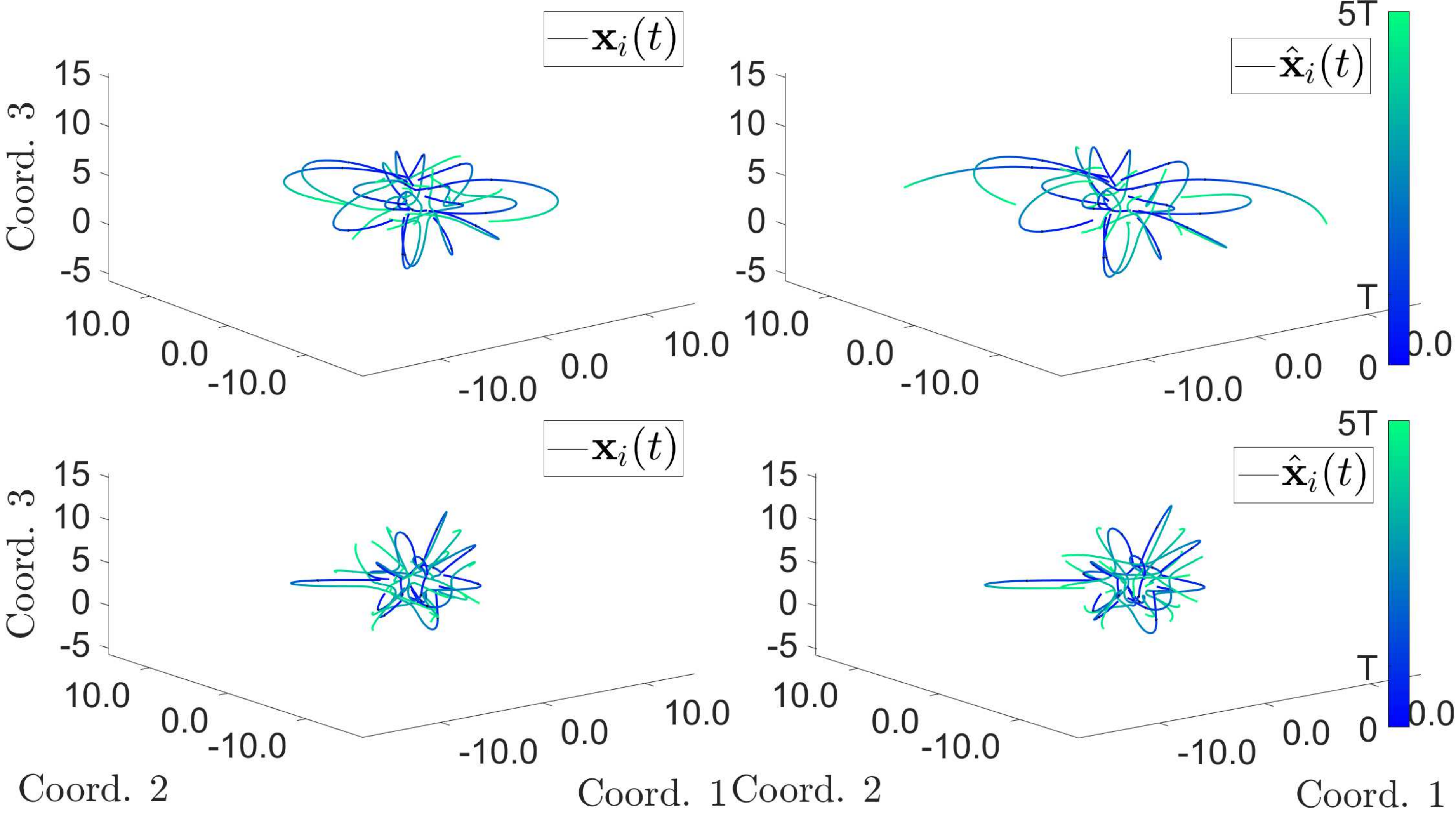} 
\end{subfigure}
\caption{(FM3D) Comparison of $\bX$ and $\hat\bX$, with the errors reported in table \ref{tab:FM3D_traj_err}. The first row of trajectories are generated from an initial condition taken from the observation data. The second row of trajectories are generated from another randomly chosen initial condition. The first column of trajectories are generated from the true interaction kernel, whereas the second column of trajectories are generated from our estimated kernel with the same initial conditions. \rev{The color of the trajectory indicates the flow of time, from deep blue (at $t = T_0$) to light green (at $t = T_f$).}}
\label{fig:FM3D_trajs}
\end{figure}

We consider the following interaction law,
\[
\intkernele(r) = N \cdot \big(\frac{1.4}{2.8r}\mathrm{e}^{-\frac{2.8r}{1.4}} - \frac{2}{r}\mathrm{e}^{-r}\big).
\]
We also use the parameters in table \ref{tab:FM3D_params} to set up the experiment.  Piece-wise linear polynomials with $n^{\bx} = 74$ basis functions are used to approximate $\intkernele$.  The comparison of the true $\intkernela$ and the estimated $\lintkernela$ is shown in Fig.\ref{fig:FM3D_phiEs}.
As it is shown in Fig. \ref{fig:FM3D_phiEs}, our estimator, $\lintkernele$, deviates from $\intkernele$ for $r$ close to $0$, for similar reasons as those discussed for the $2D$ case.  The comparison of the true trajectory $\bX(t)$ and learned $\hat\bX(t)$ is shown in Fig. \ref{fig:FM3D_trajs}.
\rev{A $3D$ milling pattern is more complicated than its $2D$ counterpart.  In the case of our experiments, some of the trajectories show a pattern of rotation about a fixed point.  And our estimated dynamics also shows similar behavior.} We provide a quantitative insight into the difference between trajectories in table \ref{tab:FM3D_traj_err}.
\begin{table}[H]
\centering
\small{\begin{tabular}{| c || c | c |} 
\hline
                                                 & $[0, T]$                                & $[T, T_f]$\\
\hline
$\text{mean}_{\text{IC}}$: Training ICs on $\bx$ & $4.9 \cdot 10^{-2} \pm 4 \cdot 10^{-3}$ & $6.8 \cdot 10^{-1} \pm 4 \cdot 10^{-2}$\\
\hline
$\text{mean}_{\text{IC}}$: Training ICs on $\bv$ & $1.6 \cdot 10^{-1} \pm 1 \cdot 10^{-2}$ & $1.2 \pm 3 \cdot 10^{-1}$\\
\hline
$\text{std}_{\text{IC}}$:  Training ICs on $\bx$ & $5.0 \cdot 10^{-3} \pm 3 \cdot 10^{-4}$ & $2.3 \cdot 10^{-1} \pm 3 \cdot 10^{-2}$\\
\hline     
$\text{std}_{\text{IC}}$:  Training ICs on $\bv$ & $3.3 \cdot 10^{-2} \pm 2 \cdot 10^{-3}$ & $4 \cdot 10^{-1} \pm 4 \cdot 10^{-1}$\\
\hline       
\hline
$\text{mean}_{\text{IC}}$: Random ICs   on $\bx$ & $4.9 \cdot 10^{-2} \pm 4 \cdot 10^{-3}$ & $6.8 \cdot 10^{-1} \pm 4 \cdot 10^{-2}$\\
\hline
$\text{mean}_{\text{IC}}$: Random ICs   on $\bv$ & $1.5 \cdot 10^{-1} \pm 1 \cdot 10^{-2}$ & $1.2 \pm 3 \cdot 10^{-1}$\\
\hline
$\text{std}_{\text{IC}}$:  Random ICs   on $\bx$ & $5.1 \cdot 10^{-3} \pm 3 \cdot 10^{-4}$ & $2.3 \cdot 10^{-1} \pm 4 \cdot 10^{-2}$\\
\hline   
$\text{std}_{\text{IC}}$:  Random ICs   on $\bv$ & $3.2 \cdot 10^{-2} \pm 1 \cdot 10^{-3}$ & $4 \cdot 10^{-1} \pm 5 \cdot 10^{-1}$\\
\hline
\end{tabular}}
\caption{(FM3D) Trajectory Errors: Initial Conditions (ICs) used in the training set (first two rows), new ICs randomly drawn from $\muX$ (second set of two rows). \rev{The trajectory errors in $\bx$/$\bv$ is calculated using \eqref{eq:traj_norm_x}/\eqref{eq:traj_norm_v}.}}
\label{tab:FM3D_traj_err}
\end{table}
We consider the center of mass velocity $\bv_{\text{CM}}(t) = \frac{\sum_{i = 1}^N m_i\bv_i(t)}{\sum_{i = 1}^N m_i}$.  In the case of $m_i = 1$ for FM3D, it becomes, $\bv_{\text{CM}}(t) = \frac{1}{N}\bv_i(t)$.  We consider the indicator score $I_s$ (at $t = T_f$) from \cite{Chuang2016}, 
\[
I_s = I_{\text{flock}} - I_{\text{mill}}.
\]
Here, the flocking score $I_{\text{flock}}$ is defined as,
\[
I_{\text{flock}} = 1 - \frac{\sum_{i = 1}^N \norm{\bv_i(T_f) - \bv_{\text{CM}}(T_f)}}{N\sqrt{\frac{\alpha}{\beta}}}.
\]
Again $I_{\text{flock}} = 1$ when perfect flocking occurs.  The milling score $I_{\text{mill}}$ has two pieces: we first define the rotational axis $\omega_i$ for agent $i$,
\[
\omega_i = \frac{\bv_i(T_f) \times (m_i\dot\bv_i(T_f))}{\norm{\bv_i(T_f)}\norm{m_i\dot\bv_i(T_f)}};
\]
next,
\[
I_{\text{mill}} = \frac{\sum_{i = 1}^N\sum_{j = 1, j \neq i}^N \inprod{\omega_i}{\omega_j}}{N(N - 1)}.
\]
When $I_{\text{mill}} = 1$ when perfect milling (around the same axis) occurs; meanwhile $I_{\text{flock}} = 0$ if $I_{\text{mill}} = 1$.  Therefore $I_s \in [-1, 1]$.  As suggested by the thresholds in \cite{Chuang2016}, we use the case, $I_s \le -0.5$, to indicate milling, see the results in table \ref{tab:FM3D_CM}.
\begin{table}[H]
\centering
\small{\begin{tabular}{| c || c | c |} 
\hline
                                & Predicted Non-Milling & Predicted Milling\\
\hline
True Non-Milling: Training ICs  & $99 \pm 2 \%$         & $1 \pm 2\%$ \\
\hline
True Milling:     Training ICs  & $0\%$                 & $0\%$\\
\hline   
\hline
True Non-Milling: Random ICs    & $99 \pm 3 \%$         & $1 \pm 3 \%$ \\
\hline
True Milling:     Random ICs    & $0   \%$              & $0 \%$\\
\hline 
\end{tabular}}
\caption{(FM3D) Confusion Matrix: ICs used in the training set (first two rows), new ICs randomly drawn from $\muX$ (second set of two rows). \rev{It is generated using table \ref{tab:general_CM}.}}
\label{tab:FM3D_CM}
\end{table}
\rev{The true FM3D systems show a non-milling\footnote{\rev{The milling score given by \cite{Chuang2016} fails to capture the case of milling around a fixed point, hence the non-milling pattern.}} pattern.  Furthermore, our predicted systems show a exceptionally similar probability of displaying the non-milling patterns.}  

We provide more statistics about the confusion matrix in order to understand our prediction of milling behavior better in table \ref{tab:FM3D_CM_details}.
\begin{table}[H]
\centering
\small{\begin{tabular}{| c || c | c | c | c |} 
\hline
             & Accuracy      & Precision      & Recall   & $F$-Score\\
\hline
Training ICs & $99 \pm 2 \%$ & $30 \pm 48\%$  & $100 \%$ & $30 \pm 48\%$ \\
\hline
\hline
Random ICs   & $99 \pm 3 \%$ & $70 \pm 48\%$  & $100 \%$ & $70 \pm 48\%$ \\
\hline
\end{tabular}}
\caption{(FM3D) Confusion Matrix Statistics: ICs used in the training set, new ICs randomly drawn from $\muX$. \rev{It is generated using table \ref{tab:general_CM_stat}.}}
\label{tab:FM3D_CM_details}
\end{table} 
Next, we use the pattern indicator scores to probe deeper into the actual large-time behavior of our predicted systems.  $\text{PI}_1$ is the relative error of $I_s$.  And $\text{PI}_2$ is the relative error of predicting the pair $(I_{\text{flock}}, I_{\text{mill}})$ (in $\ell_2$ norm).  The scores are reported in table \ref{tab:FM3D_PIs}.
\begin{table}[H]
\centering
\small{\begin{tabular}{| c || c | c |} 
\hline
                                        & $\text{PI}_1$                         & $\text{PI}_2$\\
\hline
$\text{mean}_{\text{IC}}$: Training ICs & $3 \cdot 10^{-1} \pm 1 \cdot 10^{-1}$ & $3 \cdot 10^{-1} \pm 1 \cdot 10^{-1}$\\
\hline
$\text{std}_{\text{IC}}$:  Training ICs & $3 \cdot 10^{-1} \pm 2 \cdot 10^{-1}$ & $3 \cdot 10^{-1} \pm 3 \cdot 10^{-1}$\\
\hline            
\hline
$\text{mean}_{\text{IC}}$: Random ICs   & $3 \cdot 10^{-1} \pm 1 \cdot 10^{-1}$ & $3 \cdot 10^{-1} \pm 1 \cdot 10^{-1}$\\
\hline
$\text{std}_{\text{IC}}$:  Random ICs   & $3 \cdot 10^{-1} \pm 3 \cdot 10^{-1}$ & $3 \cdot 10^{-1} \pm 3 \cdot 10^{-1}$\\
\hline   
\end{tabular}}
\caption{(FM3D) Pattern Indicator Scores: ICs used in the training set (first two rows), new ICs randomly drawn from $\muX$ (second set of two rows).}
\label{tab:FM3D_PIs}
\end{table}
\rev{Not only can we predict $I_s$ with relatively high accuracy, but also can we offer insight into the actual pair of scores, $(I_{\text{flock}}, I_{\text{mill}})$.}
\section{Emergent Behaviors Induced by $\intkernel(r, s)$}\label{sec:examples_SOD}
The flexibility of the learning algorithm given in \cite{PNASLU} allows for a generalization of the dynamical system where the interaction kernels can depend on more than just one variable, i.e., more than just $r$ (the pairwise distance data).  For example, in modeling the movement of groups of animals, field of vision can affect how individuals influence each other;  in synchronized fireflies, not only can the fireflies form spatial patterns, their light-emitting states can be also be locked in synchronization.  We consider an important example in \cite{OKeeffe2017} which models how oscillators can sync and swarm together, hence the interaction kernels depend on both $r$ and $\xi$ (the pairwise difference in phases).  Further study of this type of model has been done in \cite{Gupta2018,Levis2019,Kruk2018,OKeeffe2018}, a review with applications to computation is given in \cite{Okeeffe2019}, and a historical review of the development of the synchronization models can be found in \cite{Strogatz2000}.

These authors sought to develop a plausible model that could explain systems where a phase or real-valued feature affects the motion -- and vice versa. They called such systems ``swarmalators'' to emphasize the combined behavior of swarming and synchronized oscillation of phases in the system. 

\begin{figure}[H]
\centering
\begin{subfigure}{.48\textwidth}
  \includegraphics[width=.95\linewidth]{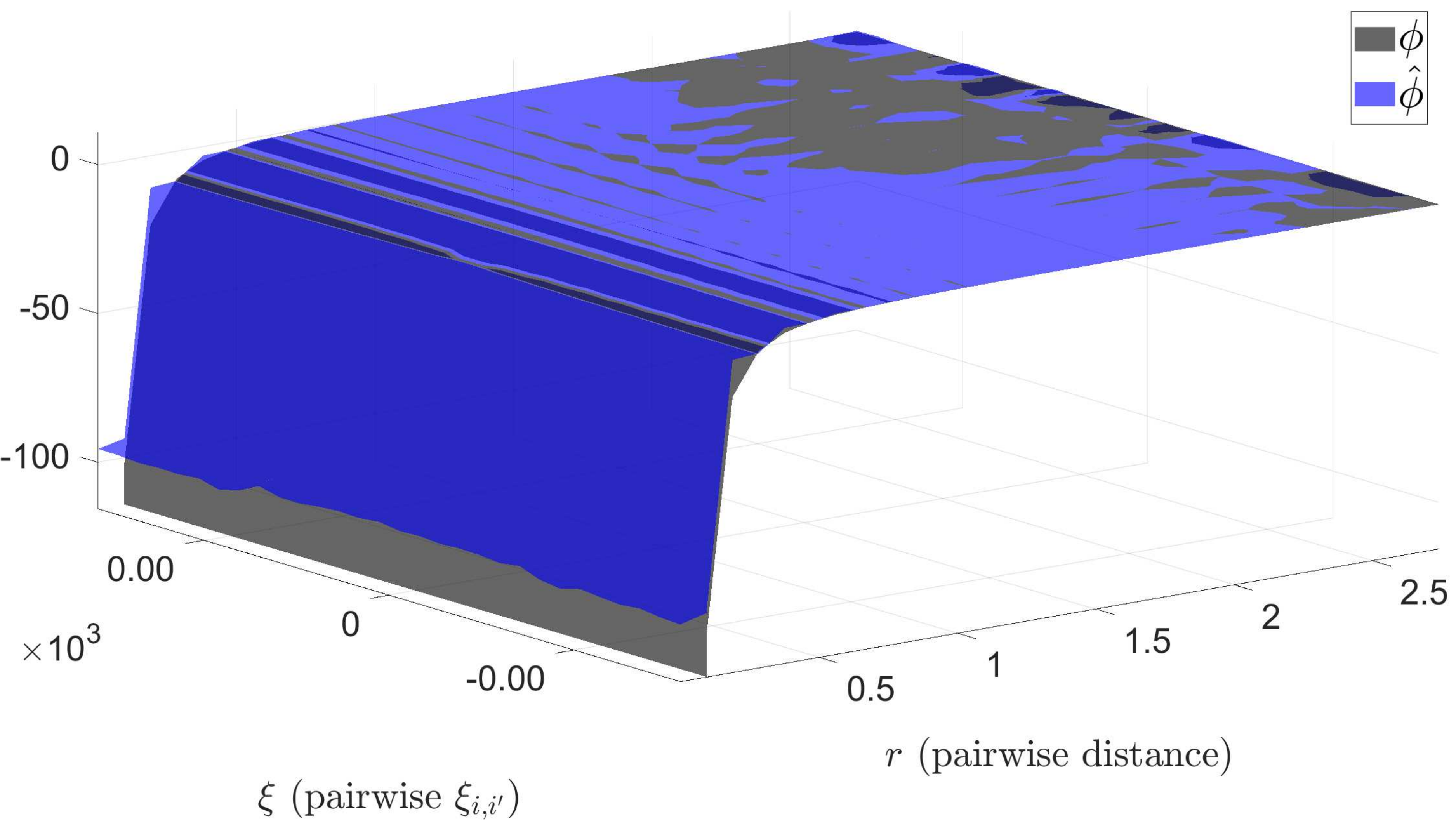}  
  \caption{Comparison of $\intkernele$ and $\lintkernele$, the relative error is $4.5 \cdot 10^{-1} \pm 9 \cdot 10^{-2}$ \rev{ (calculated using \eqref{eq:L2rhoTE})}. }
  \label{fig:SOD_phiEs}
\end{subfigure}
~
\begin{subfigure}{.48\textwidth}
  \centering
  \includegraphics[width=.95\linewidth]{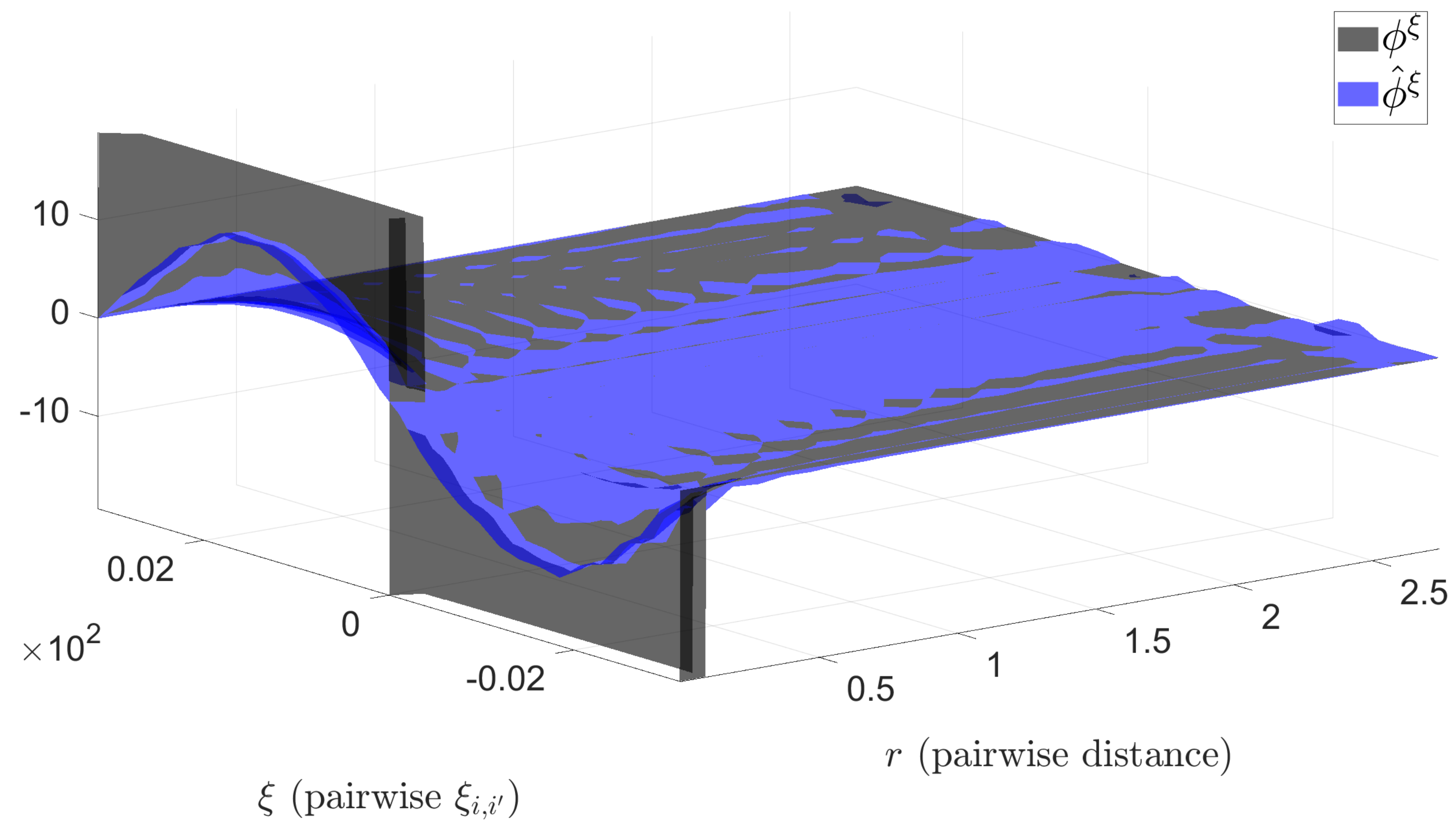}  
  \caption{Comparison of $\intkernelxi$ and $\lintkernelxi$, the relative error is $2 \cdot 10^{-1} \pm 1 \cdot 10^{-1}$ \rev{ (calculated using \eqref{eq:L2rhoTXi_1stOrder})}.}
  \label{fig:SOD_phiXis}
\end{subfigure}
\caption{(SOD) The true interaction laws are shown in black, and the mean estimated interaction laws are shown in blue.}
\label{fig:SOD_phis}
\end{figure}

\begin{figure}[H]
\begin{subfigure}{\textwidth}
  \centering
  \includegraphics[width=0.7\textwidth]{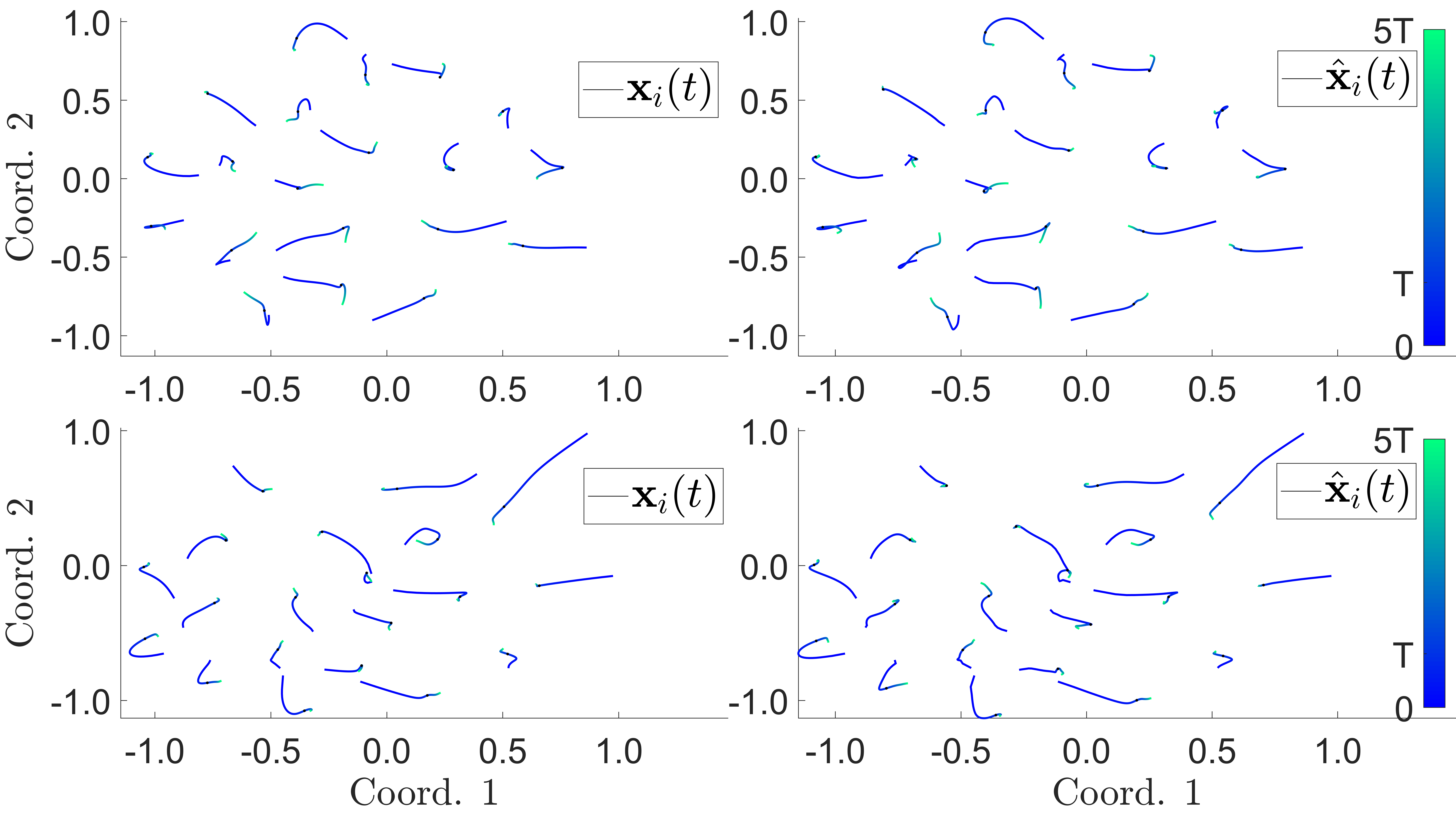} 
\end{subfigure}
\caption{(SOD) Comparison of $\bX$ and $\hat\bX$, with the errors reported in table \ref{tab:SOD_traj_err}. The first row of trajectories are generated from an initial condition taken from the observation data. The second row of trajectories are generated from another randomly chosen initial condition. The first column of trajectories are generated from the true interaction kernel, whereas the second column of trajectories are generated from our estimated kernel with the same initial conditions. \rev{The color of the trajectory indicates the flow of time, from deep blue (at $t = T_0$) to light green (at $t = T_f$).}}
\label{fig:SOD_trajs}
\end{figure}

For the Synchronized Oscillator Dynamics (SOD), each agent is indexed by $i$, $\xi_i$ is its phase, $\bx_i$ is (as usual) its position, $\omega_i$ is the fixed natural frequency, $\bv_i$ is the fixed self-propulsion velocity. The dynamics of $\bx_i$ and $\xi_i$ are governed by the following equations,
\begin{equation}\label{eq:swarmalators}
\begin{cases}
\dot\bx_i &= \bv_i + \frac{1}{N}\sum_{i'=1}^{N}\big(\frac{\bx_{i'} - \bx_i}{\norm{\bx_{i'} - \bx_i}}(A+J \cos(\xi_{i'} - \xi_i)) - B\frac{\bx_{i'} - \bx_i}{\norm{\bx_{i'} - \bx_i}^2} \big) \\
\dot\xi_i &= \omega_i + \frac{K}{N} \sum_{i'=1}^{N} \frac{\sin(\xi_{i'} - \xi_i)}{\norm{\bx_{i'} - \bx_i}}
\end{cases}
\end{equation}
Table \ref{tab:SOD_notation} shows how the SOD dynamics fits into the framework of \eqref{eq:1stOrder}.
\begin{table}[H]
\centering
\renewcommand{\arraystretch}{1}
\small{\begin{tabular}{c | c | c | c | c | c} 
\hline
Category & $\numcl$ & $\forcex(\bx_i, \xi_i)$ & $\forcexi(\bx_i, \xi_i)$ & $\intkernela(r, s^{\bx})$                     & $\intkernelxi(r, s^{\xi})$ \\
\hline
Value    & $1$      & $\bv_i$                 & $\omega_i$               & $\frac{A+J \cos(s^{\bx})}{r} - \frac{B}{r^2}$ & $\frac{K\sin(s^{\xi})}{r}$ \\
\hline
\end{tabular}}
\renewcommand{\arraystretch}{1}
\caption{Mapping of SOD to \eqref{eq:1stOrder}}
\label{tab:SOD_notation}
\vspace{0.3cm}
\renewcommand{\arraystretch}{1}
\small{\begin{tabular}{ c | c | c | c | c | c | c | c }
\hline
$M$    & $\bv_i$ & $\omega_i$ & $d$ & $T_f$  & $T$  & $\muX$               & $\muXi$\\
\hline
$1000$ & $0$     & $0$        & $4$ & $20$   & $4$ & Unif. on $[-1, 1]^2$  & Unif. on $[-\pi, \pi]$\\
\hline
\end{tabular}}  
\renewcommand{\arraystretch}{1}
\caption{(SOD) Parameters for Experiment Setup}
\label{tab:SOD_params} 
\end{table}
With certain choices of $A, J, B$ and $K$, the SOD dynamics is going to produce either a static or a non-static spatial pattern with either phases in sync or out of sync (a total of $5$ different states, see \cite{OKeeffe2017} for details). We consider the following interaction law,
\[
\intkernele(r, s^{\bx}) = \frac{1 + J\cos(s^{\bx})}{r} - \frac{1}{r^2} \mand \intkernelxi = \frac{K\sin(s^{\xi})}{r}.
\]
Here $K$ and $J$ are changing and we take $s^{\bx} = s^{\xi} = \xi$ (the pairwise difference in the phases, i.e., $\xi_{i'} - \xi_i$).   We consider a particular set of $(J, K)$ values, i.e. $(J, K) = (0.1, 1)$, which gives a static synchronous state.  In table \ref{tab:SOD_params} we describe the other parameters that we use to set up the experiment.  \rev{Here we use }piecewise linear polynomials with $n^{\bx} = 900$ (with $30$ basis functions in each dimension) and $n^{\xi} = 900$ (with $30$ basis functions in each dimension) basis functions to approximate $\intkernele$ and $\intkernelxi$.  The comparison of the true $\intkernela$ and the estimated $\lintkernela$ is shown in Fig.\ref{fig:SOD_phis}.
As is shown in Fig. \ref{fig:SOD_phiEs} and Fig. \ref{fig:SOD_phiXis}, even with the interaction laws being $2$-dimensional, we can still infer from the data with around $10^{-1}$ relative accuracy with relatively small number of basis functions.  A comparison of the true trajectory $\bX(t)$ and learned $\hat\bX(t)$ is shown in Fig. \ref{fig:SOD_trajs}.
A visual comparison of the trajectories between the true and estimated dynamics shows that the difference is small.  An more quantitative insight into the difference between trajectories is provided in table \ref{tab:SOD_traj_err}.
\begin{table}[H]
\centering
\small{\begin{tabular}{| c || c | c |} 
\hline
                                                 & $[0, T]$                                & $[T, T_f]$\\
\hline
$\text{mean}_{\text{IC}}$: Training ICs on $\bx$ & $4.69 \cdot 10^{-2} \pm 5 \cdot 10^{-4}$& $7.2 \cdot 10^{-2} \pm 1 \cdot 10^{-3}$\\
\hline
$\text{mean}_{\text{IC}}$: Training ICs on $\xi$ & $2.97 \cdot 10^{-2} \pm 5 \cdot 10^{-4}$& $3 \cdot 10^{-1} \pm 7 \cdot 10^{-1}$\\
\hline
$\text{std}_{\text{IC}}$:  Training ICs on $\bx$ & $9 \cdot 10^{-3} \pm 1 \cdot 10^{-3}$   & $3.5 \cdot 10^{-2} \pm 1 \cdot 10^{-3}$\\
\hline      
$\text{std}_{\text{IC}}$:  Training ICs on $\xi$ & $3.0 \cdot 10^{-2} \pm 2 \cdot 10^{-3}$ & $8 \pm 22$\\
\hline      
\hline 
$\text{mean}_{\text{IC}}$: Random ICs   on $\bx$ & $4.67 \cdot 10^{-2} \pm 6 \cdot 10^{-4}$& $7.2 \cdot 10^{-2} \pm 1 \cdot 10^{-3}$\\
\hline
$\text{mean}_{\text{IC}}$: Random ICs   on $\xi$ & $3.0 \cdot 10^{-2} \pm 1 \cdot 10^{-3}$ & $1.2 \cdot 10^{-1} \pm 8 \cdot 10^{-2}$\\
\hline
$\text{std}_{\text{IC}}$:  Random ICs   on $\bx$ & $9 \cdot 10^{-3} \pm 2 \cdot 10^{-3}$   & $3.5 \cdot 10^{-2} \pm 2 \cdot 10^{-3}$\\
\hline   
$\text{std}_{\text{IC}}$:  Random ICs   on $\xi$ & $2.9 \cdot 10^{-2} \pm 3 \cdot 10^{-3}$ & $2 \pm 3$\\
\hline
\end{tabular}}
\caption{(SOD) Trajectory Errors: Initial Conditions (ICs) used in the training set (first two rows), new ICs randomly drawn from $\muX$ (second set of two rows). \rev{The trajectory errors in $\bx$/$\xi$ is calculated using \eqref{eq:traj_norm_x}/\eqref{eq:traj_norm_xi}.}}
\label{tab:SOD_traj_err}
\end{table}
\rev{The Synchronized Oscillator dynamics is a complex dynamical system with $\bx_i$ and a periodic $\xi_i$ interacting with each other within the agents themselves and also collectively among the other agents; however we are still able to maintain $2$-digit relative accuracy in reproducing the trajectories, and in predicting the large-time behavior of the trajectories, the errors do not grow exponentially.  We are considering the static synchronous state, hence we check the distribution of the phases at time $T_f$ to see if the phases are in sync.  In particular, we use the mean and the variance of the phases at time $T_f$.  If the variance of $\{\xi_i(T_f)\}_{i = 1}^N$ is smaller than $0.01$, we say that the dynamics is in static synchronous state.  Since the true systems always show synchronous state, our estimated systems also shows synchronization of phases with the same certainty.}

Next, we use the following pattern indicator scores to discuss the quantitative predication performance of our estimators.  \rev{$\text{PI}_1$ is the relative error of the variance of the phases at time $T_f$, and $\text{PI}_2$ is the relative error of the mean of the phases at time $T_f$.  The scores are reported in table \ref{tab:SOD_PIs}.}
\begin{table}[H]
\centering
\small{\begin{tabular}{| c || c | c |} 
\hline
                                        & $\text{PI}_1$ & $\text{PI}_2$\\
\hline
$\text{mean}_{\text{IC}}$: Training ICs & $0$           & $8 \cdot 10^{-2} \pm 2 \cdot 10^{-1}$\\
\hline
$\text{std}_{\text{IC}}$:  Training ICs & $0$           & $2 \pm 5$\\
\hline         
\hline   
$\text{mean}_{\text{IC}}$: Random ICs   & $0$           & $3 \cdot 10^{-2} \pm 3 \cdot 10^{-2}$\\
\hline
$\text{std}_{\text{IC}}$:  Random ICs   & $0$           & $4 \cdot 10^{-1} \pm 6 \cdot 10^{-1}$\\
\hline   
\end{tabular}}
\caption{(SOD) Pattern Indicator Scores: ICs used in the training set (first two rows), new ICs randomly drawn from $\muX$ (second set of two rows).}
\label{tab:SOD_PIs}
\end{table}
\rev{Our estimated systems display exactly the same synchronization behavior as the true system (variance of the phases is exactly $0$).  Meanwhile, we can reproduce the final synchronized phase with relative high accuracy, however it comes with a big uncertainty.}
\section{Emergent Behaviors Induced by Parametric Families of Interaction Kernels}\label{sec:examples_GSS}
In recent years, there has been rapid growth in developing algorithms (either theoretical or numerical) to identify the governing equations of physical systems based on observed data.  A notable collection of these approaches assume a parametric form of the equations to perform various kinds of regression, usually sparse regression against an enormous library of standard mathematical functions, to fit the parameters \cite{BPK2016, RBPK2017, BKP2017}. Other approaches use force-based, statistical mechanics, and multiscale methods -- see the works  \cite{BCCCCGLOPPVZ2008,LLEK2010,KTIHC2011,BCGMSVW2012,Schaeffer6634}.  Our non-parametric learning approach can also be used to discover the elaborate structure of the true interaction law, i.e., $\intkernel(r) = \intkernel(r; \bP)$, where $\bP = \begin{bmatrix} p_1 & \cdots & p_k\end{bmatrix}$ is a vector of parameters.  In many settings, $\intkernel$ can be written as $\intkernel(r; \bP) = J(\bP)\intkernel_m(r)$, where $J(\cdot)$ might offer physical insight through its effect on the parameters.  In this paper, we will focus on the case when $\bP$ is $1$-dimensional, i.e., a family of one-parameter interaction laws.\\

\begin{figure}[H]
\begin{subfigure}{\textwidth}
  \centering
   \includegraphics[width=0.8\textwidth]{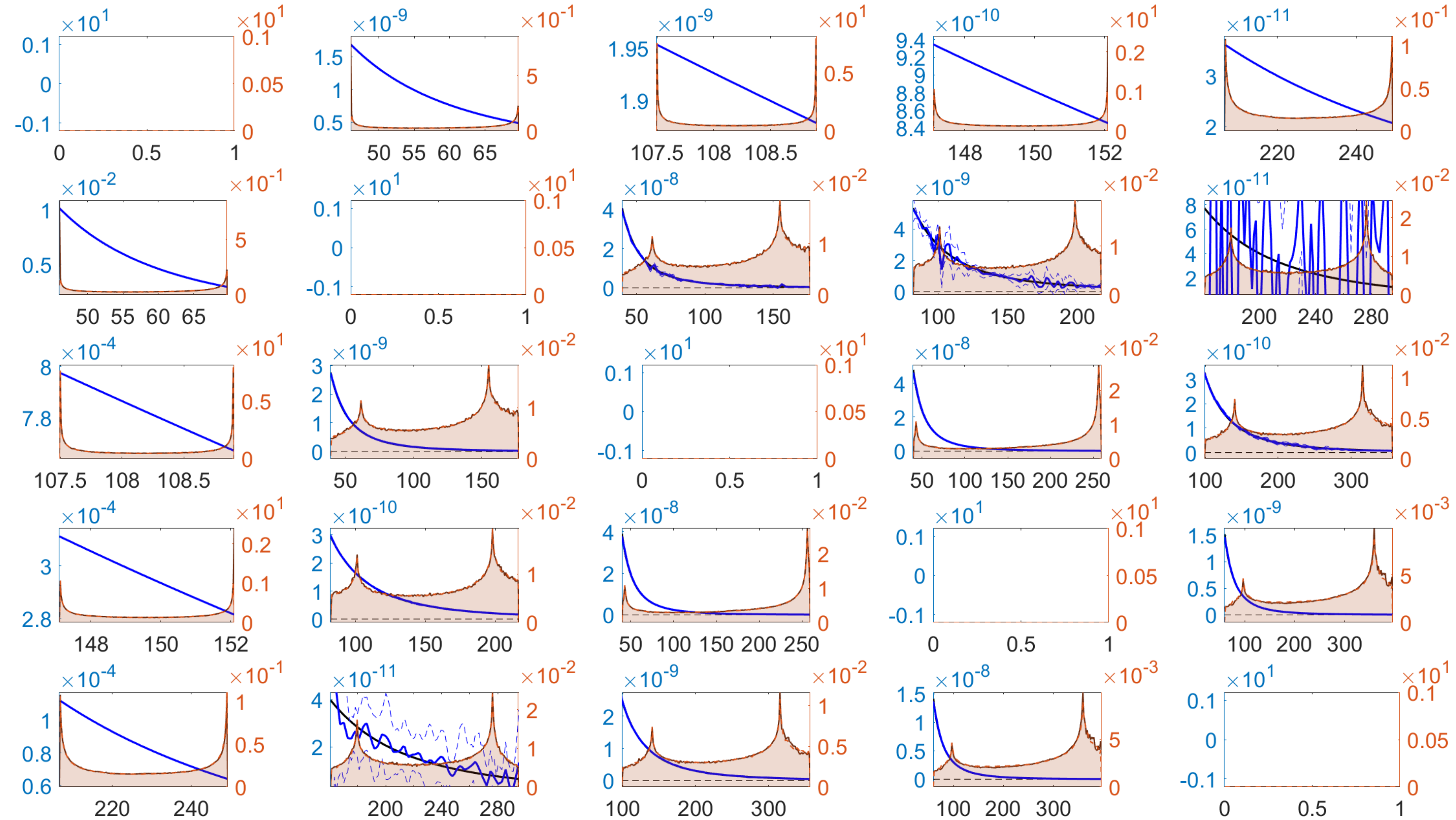}
\end{subfigure}
\caption{(GSS) Comparison of $\intkernele_{\idxcl, \idxcl'}$'s and $\lintkernele_{\idxcl, \idxcl'}$'s , the relative errors are reported in table \ref{tab:GSS_phi_err}.  For example, $\intkernele_{1, 2}$ on cell $(1, 2)$ of the plot represents the true force Mercury has on Sun.  Within each sub-plot, the true interaction kernel is shown in black solid line, whereas the mean estimated interaction kernel is shown in blue solid line with its confidence interval shown in blue dotted lines.  Shown in the background of each sub-plot is the comparison of approximated $\rho_{T, r}^{L, \idxcl, \idxcl'}$ (in lighter color) versus the empirical $\rho_{T, r}^{L, M, \idxcl, \idxcl'}$ (in darker color).}
\label{fig:GSS_phis}
\end{figure}

\begin{figure}[H]
\begin{subfigure}{\textwidth}
  \centering
   \includegraphics[width=0.8\textwidth]{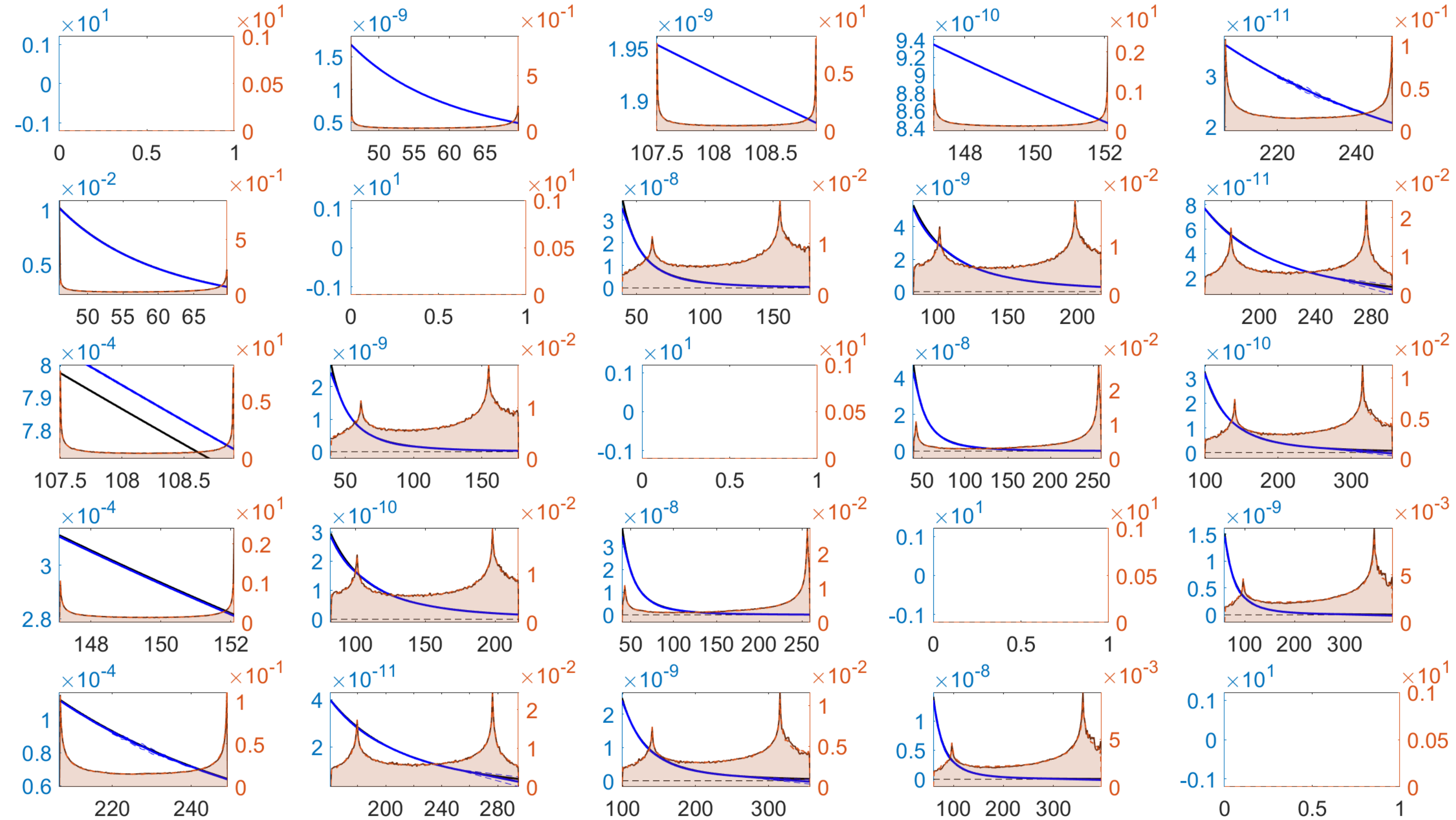}
\end{subfigure}
\caption{(GSS) Comparison of $\intkernele_{\idxcl, \idxcl'}$'s and cleaned-up $\lintkernele_{\idxcl, \idxcl'}$'s , the relative errors are reported in table \ref{tab:GSS_phi_err_clean}.  Similar layout and setup as in Fig. \ref{fig:GSS_phis}.}
\label{fig:GSS_phis_clean}
\end{figure}

We consider a simplified planetary movement in our solar system (GSS) as a second order collective dynamical system example with parametric interaction laws.  We take $\bx_i(t) \in \R^2$ or $\R^3$ as the position of each planet (only the planets in the inner-solar system are considered, i.e., Mercury, Venus, Earth and Mars, hence $N = 5$).  Their positions are governed by the following form of Newton's Law,
\begin{equation}\label{eq:solar_system}
\tilde{m}_{i}^{\mI}\ddot\bx_i(t) = \sum_{\substack{i' = 1 \\ i' \neq i}}^N \frac{G\tilde{m}_i^{\mG}\tilde{m}_{i'}}{\norm{\bx_{i'} - \bx_i}^2}\cdot\frac{\bx_{i'} - \bx_i}{\norm{\bx_{i'} - \bx_i}}, \quad \text{for $i = 1, \cdots, N$}.
\end{equation}
$\tilde{m}_i^{\mI}$ is the inertia mass of the $i^{th}$ astronomical object (AO), and $\tilde{m}_i^{\mG}$ is the gravitational mass of the corresponding AO.  In our setting we will assume that they are the same, hence \eqref{eq:solar_system} can be simplified to,
\begin{equation}\label{eq:solar_system_sim}
\ddot\bx_i(t) = \sum_{i'=1}^N \frac{G\tilde{m}_{i'}}{\norm{\bx_{i'} - \bx_i}^3}(\bx_{i'} - \bx_i), \quad \text{for $i = 1, \cdots, N$}.
\end{equation}
Here $\tilde{m}_i$ is the unknown mass of the $i^{th}$ AO, and $G = 6.67408 \cdot 10^{-11} \text{m}^3\text{kg}^{-1}\text{s}^{-2}$ is the gravitational constant (known to the algorithm). There are a total of $5$ different types of agents (each AO is of its own type) in this system, and the true interaction laws are
\[
\intkernele_{\idxcl, \idxcl'}(r; \tilde{m}_{\idxcl'}) = G\tilde{m}_{\idxcl'} \cdot \frac{1}{r^3}, \quad \text{for $\idxcl, \idxcl' = 1, \cdots, 5$}.
\]
Here, the $\intkernele_{\idxcl, \idxcl'}$ is parameterized by $J(p_1) = Gp_1$ with $p_1 = \tilde{m}_{\idxcl'}$.  Table \ref{tab:GSS_notation} shows how the GSS dynamics fits into the framework of \eqref{eq:2ndOrder}.
\begin{table}[H]
\centering
	\renewcommand{\arraystretch}{1}
	\small{\begin{tabular}{c | c | c | c | c | c | c | c } 
	\hline
	Category & $m_i$ & $\xi_i$         & $\numcl$  & $s^{\bx}_{i, i'}$ & $\forcex(\bx_i, \dot\bx_i)$ & $\intkernele_{\idxcl, \idxcl'}(r)$    & $\intkernela$\\
	\hline
	Value    & $1$   & $\emptyset$ & $N$           & $\emptyset$    & $\emptyset$                 & $\frac{G\tilde{m_{\idxcl'}}}{r^3}$   & $\emptyset$\\
	\hline
	\end{tabular}}
	\renewcommand{\arraystretch}{1}
	\caption{(GSS) Mapping \eqref{eq:2ndOrder}}
	\label{tab:GSS_notation}
\vspace{0.3cm}
	\renewcommand{\arraystretch}{1}
	\small{\begin{tabular}{ c | c | c | c }
	\hline
	$M$   & $d$ & $T_f$              & $T$         \\
	\hline
	$500$ & $2$ & $913 \text{day}$ & $182.6 \text{day}$ \\
	\hline
	\end{tabular}}  
	\renewcommand{\arraystretch}{1}
	\caption{(GSS) Parameters for Experiment Setup}
	\label{tab:GSS_params} 
\end{table}
We also use the parameters in table \ref{tab:GSS_params} to set up the experiment. These parameters are based on simple astronomical features of the system and are used for simulation of the dynamics and getting an appropriate and realistic number of observations. We use piecewise linear polynomials with $n_{\idxcl, \idxcl'}^E = 100$ for $\intkernele_{\idxcl, \idxcl'}$ when $\idxcl \neq \idxcl'$, and piecewise constant polynomials with $n_{\idxcl, \idxcl}^E = 1$ for $\intkernele_{\idxcl, \idxcl}$.  Each AO is given an index as follows: the Sun is assigned an index $1$, and depending on the distance from the Sun, the index is increased gradually, and stopping at $5$ for Mars.  We use the following units: $1$ day for time scale, $10^{6}$ km for length scale, and $10^{24}$ kg for mass scale.  The gravitational constant $G$ becomes $8.64^2 \cdot 6.67408 \cdot 10^{-6} (10^{6}\text{km})^3(10^{24}\text{kg})^{-1}\text{day}^{-2}$.  We also use the following data from NASA in table \ref{tab:nasa_data}.
\begin{table}[H]
\centering
\small{\begin{tabular}{ c | c | c | c | c | c }
\hline
Category                       & Sun                 & Mercury & Venus  & Earth   & Mars    \\
\hline
Mass ($10^{24}\text{kg}$)      & $1.989\cdot 10^{6}$ & $0.33$ & $4.87$  & $5.97$  & $0.642$ \\
\hline
Perihelion ($10^{6}\text{km}$) &          N/A        & $46$   & $107.5$ & $147.1$ & $206.6$ \\
\hline
Aphelion  ($10^{6}\text{km}$)  &          N/A        & $69.9$ & $108.9$ & $152.1$ & $249.2$ \\
\hline
Orbital Period ($\text{day}$)  &          N/A        & $88$   & $224.7$ & $365.2$ & $687$   \\
\hline
\end{tabular}}  
\caption{(GSS) NASA Data for Each AO}
\label{tab:nasa_data} 
\end{table}
The initial position distribution for the astronomical objects, $\muX$, is constructed as follows:  the Sun is always placed at the origin, whereas the planets are randomly placed on ellipses with their corresponding perihelion and aphelion data, and the Sun is sitting at one of the foci (Sun is the common focus of all initial elliptical trajectory).  We construct a distribution, $\muV$, which gives the initial velocities for the astronomical objects as follows: the Sun always has zero initial velocity, whereas the planets will have their initial velocity depending on their initial position and satisfying the Vis-Viva equation (see \cite{logsdon_1998} for details).  The comparison of the true $\intkernele_{\idxcl, \idxcl'}$'s and the estimated $\lintkernele_{\idxcl, \idxcl'}$'s is shown in Fig.\ref{fig:GSS_phis}.
We inferred a total of $N^2 = 25$ different interaction laws all together from the observation data.  As shown in Fig. \ref{fig:GSS_phis}, the interactions from planets on the Sun and the Sun on planets are estimated with high accuracy, however the estimated inter-planet interactions offer little valuable insight of the original interactions. This is likely driven by the domination of the sun in terms of effect on the dynamics -- due to its mass. The effect of the Sun's mass creates a form of ill-posedness of the system which affects the accuracy of our estimation. Realizing the possibility of a parametric form of the interaction laws, we go through a delicate decoupling procedure detailed in \ref{sec:gss_parametric_form}, and produce a cleaned up version of $\lintkernele_{\idxcl, \idxcl'}$'s, shown in Fig. \ref{fig:GSS_phis_clean}.
As shown in Fig. \ref{fig:GSS_phis_clean}, we are able to de-noise the original estimators and obtain a much cleaner presentation of the interaction laws.  Relative $L^2(\rho_T)$-errors for each $\intkernele_{\idxcl, \idxcl'}$ are provided in tables \ref{tab:GSS_phi_err} and \ref{tab:GSS_phi_err_clean} in order to re-affirm our claim.
\begin{table}[H]
\centering
\resizebox{\columnwidth}{!}{%
\small{\begin{tabular}{c || c | c | c | c | c} 
\hline
            &$\idxcl' = 1$                               &$\idxcl' = 2$                             &$\idxcl' = 3$                            &$\idxcl' = 4$                             &$\idxcl' = 5$  \\
\hline
\hline
$\idxcl = 1$&$0$                                         &$1.6199 \cdot 10^{-4} \pm 7 \cdot 10^{-8}$&$1.13 \cdot 10^{-7} \pm 3 \cdot 10^{-9}$ &$7.61 \cdot 10^{-7} \pm 6 \cdot 10^{-9}$&$2.71 \cdot 10^{-5} \pm 2 \cdot 10^{-7}$\\
\hline
$\idxcl = 2$&$1.6196 \cdot 10^{-4} \pm 8 \cdot 10^{-8}$&$0$                                       &$1.5 \cdot 10^{-1} \pm 2 \cdot 10^{-2}$  &$3.5 \cdot 10^{-1} \pm 7 \cdot 10^{-2}$ &$9 \pm 1.7$\\
\hline            
$\idxcl = 3$&$1.03 \cdot 10^{-7} \pm 6 \cdot 10^{-9}$  &$3.1 \cdot 10^{-2} \pm 8 \cdot 10^{-3}$   &$0$                                        &$1.42 \cdot 10^{-4} \pm 3 \cdot 10^{-4}$&$10 \cdot 10^{-2} \pm 1 \cdot 10^{-2}$\\
\hline
$\idxcl = 4$&$7.57 \cdot 10^{-7} \pm 3 \cdot 10^{-9}$  &$3 \cdot 10^{-2} \pm 1 \cdot 10^{-2}$     &$1.402 \cdot 10^{-2} \pm 7 \cdot 10^{-5}$&$0$                                       &$2.2 \cdot 10^{-2} \pm 2 \cdot 10^{-3}$\\
\hline   
$\idxcl = 5$&$2.717 \cdot 10^{-5} \pm 3 \cdot 10^{-8}$ &$8 \cdot 10^{-1} \pm 2 \cdot 10^{-1}$     &$3.3 \cdot 10^{-2} \pm 4 \cdot 10^{-3}$  &$1.8 \cdot 10^{-2} \pm 3 \cdot 10^{-3}$ &$0$\\
\hline 
\end{tabular}}
}
\caption{(GSS) Relative errors for the estimators, $\lintkernele_{\idxcl, \idxcl'}$ \rev{ (calculated using \eqref{eq:L2rhoTE})}.}
\label{tab:GSS_phi_err}
\end{table}

\begin{figure}[H]
\begin{subfigure}{\textwidth}
  \centering
   \includegraphics[width=0.7\textwidth]{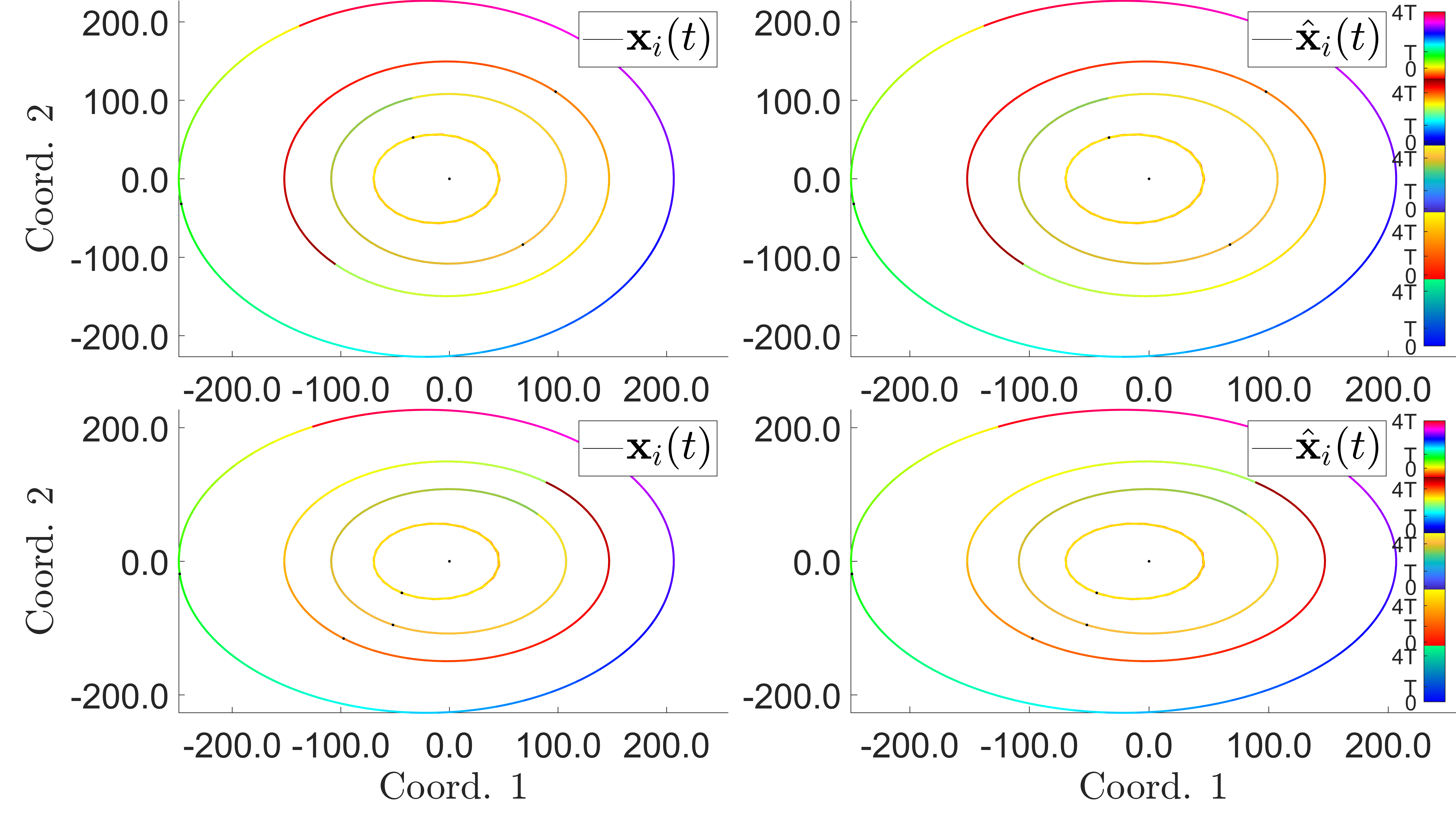} 
\end{subfigure}
\caption{(GSS) Comparison of $\bX$ and $\hat\bX$, with the errors reported in table \ref{tab:GSS_traj_err}. The first row of trajectories are generated from an initial condition taken from the observation data. The second row of trajectories are generated from another randomly chosen initial condition. The first column of trajectories are generated from the true interaction kernel, whereas the second column of trajectories are generated from our estimated kernel with the same initial conditions. \rev{The color of the trajectory indicates the flow of time, from $t = 0$ to $t = 4T$; and each AO uses a different set of colors, as given by the color bar on the right.}}
\label{fig:GSS_trajs}
\end{figure}

\begin{table}[H]
\centering
\resizebox{\columnwidth}{!}{%
\small{\begin{tabular}{c || c | c | c | c | c} 
\hline
            &$\idxcl' = 1$                          &$\idxcl' = 2$                              &$\idxcl' = 3$                           &$\idxcl' = 4$                           &$\idxcl' = 5$  \\
\hline
\hline
$\idxcl = 1$&$0$                                    &$1.742 \cdot 10^{-4} \pm 4 \cdot 10^{-7}$&$2 \cdot 10^{-5}   \pm 2 \cdot 10^{-5}$ &$8 \cdot 10^{-5} \pm 8 \cdot 10^{-5}$   &$5 \cdot 10^{-3} \pm 4 \cdot 10^{-3}$\\
\hline
$\idxcl = 2$&$2.9 \cdot 10^{-3} \pm 5 \cdot 10^{-4}$&$0$                                        &$1.5 \cdot 10^{-1} \pm 2 \cdot 10^{-2}$ &$2.4 \cdot 10^{-2} \pm 2 \cdot 10^{-3}$ &$5 \cdot 10^{-2} \pm 3 \cdot 10^{-2}$\\
\hline            
$\idxcl = 3$&$9.3 \cdot 10^{-3} \pm 4 \cdot 10^{-4}$&$3.49 \cdot 10^{-2} \pm 2 \cdot 10^{-4}$   &$0$                                     &$3.81 \cdot 10^{-2} \pm 3 \cdot 10^{-4}$&$1.5 \cdot 10^{-1} \pm 9 \cdot 10^{-2}$\\
\hline
$\idxcl = 4$&$1.7 \cdot 10^{-3} \pm 3 \cdot 10^{-4}$&$2.3 \cdot 10^{-2} \pm 2 \cdot 10^{-3}$    &$4.47 \cdot 10^{-2} \pm 4 \cdot 10^{-4}$&$0$                                     &$1.7 \cdot 10^{-1} \pm 9 \cdot 10^{-2}$\\
\hline   
$\idxcl = 5$&$7 \cdot 10^{-3} \pm 3 \cdot 10^{-3}$  &$5 \cdot 10^{-2} \pm 3 \cdot 10^{-2}$      &$1.5 \cdot 10^{-1} \pm 9 \cdot 10^{-2}$ &$1.7 \cdot 10^{-1} \pm 9 \cdot 10^{-2}$ &$0$\\
\hline 
\end{tabular}}
}
\caption{(GSS) Relative errors for the cleaned-up estimators, $\lintkernele_{\idxcl, \idxcl'}$ \rev{ (calculated using \eqref{eq:L2rhoTE})}.}
\label{tab:GSS_phi_err_clean}
\end{table}
The comparison of true trajectory $\bX(t)$ and $\hat\bX(t)$ is shown in Fig. \ref{fig:GSS_trajs}.

\begin{table}[H]
\centering
\small{\begin{tabular}{| c || c | c |} 
\hline
                                                 & $[0, T]$                                  &$[T, T_f]$\\
\hline
$\text{mean}_{\text{IC}}$: Training ICs on $\bx$ & $6.6 \cdot 10^{-4} \pm 2 \cdot 10^{-5}$ &$3.9 \cdot 10^{-3} \pm 2 \cdot 10^{-4}$\\
\hline
$\text{mean}_{\text{IC}}$: Training ICs on $\bv$ & $3.9 \cdot 10^{-3} \pm 1 \cdot 10^{-4}$&$2.13 \cdot 10^{-2} \pm 8 \cdot 10^{-4}$\\
\hline
$\text{std}_{\text{IC}}$:  Training ICs on $\bx$ & $5 \cdot 10^{-4} \pm 1 \cdot 10^{-4}$ &$2.7 \cdot 10^{-3} \pm 4 \cdot 10^{-4}$\\
\hline        
$\text{std}_{\text{IC}}$:  Training ICs on $\bv$ & $2.5 \cdot 10^{-3} \pm 3 \cdot 10^{-4}$ &$1.30 \cdot 10^{-2} \pm 2 \cdot 10^{-4}$\\
\hline    
\hline
$\text{mean}_{\text{IC}}$: Random ICs   on $\bx$ & $6.8 \cdot 10^{-4} \pm 2 \cdot 10^{-5}$ &$3.9 \cdot 10^{-3} \pm 1 \cdot 10^{-4}$\\
\hline
$\text{mean}_{\text{IC}}$: Random ICs   on $\bv$ & $3.9 \cdot 10^{-3} \pm 1 \cdot 10^{-4}$ &$2.13 \cdot 10^{-2} \pm 6 \cdot 10^{-4}$\\
\hline
$\text{mean}_{\text{IC}}$: Random ICs   on $\bx$ & $5.3 \cdot 10^{-4} \pm 1 \cdot 10^{-4}$ &$2.5 \cdot 10^{-3} \pm 3 \cdot 10^{-4}$\\
\hline
$\text{std}_{\text{IC}}$:  Random ICs   on $\bv$ & $2.6 \cdot 10^{-3} \pm 4 \cdot 10^{-4}$ &$1.2 \cdot 10^{-2} \pm 1 \cdot 10^{-3}$\\
\hline   
\end{tabular}}
\caption{(GSS) Trajectory Errors: Initial Conditions (ICs) used in the training set (first two rows), new ICs randomly drawn from $\muX$ (second set of two rows). \rev{The trajectory errors in $\bx$/$\bv$ is calculated using \eqref{eq:traj_norm_x}/\eqref{eq:traj_norm_v}.}}
\label{tab:GSS_traj_err}
\end{table}
Since the conservation of the sum of gravitational potential energy and kinetic energy of each planet produces the elliptical orbits around the Sun, we will consider the conservation of total energy (as the sum of gravitational potential energy and kinetic energy) of each planet and Sun as the emergent behavior.  The total energy for each planet at time $t$ is calculated as,
\[
E^{\text{total}}_i(t) = -\frac{G\tilde{m}_1\tilde{m}_i}{r_{i, 1}(t)} + \frac{\tilde{m}_is_i^2}{2}, \quad \text{for $i  = 2, \cdots, 5$}.
\]
$r_{i, 1}(t) = \norm{\bx_i(t) - \bx_1(t)}$ and $s_i = \norm{\bv_i(t)}$.  Then we consider the variance and mean of the total energy (associated to each planet) over time, i.e., 
\begin{align*}
\begin{cases}
E_i^{\text{Mean}} &= \text{Mean}_{l = 1}^L(E^{\text{total}}_i(t_l)) \\
E_i^{\text{Var}}  &= \text{Var}_{l = 1}^L(E^{\text{total}}_i(t_l)) \\
\end{cases}
\end{align*}
When $E_i^{\text{Var}} < 10^{-2}$ for $i = 2, \cdots 5$, we consider the total energy to be conserved.  Not surprisingly, with the total energy of the true system being always conserved, and with the predicted positions as well as their corresponding velocities of each AO estimated with about $10^{-2}$ relative errors, $100\%$ of the estimated systems show conservation of total energy. We also consider a set of Pattern Indicator scores to quantitatively measure the capability of our estimators to predict limit cycles correctly for GSS.  $\text{PI}_1$ measures the relative errors between the energy variance from the true system and the predicted system over $M$ trials.  And $\text{PI}_2$ measures the relative errors between the mean energy from the true system and the predicted system over $M$ trials.  The scores are reported in table \ref{tab:GSS_PIs}.
\begin{table}[H]
\centering
\small{\begin{tabular}{| c || c | c |} 
\hline
                                        &$\text{PI}_1$                             &$\text{PI}_2$\\
\hline
$\text{mean}_{\text{IC}}$: Training ICs &$2.85 \cdot 10^{-1} \pm 4 \cdot 10^{-3}$&$4.96 \cdot 10^{-6} \pm 4 \cdot 10^{-8}$\\
\hline
$\text{std}_{\text{IC}}$:  Training ICs &$1.15 \cdot 10^{-1} \pm 3 \cdot 10^{-3}$&$8.9 \cdot 10^{-7} \pm 2 \cdot 10^{-8}$\\
\hline        
\hline   
$\text{mean}_{\text{IC}}$: Random ICs   &$2.87 \cdot 10^{-1} \pm 5 \cdot 10^{-3}$&$4.94 \cdot 10^{-6} \pm 4 \cdot 10^{-8}$\\
\hline
$\text{std}_{\text{IC}}$:  Random ICs   &$1.16 \cdot 10^{-1} \pm 3 \cdot 10^{-3}$&$9.1 \cdot 10^{-7} \pm 2 \cdot 10^{-8}$\\
\hline   
\end{tabular}}
\caption{(GSS) Pattern Indicator Scores: ICs used in the training set (first two rows), new ICs randomly drawn from $\muX$ (second set of two rows).}
\label{tab:GSS_PIs}
\end{table}
Notice that the original system has its total energy variance being close to zero, and we are able to reproduce the total energy variance which is close to zero; moreover, the total energy of the predicted system for each planet resembles closely of its counterpart in the true system.

\rev{We have studied the Solar system as an interacting agent-based systems where each agent representing a different type due to the mass-based gravity.  However, it is clear that there is only one underlying interaction law with an associated parameter (the mass) for each agent.  Our learning approach performs well without any knowledge of this structure and produces each pairwise interaction as a different function.  But in fact, they are a family of functions parameterized by one single parameter, which is the mass of each agent.  Therefore, in this subsequent section, we proceed to show that the learned functions are close to the known gravitational kernel and that we can discover the underlying masses using an appropriate decoupling procedure.} 
\subsection{Discovery of the Parametric Form}\label{sec:gss_parametric_form}
\begin{figure}[H]
\begin{subfigure}{\textwidth}
  \centering
   \includegraphics[width=0.7\textwidth]{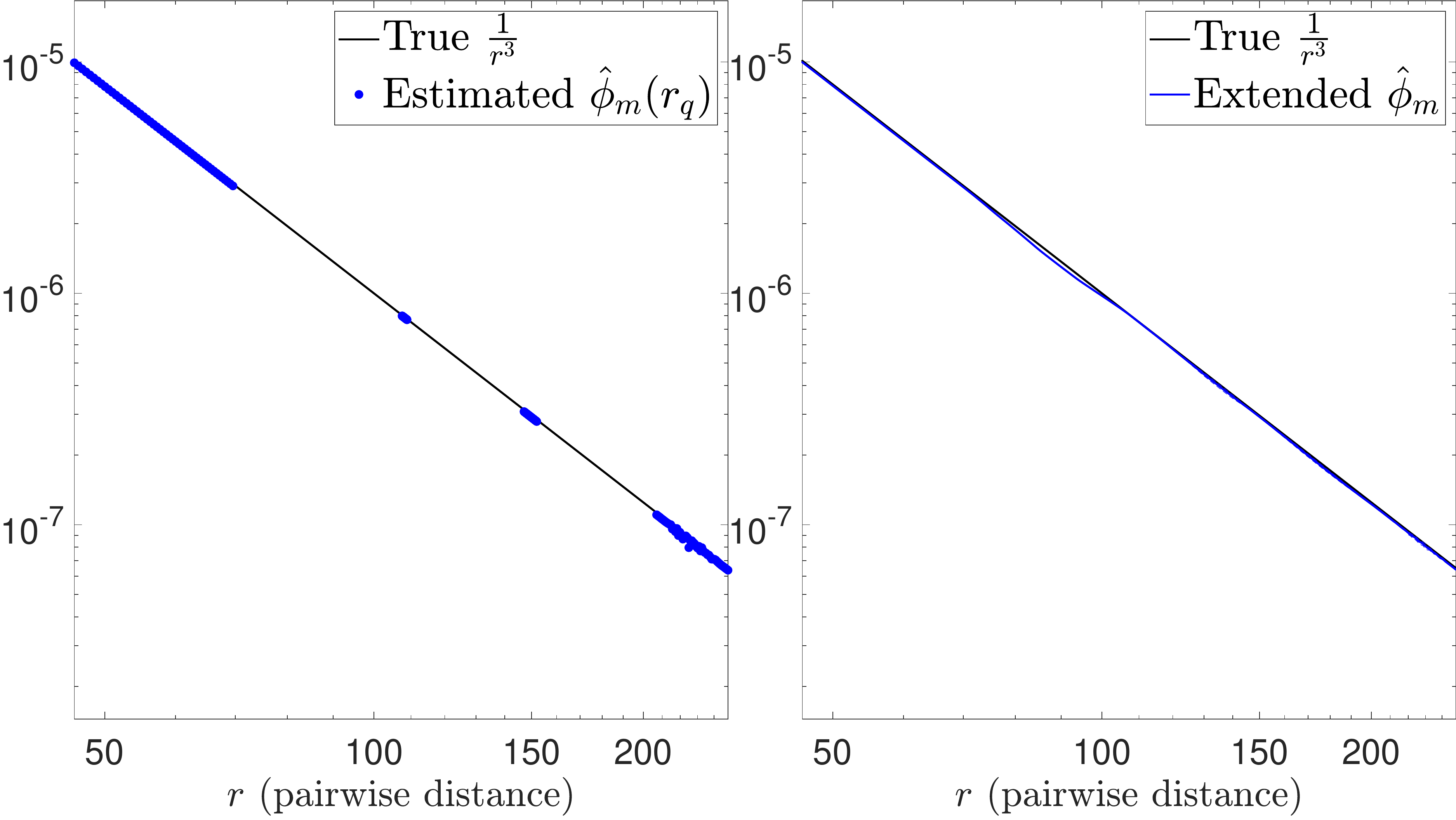} 
\end{subfigure}
\caption{(GSS) Extension from discrete $\{\lintkernel_m^{\bx, *}(r_q)\}_{q = 1}^Q$'s to a continuous $\lintkernele_m$. The $\frac{1}{r^3}$ line is shown as a reference line and it is not used in the learning of $\{\lintkernel_m^{\bx, *}(r_q)\}_{q = 1}^Q$'s nor the extension procedure.}
\label{fig:GSS_phim}
\end{figure}

Having examined the behaviors of $\lintkernele_{1, \idxcl'}$ and $\lintkernele_{\idxcl', 1}$ (for $\idxcl' = 2, \cdots, N$) closely, we observe an interesting behavior of our estimators, which is that $\lintkernele_{1, \idxcl'}$ and $\lintkernele_{\idxcl', 1}$ (for $\idxcl' \neq 1$) behave roughly the same, except at different scales.  Such behavior prompts us to consider a single-parameter parametric structure of $\lintkernele_{\idxcl, \idxcl'}$'s, i.e.,
\[
\lintkernele_{\idxcl, \idxcl'}(r) \approx \beta_{\idxcl'}\lintkernele_m(r) \quad \text{for $\idxcl \neq \idxcl'$ with $\beta_{\idxcl'} > 0$}.
\]
\begin{remark}
We do not assume any particular form of $\lintkernele_m(r)$, except that $\lintkernele_m(r)$ being continuous.
\end{remark}
In fact, the original gravitational interaction kernels are parameterized by $G\tilde{m}_{i'}$, i.e. $\intkernele_{\idxcl, \idxcl'}(r) = G\tilde{m}_{\idxcl'}\cdot\frac{1}{r^3}$.
\begin{remark}
The gravitational constant $G$ represents the length and time scales on which the experiment is conducted, and it will not be identifiable by our decoupling procedure.  Therefore, we assume that $G$ is known.  In fact, the first implicit measurement of $G$ with about $1\%$ accuracy is attributed to Henry Cavendish in the Cavendish experiment performed in $1797 - 1798$, and the result was published in Philosophical Transactions of the Royal Society.  Using the estimated $G$, with the radius of Earth first calculated by the Greek mathematician Eratosthenes in approximately $230$ BC, and the gravitational acceleration, $g \approx 9.8 \text{m}/\text{sec}^2$, determined by Galileo in the $16^{\text{th}}$ century, one can calculate the mass of the Earth, by connecting Newton's second law and universal law of gravitation, to get $M_{\text{Earth}} = 5.98 \cdot 10^{24} \text{kg}$.
\end{remark}
Since $\lintkernele_{\idxcl, \idxcl'} \approx \intkernele_{\idxcl, \idxcl'}$ (for $\idxcl = 1$ or $\idxcl' = 1$), we want to decouple $\beta_{\idxcl'}$ and $\lintkernele_m(r)$ from $\lintkernele_{\idxcl, \idxcl'}$ through a three-step optimization procedure.  First, we consider a sequence of points $\{r_q\}_{q = 1}^Q$ from the supports of $\lintkernele_{1, \idxcl'}$ for $\idxcl' = 2, \cdots N$ ($r_q$'s are taken as the centers of the sub-intervals where the basis functions are built), and the following loss function,
\begin{align*}
f_1(\beta_1, \cdots, \beta_N, \lintkernele_m(r_1), \cdots, \lintkernele_m(r_Q)) &= \sum_{\idxcl = 2}^N\sum_{q = 1}^Q(\lintkernele_{\idxcl, 1}(r_q) - \beta_1\lintkernele_m(r_q))^2d\rho_T^{L, M, \bx, \idxcl, 1}(r_q) \\
&\qquad + \sum_{\idxcl' = 2}^N\sum_{q = 1}^Q(\lintkernele_{1, \idxcl'}(r_q) - \beta_{\idxcl'}\lintkernele_m(r_q))^2d\rho_T^{L, M, \bx, 1, \idxcl'}(r_q)
\end{align*}
$f_1$ is minimized over $\beta_{\idxcl'} \ge 0$ for $\idxcl' = 1, \cdots, N$ and $\lintkernele_m(r_q) \in R$ for $q = 1, \cdots, Q$.  We only keep portion of the minimizer, namely, $\{\lintkernel_m^{\bx, *}(r_q)\}_{q = 1}^Q$, due to the fact that the Sun related terms have significantly more dominance in $f_1$.  Second, we extend the discrete values of $\{\lintkernel_m^{\bx, *}(r_q)\}_{q = 1}^Q$ to a continuous function, and express $\lintkernele_m$ as a linear combination of basis functions $\psi_{\eta}$ (clamped B-spline functions of degree $2$) over the interval $[R_1, R_2]$, where
\[
R_1 = \min_{\idxcl, \idxcl' = 1,\dots,\numcl} \{R_{\idxcl, \idxcl'}^{\min}\}, \quad R_2 = \max_{\idxcl, \idxcl'=1,\dots,\numcl}\{R_{\idxcl, \idxcl'}^{\max}\}, \quad \text{with $\text{supp}(\lintkernele_{\idxcl, \idxcl'}) = [R_{\idxcl, \idxcl'}^{\min}, R_{\idxcl, \idxcl'}^{\max}]$}.
\]
Hence, 
\[
\lintkernele_m(r) = \sum_{\eta = 1}^Q \alpha_{\eta}\psi_{\eta}(r).
\]
Then, we do a regularized least square fit to $\{\lintkernel_m^{\bx, *}(r_q)\}_{q = 1}^Q$, using the following loss function,
\[
f_2(\alpha_1, \cdots, \alpha_Q) = \sum_{q = 1}^Q \sum_{\eta = 1}^Q (\alpha_{\eta}\psi_{\eta}(r_q) - \lintkernel_m^{\bx, *}(r_q))^2 + \lambda\int_{r = R_1}^{R_2} \big|\sum_{\eta = 1}^Q \alpha_{\eta}\psi_{\eta}''(r)\big|^2 \, dr.
\]
Here we take $\lambda = 10^{-3}$.  The result of extending the discrete points to a continuous function is shown in Fig. \ref{fig:GSS_phim}.
The last step is to use the discrete values, $\{\lintkernel_m^{\bx, *}(r_q)\}_{q = 1}^Q$, to learn the $\beta_{\idxcl'}$ again, using the following loss function,
\begin{align*}
f_3(\beta_1, \cdots, \beta_N) &= \sum_{\idxcl = 2}^N\sum_{q = 1}^Q \frac{(\lintkernele_{\idxcl, 1}(r_q) - \beta_1\lintkernel_m^{\bx, *}(r_q))^2d\rho_T^{L, M, \bx, \idxcl, 1}(r_q) }{\sum_{q = 1}^Q(\lintkernele_{\idxcl, 1}(r_q))^2d\rho_T^{L, M, \bx, \idxcl, 1}(r_q)} \\
&\qquad + \sum_{\idxcl' = 2}^N\sum_{q = 1}^Q\frac{(\lintkernele_{1, \idxcl'}(r_q) - \beta_{\idxcl'}\lintkernel_m^{\bx, *}(r_q))^2d\rho_T^{L, M, \bx, 1, \idxcl'}(r_q)}{\sum_{q = 1}^Q(\lintkernele_{1, \idxcl'}(r_q))^2d\rho_T^{L, M, \bx, 1, \idxcl'}(r_q)}
\end{align*}
The re-scaling by $\sum_{q = 1}^Q(\lintkernele_{\idxcl, 1}(r_q))^2d\rho_T^{L, M, \bx, \idxcl, 1}(r_q)$ and $\sum_{q = 1}^Q(\lintkernele_{1, \idxcl'}(r_q))^2d\rho_T^{L, M, \bx, 1, \idxcl'}(r_q)$ is to keep all terms balanced, and in this particular instance it especially counters the dominance of the mass of the Sun, whose mass takes up more than $95\%$ of the mass of the whole solar system. The appropriate use of the dynamics-adapted measures enables us to identify parameters correctly. $f_3$ is minimized over $\beta_{\idxcl'} \ge 0$ for $\idxcl' = 1, \cdots, N$.  The minimizer $\beta_{\idxcl'}^*$ together with $\lintkernele_m$ (from the previous two steps), will have the following form
\[
\beta_{\idxcl'} = C_1\tilde{m}_{\idxcl'} \quad \text{for $\idxcl' = 1, \cdots, N$}\,,,
\]
and
\[
\lintkernele_m(r) = \frac{C_2}{r^3}, \quad \text{with $C_1C_2 = G$}.
\]
In order to offer deeper understanding of the difficulty of estimating the masses of each AO, we provide the relative errors for estimating the mass of each astronomical object in table \ref{tab:gss_mass_err} along with the mean and standard deviation of the estimated masses. 
\begin{table}[H]
\centering
\resizebox{\columnwidth}{!}{%
\small{\begin{tabular}{c || c | c | c | c | c } 
\hline
               &Sun                                    &Mercury                                 & Venus                                & Earth                                  & Mars\\
\hline
True Mass      &$1.9885 \cdot 10^{6}$                  &$3.3 \cdot 10^{-1}$                     & $4.87$                               & $5.97$                                 & $6.42 \cdot 10^{-1}$\\
\hline
Estimated Mass &$2.01 \cdot 10^{6} \pm 1 \cdot 10^{4}$ &$3.35 \cdot 10^{-1} \pm 2 \cdot 10^{-3}$&$4.88 \pm 3 \cdot 10^{-2}$            &$6.05 \pm 4 \cdot 10^{-2}$              &$6.52 \cdot 10^{-1} \pm 5 \cdot 10^{-3}$ \\
\hline
Rel. Err.      &$1.1 \cdot 10^{-2} \pm 7 \cdot 10^{-3}$&$1.4 \cdot 10^{-2} \pm 6 \cdot 10^{-3}$ & $4 \cdot 10^{-3} \pm 4 \cdot 10^{-3}$& $1.3 \cdot 10^{-2} \pm 6 \cdot 10^{-3}$& $1.6 \cdot 10^{-2} \pm 8 \cdot 10^{-3}$ \\
\hline
\end{tabular}}
}
\caption{(GSS) True, Estimated Masses, and the Relative Errors.  Recall that the masses are measured in unit: $10^{24}$ kg. Notice the immense difference in the scales of the masses, with the mass of the Sun taking up over $99\%$ of the whole Solar system, which makes the mass estimation problem severely ill-posed.}
\label{tab:gss_mass_err}
\end{table}
\section{Conclusion}\label{sec:conclusion}
We have demonstrated the effectiveness and efficiency of a nonparametric inference procedure to estimate the governing structure of various kinds of collective dynamics from observation of short-time trajectory data.  Such estimators can be also used to predict the correct type of emergent behaviors of the observed systems at larger timescales than those obtained from the training data.  The governing models proposed in section \ref{sec:model} encompass a wide range of dynamical systems of significant theoretical and computational interests to the physics, biology, and social science communities; and the algorithm in section \ref{sec:algorithm} scales efficiently to a large number of homogeneous or heterogeneous agents.

The systems included first-order, one-dimensional interaction kernels (Opinion Dynamics in Sec. \ref{sec:examples_OD}), second-order one-dimensional interaction kernels (Cucker-Smale, Self-Propelling Particles in $2D/3D$, in Sec. \ref{sec:examples_CS} to \ref{sec:examples_FM3D}), first-order two-dimensional interaction kernels (Synchronized Oscillator in Sec. \ref{sec:examples_SOD}), and second-order families of one-dimensional interaction kernels with underlying, but unknown, single parameters (Gravitational System in Sec. \ref{sec:examples_GSS}). In all cases, our estimators exhibit high precision in terms of standard performance measures, as well as high accuracy at capturing the proper type of emergent behaviors as measured by the confusion matrix and pattern indicator scores appropriate to the system.  Our final example studied the intrinsic parametric structure of our learned estimators, which leads to the discovery of some fundamental physical concepts, such as accurate mass and the underlying shared kernel of $\frac{1}{r^2}$ for gravitational force.


Further study of more intricate parametric structure of the interaction laws is ongoing as well as the theoretical foundations of the systems (\ref{eq:1stOrder}),(\ref{eq:2ndOrder}).  We are also preparing the study of emergent behaviors on more complex systems with more elaborate interaction laws and governing structures.

\vskip0.5cm
\noindent{\bf{Acknolwedgements}}. We acknowledge support from NSF-ATD-1737984, AFOSR FA9550-17-1-0280, NSF-IIS-1546392, NSF-IIS-1837991, NIH - T32GM119998; we thank Duke University and Prisma Analytics Inc. for free use of computing resources.

\appendix
\section{Performance Measures}\label{sec:PM_app}
Similar to what we have defined for measuring the performance of $\blintkernele$, we will use $\rho_T^{\xi, \idxcl, \idxcl'}$ to give the performance indicators of $\blintkernelxi$ in first order systems.  Similarly we have
\begin{equation}\label{eq:rhoTXi_1storder}
\begin{cases}
\rho_T^{\xi, \idxcl, \idxcl'}(r, s^{\xi}) &= \frac{1}{N_{\idxcl, \idxcl'}T}\int_{t = 0}^T \mathbb{E}_{\bY_0 \sim \muY}\Big[ \sum_{\substack{i \in \cl_{\idxcl} \\ i' \in \cl_{\idxcl'} \\ i \neq i'}} \delta_{r_{i, i'}(t), s^{\xi}_{i, i'}(t)}(r, s^{\xi})\Big] \, dt,  \\\
\rho_T^{L, \xi, \idxcl, \idxcl'}(r, s^{\xi}) &= \frac{1}{N_{\idxcl, \idxcl'}L}\sum_{l = 1}^L \mathbb{E}_{\bY_0 \sim \muY}\Big[ \sum_{\substack{i \in \cl_{\idxcl} \\ i' \in \cl_{\idxcl'} \\ i \neq i'}} \delta_{r_{i, i'}(t_l), s^{\xi}_{i, i'}(t_l)}(r, s^{\xi})\Big], \\
\rho_T^{L, M, \xi, \idxcl, \idxcl'}(r, s^{\xi}) &= \frac{1}{N_{\idxcl, \idxcl'}LM}\sum_{l, m = 1}^{L, M} \sum_{\substack{i \in \cl_{\idxcl} \\ i' \in \cl_{\idxcl'} \\ i \neq i'}} \delta_{r_{i, i'}(t_l), s^{\xi}_{i, i'}(t_l)}(r, s^{\xi}).
\end{cases}
\end{equation}
For measuring the difference, $\intkernelxi_{\idxcl, \idxcl'} - \lintkernelxi_{\idxcl, \idxcl'}$, we use the following $L^2(\rho_T)$ norm,
\begin{equation}\label{eq:L2rhoTXi_1stOrder}
\norm{\intkernelxi_{\idxcl, \idxcl'} - \lintkernelxi_{\idxcl, \idxcl'}}_{L^2(\rho_T^{\xi, \idxcl, \idxcl'})}^2 = \int_{r = 0}^\infty \int_{s^{\xi} =-\infty}^\infty (\intkernelxi_{\idxcl, \idxcl'}(r, s^{\xi}) - \lintkernelxi_{\idxcl, \idxcl'}(r, s^{\xi}))^2 \, d\rho_T^{E, \idxcl, \idxcl'}(r, s^{\xi}).
\end{equation}
For $(\blintkernele, \blintkernela, \blintkernelxi)$ learned from any second-order system,  we will need two new sets of probability distributions.  First, for $\rho_T^{E, \idxcl, \idxcl'}$, it is the same as defined in \eqref{eq:rhoTE}.  Second we define $\rho_T^{\dot\bx, \idxcl, \idxcl'}$ as follows,
\begin{equation}\label{eq:rhoTA}
\begin{cases}
\rho_T^{\dot\bx, \idxcl, \idxcl'}(r, s^{\dot\bx}, \dot{r}) &= \frac{1}{N_{\idxcl, \idxcl'}T}\int_{t = 0}^T\mathbb{E}_{\bY_0 \sim \muY}\Big[ \sum_{\substack{i \in \cl_{\idxcl} \\ i' \in \cl_{\idxcl'} \\ i \neq i'}} \delta_{r_{i, i'}(t), s^{\dot\bx}_{i, i'}(t), \dot{r}_{i, i'}(t)}(r, s^{\dot\bx}, \dot{r})\Big] \, dt,  \\
\rho_T^{L, \dot\bx, \idxcl, \idxcl'}(r, s^{\dot\bx}, \dot{r}) &= \frac{1}{N_{\idxcl, \idxcl'}L}\sum_{l = 1}^L \mathbb{E}_{\bY_0 \sim \muY}\Big[ \sum_{\substack{i \in \cl_{\idxcl} \\ i' \in \cl_{\idxcl'} \\ i \neq i'}} \delta_{r_{i, i'}(t_l), s^{\dot\bx}_{i, i'}(t_l), \dot{r}_{i, i'}(t_l)}(r, s^{\dot\bx}, \dot{r}) \Big],\\
\rho_T^{L, M, \dot\bx, \idxcl, \idxcl'}(r, s^{\dot\bx}, \dot{r}) &= \frac{1}{N_{\idxcl, \idxcl'}LM}\sum_{l, m = 1}^{L, M} \sum_{\substack{i \in \cl_{\idxcl} \\ i' \in \cl_{\idxcl'} \\ i \neq i'}} \delta_{r_{i, i'}(t_l), s^{\dot\bx}_{i, i'}(t_l), \dot{r}_{i, i'}(t_l)}(r, s^{\dot\bx}, \dot{r}). 
\end{cases}
\end{equation}
Here, $\bY = \begin{bmatrix} \bX \\ \bV \\ \bXi \end{bmatrix}$, $\muY = \begin{bmatrix} \muX \\ \muV \\ \muXi \end{bmatrix}$, and $\dot{r}$, being not the derivative $r$, rather the pairwise speed data, e.g., $\dot{r}_{i, i'}(t) = \norm{\bv_{i'}(t) - \bv_i(t)}$.  Finally, $\rho_T^{\xi, \idxcl, \idxcl'}$ is defined slightly differently from \eqref{eq:rhoTXi_1storder},
\begin{equation}\label{eq:rhoTXi_2ndorder}
\begin{cases}
\rho_T^{\xi, \idxcl, \idxcl'}(r, s^{\xi}, \xi) &= \frac{1}{N_{\idxcl, \idxcl'}T}\int_{t = 0}^T \mathbb{E}_{\bY_0 \sim \muY}\Big[ \sum_{\substack{i \in \cl_{\idxcl} \\ i' \in \cl_{\idxcl'} \\ i \neq i'}} \delta_{r_{i, i'}(t), s^{\xi}_{i, i'}(t), \xi_{i, i'}(t)}(r, s^{\xi}, \xi)\Big] \, dt,  \\
\rho_T^{L, \xi, \idxcl, \idxcl'}(r, s^{\xi}, \xi) &= \frac{1}{N_{\idxcl, \idxcl'}L}\sum_{l = 1}^L \mathbb{E}_{\bY_0 \sim \muY}\Big[ \sum_{\substack{i \in \cl_{\idxcl} \\ i' \in \cl_{\idxcl'} \\ i \neq i'}} \delta_{r_{i, i'}(t_l), s^{\xi}_{i, i'}(t_l), \xi_{i, i'}(t)}(r, s^{\xi}, \xi)\Big], \\
\rho_T^{L, M, \xi, \idxcl, \idxcl'}(r, s^{\xi}) &= \frac{1}{N_{\idxcl, \idxcl'}LM}\sum_{l, m = 1}^{L, M} \sum_{\substack{i \in \cl_{\idxcl} \\ i' \in \cl_{\idxcl'} \\ i \neq i'}} \delta_{r_{i, i'}(t_l), s^{\xi}_{i, i'}(t_l)}(r, s^{\xi}). 
\end{cases}
\end{equation}
Here, $\xi_{i, i'}(t) = \abs{\xi_{i'}(t) - \xi_i(t)}$.  The prediction error, $\intkernele_{\idxcl, \idxcl'} - \lintkernele_{\idxcl, \idxcl'}$, is measured in the same norm defined in \eqref{eq:L2rhoTE}; for $\intkernelxi_{\idxcl, \idxcl'} - \lintkernelxi_{\idxcl, \idxcl'}$, but it is weighted differently,
\begin{equation}\label{eq:L2rhoTXi_2ndOrder}
\norm{\intkernelxi_{\idxcl, \idxcl'} - \lintkernelxi_{\idxcl, \idxcl'}}_{L^2(\rho_T^{\xi, \idxcl, \idxcl'})}^2 = \int_{r = 0}^\infty \int_{s^{\xi} =-\infty}^\infty (\intkernelxi_{\idxcl, \idxcl'}(r, s^{\xi}) - \lintkernelxi_{\idxcl, \idxcl'}(r, s^{\xi}))^2 \, \xi^2d\rho_T^{E, \idxcl, \idxcl'}(r, s^{\xi}, \xi).
\end{equation}
and for $\intkernela_{\idxcl, \idxcl'} - \lintkernela_{\idxcl, \idxcl'}$, the corresponding norm is defined as follows,
\begin{equation}\label{eq:L2rhoTA}
\norm{\intkernela_{\idxcl, \idxcl'} - \lintkernela_{\idxcl, \idxcl'}}_{L^2(\rho_T^{\dot\bx, \idxcl, \idxcl'})}^2 = \int_{r = 0}^\infty \int_{s^{\dot\bx} =-\infty}^\infty (\intkernela_{\idxcl, \idxcl'}(r, s^{\dot\bx}) - \lintkernela_{\idxcl, \idxcl'}(r, s^{\dot\bx}))^2 \, \dot{r}^2d\rho_T^{\dot\bx, \idxcl, \idxcl'}(r, s^{\dot\bx}, \dot{r}).
\end{equation}
\bibliography{LimitCycles_PhysicaD}
\end{document}